\documentclass{article}
\PassOptionsToPackage{numbers,sort&compress}{natbib}

\usepackage[main, final]{neurips_2026}

\usepackage[utf8]{inputenc} 
\usepackage[T1]{fontenc}    

\usepackage{amsfonts}       
\usepackage{nicefrac}       
\usepackage{hyperref}
\usepackage{url}

\usepackage{tikz, subcaption}
\usepackage{caption}
\usepackage{wrapfig}
\usepackage{graphicx}
\usepackage{bbm}
\usepackage{algorithm}
\usepackage{algpseudocode}
\usepackage{longtable}
\usepackage{booktabs}
\usepackage{array}
\usepackage{tabularx}
\usepackage{titletoc}
\usepackage[most]{tcolorbox}
\usepackage{amsmath, amssymb, booktabs, microtype}
\usepackage{listings, xcolor}
\usepackage{longtable}
\usepackage{tabularx}
\usepackage{ltxtable}  
\usepackage{booktabs} 
\usepackage{mathtools}
\usepackage{multirow}

\usepackage{xltabular}

\newcommand{\R}{\mathbb{R}}

\lstdefinestyle{pylight}{
  language=Python,
  basicstyle=\footnotesize\ttfamily,
  keywordstyle=\color{blue!70!black},
  commentstyle=\color{gray}\itshape,
  stringstyle=\color{green!50!black},
  numberstyle=\tiny\color{gray},
  numbers=left, numbersep=8pt,
  breaklines=true, breakatwhitespace=true,
  frame=single, framesep=4pt, rulecolor=\color{gray!50},
  showstringspaces=false, columns=fullflexible,
  tabsize=4,
}

\DeclareMathOperator*{\softmax}{softmax}

\title{Interpreting Neural Combinatorial Optimization via Evolving Programmatic Bottlenecks}

\author{
Haocheng Duan$^{1*}$,
Yuxin Guo$^{1*}$,
Jieyi Bi$^2$,
Anqi Xie$^1$,\\
\textbf{Sirui Li}$^{3\dagger}$,
\textbf{Yining Ma}$^{4\ddagger}$,
\textbf{Cathy Wu}$^4$\\\\
$^1$Carnegie Mellon University,
$^2$Nanyang Technological University,\\
$^3$Microsoft Research, 
$^4$Massachusetts Institute of Technology
}

\begin{document}

\maketitle
{\let\thefootnote\relax\footnotetext{$^*$Equal contribution. \quad $^\dagger$Work done outside employment. \quad $^\ddagger$Corresponding author: Yining Ma (yiningma@mit.edu).}}

\begin{abstract} 
While Neural Combinatorial Optimization (NCO) achieves strong performance, its interpretability remains an open question. In this paper, we introduce Evolving Programmatic Bottlenecks (EPB), the first post-hoc, surrogate-based interpretability method for NCO, which explains a black-box policy by distilling it into an interpretable behavior surrogate. A natural surrogate is a Concept Bottleneck Model (CBM), yet standard concepts are input attributes, whereas NCO is better explained by decision rules that are hard to predefine. EPB thus uses LLMs to evolve a bank of programmatic decision rules as concepts, each producing an action distribution, where a learned state-dependent router then mixes them to form the surrogate action. We optimize EPB through an iterative framework: Block I fixes the bank size and introduces the hybrid textual-numerical gradient descent to jointly optimize the textual programs and the numerical router; Block II adapts the bank size via failure-targeted expansion and redundancy pruning. Extensive experiments demonstrate that the EPB surrogate closely matches the teacher’s behavior and solution quality. We then analyze the distilled rules, reveal several NCO decision patterns, and verify the rules' faithfulness to the surrogate’s decisions. Moreover, the surrogate allows targeted program edits that improve its generalization beyond the teacher. Our work advances interpretable NCO and establishes EPB as a promising direction for interpreting complex learned decision-making policies.
The code and the distilled surrogate are available at: \url{https://github.com/HaochengDuan/EPB}.
\end{abstract}

\section{Introduction}

\label{sec:intro}

Neural Combinatorial Optimization (NCO) has achieved remarkable success, with learned policies now matching strong classical heuristics on complex routing and scheduling \cite{kool2019attention, kwon2020pomo, luo2023neural}. However, this success hides a critical interpretability deficit. Current NCO frameworks function as black boxes, providing little insight into what decision strategies have been learned and why a particular action is favored by them. This hinders {informed} and {trustworthy} deployment in {industry} and prevents interpretability from catalyzing future NCO research in {academia}, as it has in computer vision~\cite{zeiler2014visualizing, steinmann2023learning, wu2023discover}.

In this paper, we study post-hoc, {surrogate-based} interpretability for NCO, which aims at distilling the black-box NCO policy into an interpretable student surrogate that mimics the teacher’s decisions. 
The surrogate serves as both an interpretable {behavioral} proxy for analysis and a potential deployable alternative (e.g., AlphaDev~\cite{mankowitz_faster_2023}). Concept Bottleneck Models (CBMs)~\cite{koh2020concept} offer a natural structure for such a surrogate~\cite{ghosh2023dividing}, grounding it in human-understandable \emph{concepts} that one can read, audit, and improve. Yet adapting CBMs to NCO is non-trivial. First, existing CBM concepts are typically atomic \emph{attributes} or \emph{features} (e.g., colors and textures in vision), whereas the more natural explanatory unit for NCO is a \emph{decision rule} (e.g., a nearest-neighbor heuristic) that determines action choices. Second, while classical heuristics offer a starting point, such a rule library may be intractable to construct since the algorithmic design space spans combinatorially many decision rules and hyperparameters.

\begin{wrapfigure}{r}{5.2cm}
\vspace{-15pt}
    \centering
\includegraphics[width=0.97\linewidth]{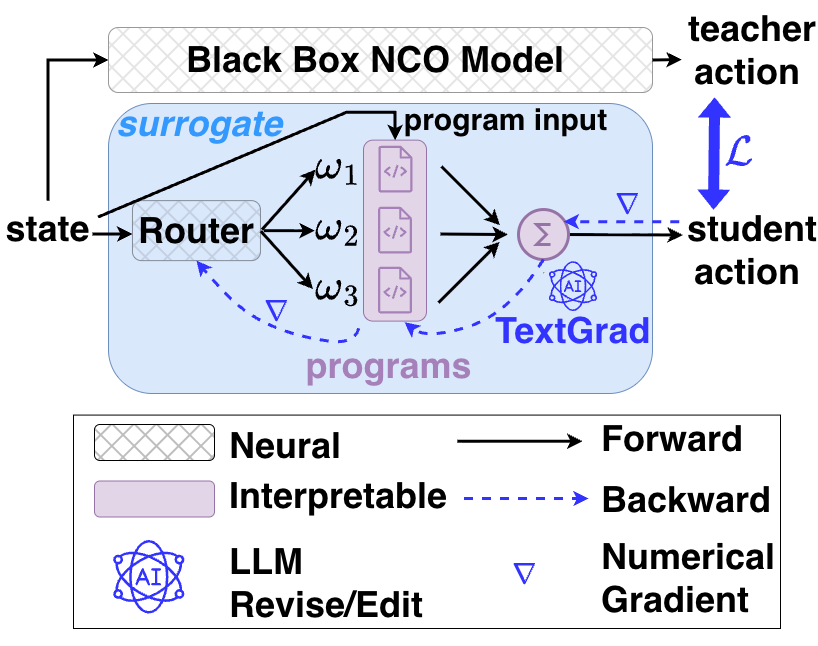}
    \caption{Framework of Evolving Programmatic Bottlenecks (EPB).}
    \label{fig:cbm_epb}
\vspace{-5pt}
\end{wrapfigure}
We introduce \textbf{Evolving Programmatic Bottlenecks (EPB)}. Rather than relying on a fixed, predefined concept library, EPB~(Figure~\ref{fig:cbm_epb}) uses large language models (LLMs) to automatically discover and evolve a bank of executable programs, each representing a decision rule as the concept. A state-dependent router then weights the programs’ action distribution. This distills the teacher NCO model into a student surrogate, where both the decision rules (programs) and their contributions to the final prediction (weights) are directly inspectable. To our knowledge, EPB is the first framework to provide post-hoc, surrogate-based interpretations of NCO.

Learning this surrogate requires jointly evolving the program bank and training the neural router to combine the programs. This coupling raises two challenges: errors arise from the mixture and do not directly identify which programs need revision, and each revision changes how the router should weight the program. Moreover, as programs evolve, the bank may leave some teacher behaviors uncovered or accumulate redundant programs. EPB addresses them through two alternating optimization blocks. \textbf{Block~I} fixes the bank size and jointly refines the programs and router through \emph{Hybrid Textual-Numerical Gradient Descent}. Routing weights and deviations from the teacher guide feedback allocation to programs, which TextGrad~\cite{yuksekgonul2024textgrad} revises, while numerical gradients update the router. \textbf{Block~II} adapts the bank through two operations: \textsc{Add} creates new programs to address unresolved failures, while \textsc{Drop} removes redundant programs.

We apply EPB to two representative NCO models, POMO \cite{kwon2020pomo} and LEHD \cite{luo2023neural}, across two routing tasks, TSP and CVRP. Results show that EPB surrogates remain close to the teachers in decision behavior and solution quality. 
Based on the surrogates, we analyze their decision patterns both at individual steps and in aggregate across instances, revealing, e.g., why they deviate from nearest neighbors and how their strategies shift as routes progress.
We further provide counterfactual and program knockout tests that support the faithfulness of these rules to the surrogate’s decisions.
Notably, while matching their teachers in distribution, the POMO surrogates outperform in generalizing to larger instances, suggesting that the distilled programs may be less tied to the training size than the teacher’s neural representations. Finally, because the surrogate contains white-box decision rules, we show that targeted program edits further improve its performance, such as out-of-distribution generalization.

\textbf{Contributions:} (i) we propose EPB, to our knowledge, the first framework to distill NCO policies into interpretable surrogates by using LLMs to evolve programs as concepts; (ii) we propose an iterative framework alternating between hybrid textual-numerical gradient descent and dynamic bank-size management; (iii) we validate the surrogate’s fidelity in behavior and solution cost, and provide faithful explanations of individual and aggregated decision patterns; and (iv) we demonstrate its generalization to larger instances, and further gains beyond the teacher through program edits.

\section{Related Work}
\textbf{Neural combinatorial optimization (NCO).}
Classical heuristics rely on hand-crafted decision rules that require problem-specific expertise. NCO instead learns policies from data, through constructive policies \cite{vinyals2015pointer,bello2016neural,nazari2018reinforcement,kool2019attention,kwon2020pomo,luo2023neural} and diffusion-based solvers \cite{sun2023difusco}, with test-time scaling and search further enhancing performance \cite{choo2022sgbs,hottung2022efficient,ma2021learning,wu2021learning}. However, the learned decision logic remains implicit in neural parameters and difficult to interpret at the level of individual choices. Early NCO interpretability work, such as probing~\cite{zhang2025probing}, reveals what information NCO representations encode but does not yield executable decision rules. EPB pioneers the first behavioral surrogate study for NCO.

\textbf{Surrogate-based Interpretability.}
Interpretability can be pursued by analyzing a model’s internals, as in mechanistic interpretability~\cite{geiger2023causal, michaud2024mips}, by learning inherently interpretable models, such as programmatic policies in reinforcement learning (RL)~\cite{delfosse2023nudge, trivedi2021synthesis}, or by fitting a post-hoc interpretable surrogate that mimics a trained model’s behavior~\cite{craven1995extracting}. A surrogate can be inspected, verified, or deployed in place of the black box \cite{bastani2018verifiable}, though it explains behavior rather than mechanism. Common surrogate families include trees \cite{craven1995extracting, frosst2017distilling, bastani2018verifiable}, additive models \cite{agarwal2021neural}, tree mixtures \cite{vasic2022moet}, and concept-based models \cite{ghosh2023dividing, moreira2024diconstruct}. Conceptually close to EPB, PIRL~\cite{verma2018programmatically} learns a single program that imitates a neural policy within a user-provided sketch for continuous control tasks. EPB instead uses LLMs to evolve a bank of programs as concepts, combined by a neural router to distill and interpret NCO policies.

\textbf{Concept bottleneck models (CBMs).} CBMs \cite{koh2020concept, kim2018tcav, ghorbani2019ace} link predictions to human-readable concepts, mostly in the form of input attributes. Later CBM extensions reduce annotation requirements \cite{yuksekgonul2022post, oikarinen2023label, srivastava2024vlgcbm} and broaden applications \cite{kulkarni2025cbae, sun2025cbllm}. Further work explores concept search \cite{feng2026bayesian}, expert composition \cite{de2026mixture}, and concept-based RL \cite{ye2025licorice}. More recently, DiSciPLE \cite{mall2025disciple} evolves programs with LLMs, but uses them as feature extractors rather than as state-to-action rules. Explaining NCO decisions, however, requires showing how each decision state gives rise to the action choices. EPB therefore uses decision rules as concepts, each implemented as a program that maps a state to an action distribution.

\textbf{LLM-guided algorithm discovery.}
EPB relies on LLMs to generate and revise programs, building on their ability to write code \cite{li2022alphacode, chen2021codex, austin2021program}, evolve programs and heuristics \cite{romera2024funsearch, liu2024eoh, ye2024reevo}, optimize text from feedback \cite{yuksekgonul2024textgrad}, and refine outputs through reflection or search \cite{shinn2023reflexion, yao2023tree}. Among them, FunSearch~\cite{romera2024funsearch}, EoH~\cite{liu2024eoh}, AlphaEvolve~\cite{novikov2025alphaevolve}, and ShinkaEvolve~\cite{lange2026shinkaevolve} evolve code to optimize task performance. EPB instead distills a trained NCO policy into interpretable programs to explain its behavior.

\section{Preliminaries}

\textbf{Neural combinatorial optimization (NCO).} 
A combinatorial optimization (CO) problem defines, for each problem instance $s \sim \mathcal{D}$, a feasible set $\mathcal{Y}(s)$ and an objective $J(y;s)$ over solutions $y = (a_{1:T})$, with the goal of finding $y^*(s) \in \arg\min_{y \in \mathcal{Y}(s)} J(y;s)$. We focus on routing problems, i.e., the Traveling Salesman Problem (TSP) and Capacitated Vehicle Routing Problem (CVRP). NCO learns to construct a solution $y=(a_1,\ldots,a_T)$ through sequential decisions (details in \S\ref{sec: tsp_cvrp}). Let $x_t = (s, a_{1:t-1})$ denote the decision state at step $t$, including the problem instance and the partial solution constructed so far.  An NCO model defines an autoregressive policy $\pi_\theta(a_t \mid x_t)$ to select the next node to visit, until a feasible solution is constructed. As representative teachers, we evaluate POMO~\cite{kwon2020pomo}, which leverages symmetries via multi-start rollouts, and LEHD~\cite{luo2023neural}, a light-encoder/heavy-decoder model trained on partial solutions for large-scale generalization (\S\ref{sec: teacher_models}).

\textbf{Concept bottleneck models (CBM).} A standard CBM~\cite{koh2020concept} employs a two-stage pipeline, $y = f(g(x))$, where an encoder $g$ first maps the input $x$ to a set of human-interpretable concepts before a predictor $f$ makes the final decision based solely on these concepts. The key idea is that the decision is bottlenecked by human-understandable concepts, making it more trustworthy. EPB borrows this idea with evolved programs as concepts: at each state $s$, $g$ is a bank of LLM-evolved programs outputting action distributions $V_m(s)$, and $f$ is their router-weighted mixture $\sum_m w_m(s)\,V_m(s)$.

\newcommand{\teacher}{\pi_T}                
\newcommand{\student}{\pi_S}                
\newcommand{\bank}{\mathcal{B}}             

\newcommand{\loss}{\mathcal{L}}             
\newcommand{\KLL}{\mathrm{KL}}               

\newcommand{\indic}[1]{\mathbbm{1}\!\left[#1\right]}   

\newcommand{\Acal}{\mathcal{A}}             
\newcommand{\Scal}{\mathcal{S}}             
\newcommand{\Dcal}{\mathcal{D}}             
\newcommand{\nstar}{n^{\star}}              
\newcommand{\Vm}{V_m}                       
\newcommand{\wm}{w_m}                       
\newcommand{\KLm}{\KL_m}                    

\section{Methodology}
\label{sec:method}

An NCO policy outputs action distributions, so its natural explanatory unit is a decision rule: an executable program that maps the current state to preferences over actions. Following the CBM idea, we construct a student surrogate $\student(\cdot\mid s;\theta,\bank)$ including a bank $\bank$ of human-readable programs evolved by LLMs, each producing an action distribution $V_m(s)$ and serving as a \emph{programmatic bottleneck}, and a router with parameters $\theta$ that mixes these distributions into $\student(\cdot\mid s)$ at each state. 
We first describe the surrogate architecture and then introduce how it is trained via a distillation loss so that $\student(\cdot\mid s)$ matches the teacher’s (original NCO model's) distribution $\teacher(\cdot\mid s)$ at every state $s$.

\begin{figure}[t]
  \centering
  \includegraphics[width=\linewidth]{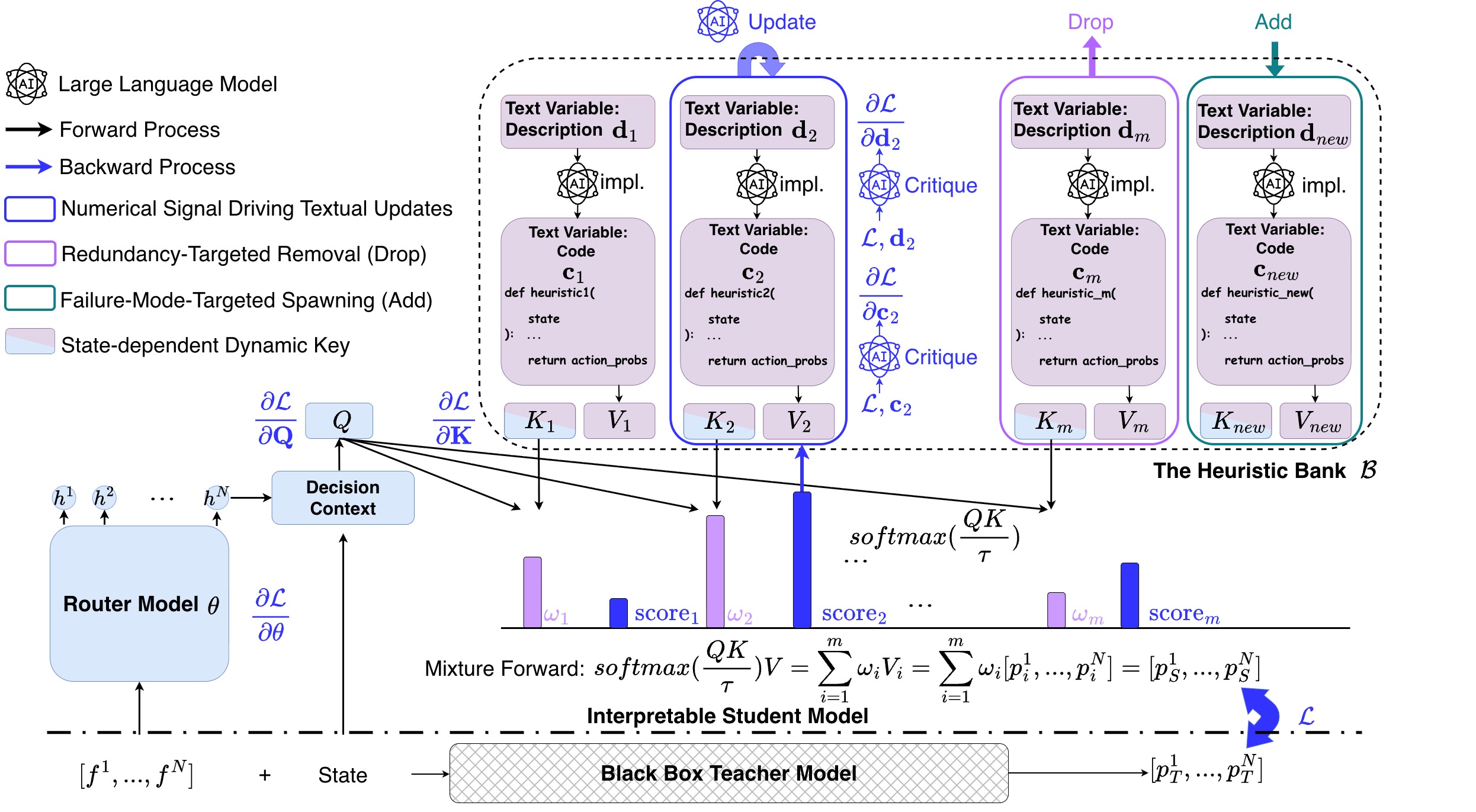}
  \caption{Overview of our Evolving Programmatic Bottlenecks (EPB) architecture.}
  \label{fig:overview}
\end{figure}
\subsection{The Student Surrogate with Programmatic Bottlenecks}
\label{sec:aacb}
At each decision state $s$, the surrogate produces an action distribution in three steps: (1) each program executes its code on $s$ to produce an action distribution $V_m(s)$; (2) the router assigns each program a state-dependent weight $w_m(s)$; and (3) the surrogate outputs the weighted mixture $\sum_m w_m(s)\,V_m(s)$. We describe each component below and depict the architecture of the surrogate in Fig.~\ref{fig:overview}.

\textbf{The program bank.} The bank $\bank = \{(c_m, d_m)\}_{m=1}^{M}$ collects $M$ programs, where $c_m$ is a program that maps the decision state to per-action scores and $d_m$ is a few-sentence description of the strategy it implements. The description guides revisions of the code (\S\ref{sec:hierarchical}), while the code runs in the forward pass. The bank size $M$ changes adaptively during training (\S\ref{sec:bank_evolution}).

\textbf{Per-program distributions.}
Each program $m$ produces a per-action distribution by executing its code $c_m$ on the current state and applying a softmax with temperature $\tau_h$:
\begin{equation}
  V_m(s)
  \;=\;
  \softmax\!\left(\, c_m(s) / \tau_h \,\right).
  \label{eq:vm}
\end{equation}

\textbf{State-dependent dynamic routing.}
\label{sec:ocdr}
The router, an attention module parameterized by $\theta$, determines how much each program contributes at each state. It assigns program $m$ the weight,
\begin{equation}
  w_m\left(s; \theta, \operatorname{sg}\left(\bank\right)\right)
  \;=\;
  \softmax_m\!\left(
    Q\left(s; \theta\right) \cdot K_m\left(s; \theta, \operatorname{sg}\left(\bank\right) \right) \,/\, \tau_r
  \right),
  \label{eq:routing}
\end{equation}
where $\tau_r$ is a temperature. The query $Q(s;\theta) \in \R^D$ embeds the current state, following the teacher's context encoding (e.g., for TSP: global information, the current-state context, and the first-node context). The key $K_m$ represents each program $m$, which is also state dependent as follows:

\begin{equation}
  K_m\left(s; \theta, \operatorname{sg}(\bank)\right)
  \;\coloneqq\;
  \sum_n \operatorname{sg}\left(V_m(s,n)\right) h(s,n;\theta)
  \;\in\; \R^D,
  \label{eq:K_dynamic}
\end{equation}
where $h(s,n;\theta)$ is the embedding of candidate node $n$. The dot product $Q \cdot K_m$ thus scores how well program $m$'s output fits the current state. The stop-gradient $\operatorname{sg}(\cdot)$ lets the router read the programs' outputs in the forward pass without passing gradients into them during backpropagation (\S\ref{sec:hybrid_gradient}).

We construct each key from the program’s output $V_m(s)$, rather than learning a separate embedding for each program. First, the key adds no additional trainable parameters, so programs can be added or dropped (\S\ref{sec:bank_evolution}) flexibly without changing the router’s architecture. Second, it represents each program in a simple, direct, yet effective way: a standard CBM concept is described by its activation on the input, whereas an EPB concept is a program, better described by what it recommends at the current state. Our ablation confirms that these keys achieve higher fidelity than freely learned static embeddings and fixed text embeddings with a learned projection (\S\ref{app:key_design}).

\textbf{Mixture forward.}
The student’s prediction is the routing-weighted mixture of the bank’s outputs:
\begin{equation}
  \student(\cdot \mid s)
  \;=\;
  \sum_{m=1}^{M}\, w_m(s; \theta, \operatorname{sg}(V)) \cdot V_m(s).
  \label{eq:mixture}
\end{equation}

\begin{algorithm}[t]
\caption{Two-Block Training}
\label{alg:two_block}
\begin{algorithmic}[1]
\Require Initial bank $\bank_0$, routing parameters $\theta_0$, dataset $\Dcal$, validation set $\Dcal_{\text{val}}$
\State $\bank \gets \bank_0$;\quad $\theta \gets \theta_0$
\Repeat
  \Statex \quad\textbf{// Inner block: hybrid optimization at fixed $M$}
  \Repeat
    \State $\theta \gets \theta - \eta\, \nabla_\theta \loss(\theta, \bank)$ \Comment{SGD on Continuous Routing Weights}
    \State $\bank \gets \textsc{TextGradStep}(\bank, \theta, \Dcal)$ \Comment{TextGrad on Symbolic Bank}
  \Until{$\loss_{\text{val}}$ plateaus}
  \Statex \quad\textbf{// Outer block: bank size revision}
  \State $\bank \gets \textsc{Add}(\bank, \Dcal, \Dcal_{\text{val}})$ \Comment{Alg.~\ref{alg:fault_pretrain}}
  \State $\bank \gets \textsc{Drop}(\bank, \Dcal_{\text{val}})$ \Comment{Alg.~\ref{alg:drop}}
\Until{$\bank$ unchanged by both \textsc{Add} and \textsc{Drop}}
\State \Return $(\theta, \bank)$
\end{algorithmic}
\end{algorithm}

\subsection{Two-Block Optimization for EPB}
\label{sec:two_block}

Training EPB requires resolving three quantities: the programs in $\bank$, the router parameters $\theta$, and the bank size $M$. They live in different spaces, as $\theta$ is continuous, $\bank$ is symbolic, and $M$ is discrete, so no single optimizer handles all three. We therefore update them with different mechanisms in a nested two-block scheme (Algorithm~\ref{alg:two_block}). The inner \textbf{Block~I} fixes $M$ and jointly refines the programs $\bank$ and router $\theta$ through Hybrid Textual-Numerical Gradient Descent (\S\ref{sec:hybrid_gradient}), while letting an LLM diagnose the programs’ failures and rewrite them. The outer \textbf{Block~II} adjusts $M$ by adding programs for unresolved failure modes and dropping redundant ones (\S\ref{sec:bank_evolution}), after which we repeat 
Block~I at the new size, continuing the loop until optimization convergence.

\subsubsection{Block I: Hybrid Textual-Numerical Gradient Descent}
\label{sec:hybrid_gradient}
Block~I jointly trains the continuous routing parameters $\theta$ and the symbolic bank $\bank$ under a shared distillation loss $\loss(\theta,\bank)$: $\theta$ is updated by SGD, while $\bank$ is updated through TextGrad proposed in~\citep{yuksekgonul2024textgrad}. TextGrad uses LLM calls to generate feedback from the loss and then revise the code and its natural language descriptions in $\bank$ based on that feedback.

Specifically, for a decision state $s$, we define the per-state distillation loss and its expectation as
\begin{equation}
\loss(\theta,\bank)
  =
  \mathbb{E}_{s\sim\Dcal}
  \left[
    \ell_s(\theta,\bank)
  \right],
  \qquad  
  \ell_s(\theta,\bank)
  =
  \KLL\!\left(
    \teacher(\cdot\mid s)
    \,\|\,
    \student(\cdot\mid s;\theta,\bank)
  \right)
  .
  \label{eq:shared_distillation_loss}
\end{equation}

Programs are code and cannot be updated by numerical gradients, but gradients with respect to their outputs can help guide textual feedback. Each program output $V_m(s)$ affects the loss through two paths: directly, as a component of the mixture in Eq.~\ref{eq:mixture}, and indirectly, through its routing key $K_m$ in Eq.~\ref{eq:K_dynamic}. To isolate the direct contribution, we use $\operatorname{sg}(V)$ in the router, so $\partial w_j/\partial V_m = 0$ for all $j,m$. The router still reads $V_m(s)$ in the forward pass, but no gradient flows from the routing weights back to the program outputs (\S\ref{sec: stop_gradient_explaination}). The resulting gradients with respect to the program outputs and router parameters are given below, with derivations in Appendix~\S\ref{app:hybrid_gradient_derivation}:
\begin{equation}
  \nabla_{V_m(s)} \ell_s
  =
  w_m(s;\theta,\operatorname{sg}(V))\,r_s,
  \qquad
  \nabla_\theta \ell_s
  =
  \sum_{m=1}^{M}
  \left\langle r_s,V_m(s)\right\rangle
  \nabla_\theta w_m(s;\theta,\operatorname{sg}(V)),
  \label{eq:gradient_split_main}
\end{equation}
where $r_s = \nabla_{\student(\cdot\mid s)} \ell_s(\theta,\bank)$ is the gradient of the loss with respect to the student's output distribution. Thus, the distillation loss induces two coupled signals: an output-space sensitivity for attributing failures to program $m$, and a standard numerical gradient for updating the routing network $\theta$.

\textbf{Numerical signal driving textual updates.}
A standalone TextGrad backward call derives a textual gradient for a variable from that variable's own failures. In EPB, however, the $M$ programs jointly produce each prediction through the mixture, so a failure belongs to the mixture rather than to any single program, and we must first decide which programs to revise. The program-output gradient in Eq.~\ref{eq:gradient_split_main}, i.e., $\nabla_{V_m(s)} \ell_s = w_m(s)\, r_s$, is a natural starting point {but is insufficient on its own: since the mixture error $r_s$ is shared across the bank, the gradient differs across programs only through the routing weight. It thus tells us how much a program contributes to the mixture, but not whether the program itself is wrong. We therefore keep the routing weight and multiply it by a program-specific error term, measuring how far the program's output deviates from the teacher's with $\KLL\bigl(\teacher(\cdot\mid s)\,\|\,V_m(s)\bigr)$. Since our teachers are near-deterministic and fidelity is measured by top-action agreement, we count this deviation only when the program's top action differs from the teacher's, $\nstar(s)=\arg\max_{a\in\Acal(s)}\teacher(a\mid s)$.} This yields the feedback allocation score
\begin{equation}
\mathrm{score}_m(s)
  =
  \underbrace{w_m(s)}_{\text{contribution}}
  \cdot
  \underbrace{\indic{\arg\max_{a\in\Acal(s)} V_m(s,a) \neq \nstar(s)}}_{\text{wrong top action}}
  \cdot
  \underbrace{\KLL\bigl(\teacher(\cdot\mid s)\,\|\,V_m(s)\bigr)}_{\text{deviation from teacher}} .
  \label{eq:score}
\end{equation}
For each program, we rank states by $\mathrm{score}_m(s)$ and pass its top-$K$ states, $\mathcal F_m^t = \operatorname{TopK}_{s}\, \mathrm{score}^t_m(s)$, to TextGrad, which converts these failures into natural-language update directions for rewriting $c_m$ and $d_m$. Restricting feedback to the most relevant failures keeps the LLM calls manageable.

\textbf{Textual outputs driving numerical updates.}
The coupling also runs in reverse. In Eq.~\ref{eq:gradient_split_main}, the gradient for $\theta$ depends on the program outputs through $\langle r_s, V_m(s)\rangle$, so a TextGrad rewrite that changes a program's output distribution also changes the subsequent SGD gradient that trains the router.

\textbf{Closed-loop textual-numerical optimization.}
The two coupling directions close the training loop under the shared loss $\loss$, yielding an approximate \emph{block coordinate descent} process:
\begin{equation}
\begin{aligned}
  \theta^{t+1} &= \theta^t - \eta_t \nabla_\theta \loss(\theta^t,\bank^t), \quad
  \mathcal F_m^t = \operatorname{TopK} \mathrm{score}^t_m(s), \\
  \Delta_m^{\mathrm{text},t} &= \operatorname{TG\text{-}Backward}\!\left(c_m^t, d_m^t, \mathcal F_m^t\right), \quad
  (c_m^{t+1}, d_m^{t+1}) = \operatorname{TG\text{-}Step}\!\left(c_m^t, d_m^t, \Delta_m^{\mathrm{text},t}\right).
\end{aligned}
\label{eq:hybrid_bcd}
\end{equation}
Here $\operatorname{TG\text{-}Backward}$ uses an LLM to turn the selected failures $\mathcal F_m^t$ into a critique $\Delta_m^{\mathrm{text},t}$ of $(c_m^t, d_m^t)$, which $\operatorname{TG\text{-}Step}$ then applies as a rewrite. The continuous block $\theta$ is updated by SGD at every batch, while the symbolic block $B$ is updated periodically. This asymmetric schedule reflects the different costs: SGD performs small local steps, whereas each TextGrad step rewrites executable programs and therefore corresponds to a larger, non-differentiable move in function space.

\subsubsection{Block II: Bank Size Evolution via Add and Drop}

\label{sec:bank_evolution}
During training, we adjust the bank size $M$ by adding or removing programs. A static bank is ill-suited because optimization faces two competing pressures: previously unseen failure modes may emerge that no existing heuristic can resolve (requiring expansion), while refining different heuristics on similar teacher samples can make initially distinct programs behave similarly, leaving the bank larger than necessary. We thus introduce \textsc{Add}, which spawns new specialists for uncovered failure modes, and \textsc{Drop}, which prunes redundant or inactive heuristics. Both are triggered upon a validation plateau, a regime in which the validation loss changes only marginally.

\textbf{Failure-mode-targeted spawning (Add).} Add searches for failure cases that suggest missing programs: decision steps where \emph{every} program fails, which are unlikely to be fixed by re-routing alone. We pass the fault set to the LLM, which groups it into $K$ hypothesized failure modes, each with a candidate program. We then refine candidates on the fault subset over several rounds. Each fault state is assigned to the candidate currently closest to the teacher. Each candidate receives TextGrad feedback from its assigned states, and we accept edits only if they improve agreement with the teacher on a held-out split. Surviving candidates are added to the bank (Algorithm~\ref{alg:fault_pretrain}, \S\ref{sec:add_alg}).

\textbf{Redundancy-targeted removal (Drop).} Drop uses greedy leave-one-out pruning. It evaluates each removal, drops the program with the smallest increase in validation loss if this increase is within a preset tolerance, and recomputes. Pruning stops when no removal meets the tolerance, the per-call drop limit is reached, or the bank reaches its minimum size. This keeps the portfolio compact by eliminating programs whose contribution is covered by the rest of the bank (Algorithm~\ref{alg:drop}, \S\ref{sec:add_alg}).
\section{Experiments}
\label{sec:results}

\begin{wrapfigure}[12]{r}{0.38\textwidth}
  \centering  
  \vspace{-4pt}
  \includegraphics[width=0.92\linewidth]{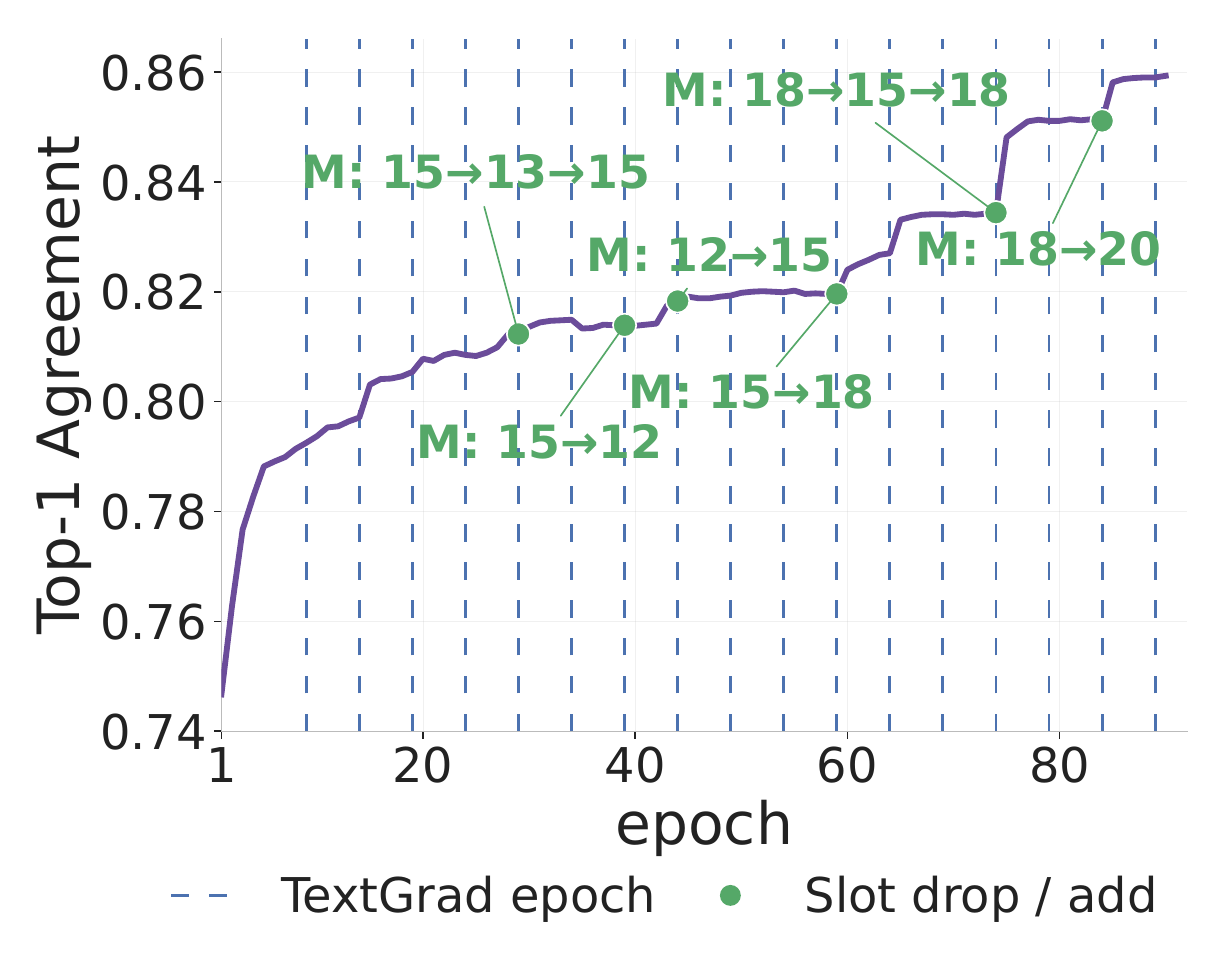}
  \vspace{-4pt}
  \caption{Training top-1 agreement.}
  \label{fig:training}
\end{wrapfigure}

We evaluate EPB by distilling POMO (TSP-50/100, CVRP-50/100) and LEHD (TSP-100, CVRP-100). Detailed experimental settings, additional results and discussions such as LLM costs, sensitivity to random seeds, and the distilled programs are provided in Appendices \S\ref{sec: setting} and \S\ref{sec:appendixresults}.
Fig.~\ref{fig:training} shows an example training curve for POMO-CVRP-50, where top-1 agreement measures teacher–student agreement given the same state. Hybrid gradient descent and bank size management jointly refine the program bank, improve the student surrogate's performance, and help escape plateaus. 

\subsection{Surrogate Performance}
\label{sec:fidelity}

\begin{table}[t]
\centering\footnotesize
\caption{Fidelity comparison of the surrogate to the teacher model. $N$: number of nodes.}
\vspace{10pt}
\setlength{\tabcolsep}{3.5pt}
\label{tab:teacher_vs_student}
\begin{tabular}{@{}l@{\hspace{4pt}}r*{6}{c}@{\hspace{6pt}}*{6}{c}@{}}
\toprule
& & \multicolumn{6}{c}{\textbf{TSP}} & \multicolumn{6}{c}{\textbf{CVRP}} \\
\cmidrule(lr){3-8}\cmidrule(l){9-14}
\textbf{Teacher} & $N$
& GPT-6 & Fable 5.1 & Tree & NAM & Mo\"ET & EPB
& GPT-6 & Fable 5.1 & Tree & NAM & Mo\"ET & EPB \\
\midrule
\multicolumn{14}{@{}l}{\textit{Top-1 agreement (\%)} $\uparrow$}\\
POMO & 50  & 85.9 & 84.7 & 85.6 & \underline{87.2} & 83.5 & \textbf{94.8} & 68.6 & 66.7 & 70.8 & 69.7 & \underline{71.6} & \textbf{85.8} \\
POMO & 100 & 84.2 & 84.6 & 84.7 & \underline{86.5} & 83.4 & \textbf{94.1} & 67.8 & 66.2 & 69.2 & 68.6 & \underline{71.0} & \textbf{86.5} \\
LEHD & 100 & 84.6 & 85.2 & 85.4 & \underline{86.9} & 83.1 & \textbf{94.7} & 64.8 & 63.5 & 66.9 & \underline{68.7} & 68.0 & \textbf{85.1} \\
\midrule
\multicolumn{14}{@{}l}{\textit{Greedy gap (\%)} $\downarrow$}\\
POMO & 50  & 9.65  & \underline{8.09} & 16.50 & 11.16 & 18.36 & \textbf{3.02} & 14.90 & \underline{13.23} & 25.95 & 17.89 & 14.16 & \textbf{2.88} \\
POMO & 100 & 10.98 & \underline{9.04} & 20.05 & 13.69 & 20.82 & \textbf{3.79} & 16.37 & \underline{14.43} & 28.63 & 19.38 & 15.62 & \textbf{3.94} \\
LEHD & 100 & 11.47 & \underline{9.53} & 20.08 & 13.75 & 22.34 & \textbf{2.57} & 15.89 & \underline{14.09} & 21.90 & 18.20 & 15.42 & \textbf{3.53} \\
\midrule
\multicolumn{14}{@{}l}{\textit{Search gap (\%)} $\downarrow$}\\
POMO & 50  & 2.54 & \underline{2.45} & 4.59 & 2.61 & 7.08  & \textbf{0.03} & 10.35 & 9.27  & 8.60  & 9.88  & \underline{7.52}  & \textbf{1.13} \\
POMO & 100 & 4.58 & \underline{3.90} & 8.31 & 4.98 & 10.89 & \textbf{0.07} & 12.85 & 11.27 & 12.06 & 12.37 & \underline{9.12}  & \textbf{1.25} \\
LEHD & 100 & 2.09 & \underline{1.64} & 3.46 & 2.61 & 7.15  & \textbf{0.34} & 14.34 & 12.69 & 14.47 & 12.13 & \underline{11.53} & \textbf{2.16} \\
\bottomrule
\end{tabular}
\end{table}
Table~\ref{tab:teacher_vs_student} reports the fidelity of the students to the teacher, measured by top-1 action agreement on the teacher’s greedy trajectories, and solution quality, measured by cost gaps under greedy decoding and search (multi-start for POMO, random reconstruction for LEHD). We compare against two kinds of interpretable surrogates. The first kind is a single scoring program written and tuned by an LLM coding agent, using either GPT-6 or Fable 5.1. The second kind is based on a fixed set of hand-crafted features, and we compare with three methods in Table~\ref{tab:teacher_vs_student}: Tree means a CART decision tree; NAM \cite{agarwal2021neural} sums the outputs of per-feature neural networks; and Mo\"ET \cite{vasic2022moet} combines decision-tree experts through a learned, state-dependent softmax gate. EPB achieves the highest top-1 agreement in all six settings, substantially outperforming the strongest baseline. It also attains the lowest cost gaps under both greedy and search decoding. Notably, the margin becomes larger on CVRP, where capacity-dependent decisions involve conditional logic that feature-based surrogates struggle to capture. Absolute costs (Table~\ref{tab:absolute_costs}) and instance-level tour comparisons are depicted in \S\ref{sec:absolute_costs} and \S\ref{sec: touragreement}.

\subsection{Analysis on Distilled Programs}
\begin{table}[b]
\centering
\footnotesize
\setlength{\tabcolsep}{5pt}
\renewcommand{\arraystretch}{1.12}
\caption{High-level summary of distilled program banks learned by EPB.}
\vspace{8pt}
\label{tab: heu_stats}
\begin{tabularx}{\linewidth}{l X}
\toprule
\textbf{Bank} & \textbf{High-level behavior} \\
\midrule
POMO TSP &
\textbf{Priorities shift with tour progress.}
Programs use progress or remaining-city counts to emphasize peripheral and isolated cities early, then start-aware closure and multi-step lookahead. \\

LEHD TSP &
\textbf{Local geometry and tour progress jointly guide decisions.}
Programs balance short travel with insertion and closure costs, using remaining-city geometry and directions relative to the start-to-current heading. Closure receives greater weight as fewer cities remain. \\

POMO CVRP &
\textbf{Customer preferences adapt to available capacity.}
Programs combine proximity, route savings, and directional coherence with how well customer demand fits the remaining capacity, including whether to continue serving customers or return to the depot. \\

LEHD CVRP &
\textbf{Local geometry and vehicle load jointly guide decisions.}
Programs combine short moves and route savings with demand fit and angular ordering around the depot, using load to regulate depot returns. \\
\bottomrule
\end{tabularx}
\end{table}
Table~\ref{tab: heu_stats} summarizes the distilled surrogates. POMO-TSP programs shift priorities across construction phases, while LEHD-TSP programs increasingly emphasize tour closure alongside local geometry (\S\ref{sec:pomo_lehd}). Both CVRP banks coordinate spatial preferences, demand, and depot returns based on remaining capacity. Next, we analyze their decision patterns both at individual steps and in aggregate across instances.
The full gallery and sample code are provided in \S\ref{sec: heu_gallery} and \S\ref{sec: appendix_code}, respectively.

\label{sec: insights}
\begin{wrapfigure}{r}{0.6\textwidth}
  \centering
  \vspace{2pt}
\includegraphics[width=\linewidth]{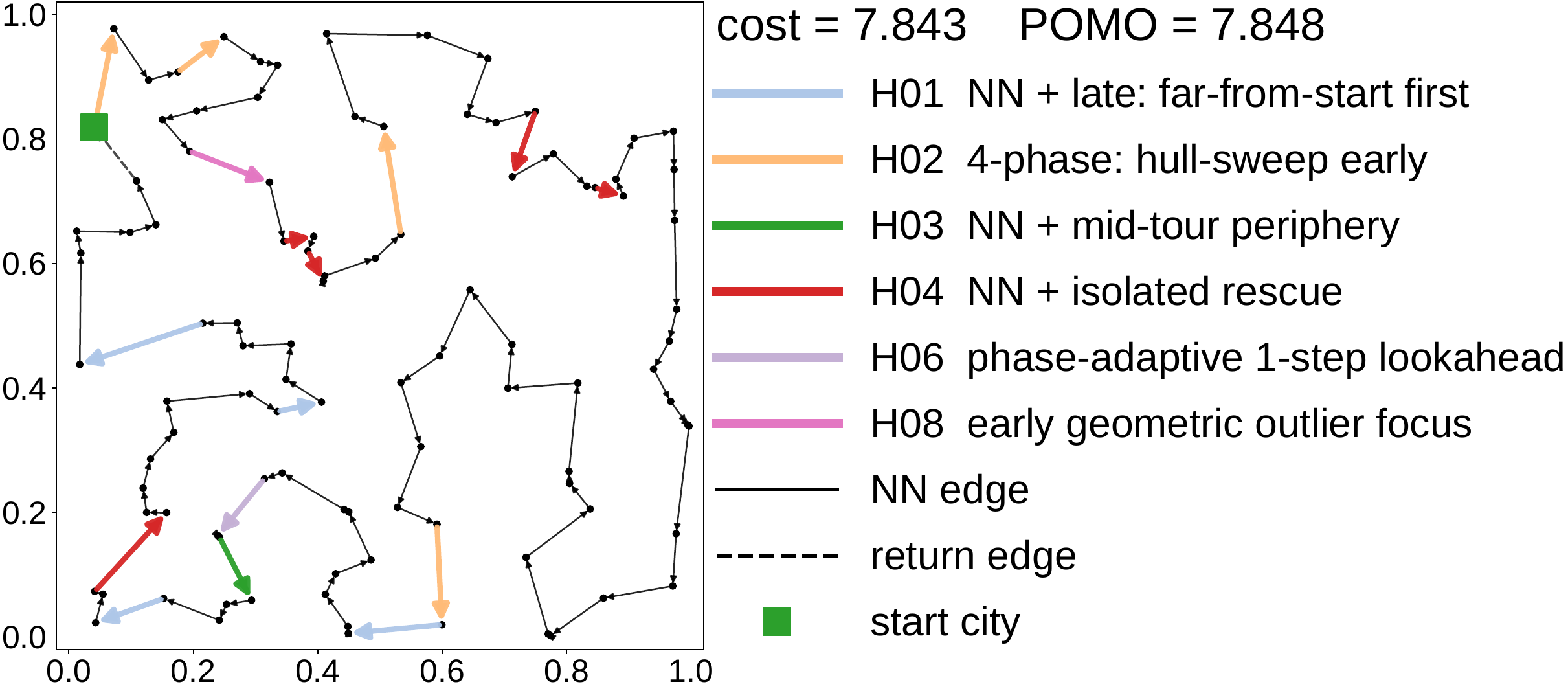}
  \caption{POMO-TSP-100 case study. The highlighted heuristic has the highest router weight at that step.}
  \label{fig:tsp_100_demo}
\end{wrapfigure}
\textbf{Interpreting individual decision patterns.} EPB explains individual decision steps by linking action choices to human-readable programs. Figure~\ref{fig:tsp_100_demo} highlights the steps where the POMO-TSP-100 student does not choose the nearest unvisited city. We focus on these steps because the nearest neighbor (NN) rule matches approximately 83\% of local decisions in the analyzed high-quality tours. At the highlighted steps, the highest-weight programs express distinct geometric preferences. \textsc{H02} (\emph{hull-sweep early}) rewards candidates farther from the centroid of unvisited cities while penalizing travel distance, favoring peripheral cities when the extra travel is justified. \textsc{H04} (\emph{isolated rescue}) gives extra preference to isolated cities that might otherwise require costly later detours. \textsc{H01} (\emph{far-from-start first}) balances short moves with a preference for cities farther from the starting city, leaving nearby cities available for tour closure. Together, these cases illustrate how the student balances immediate proximity with visiting peripheral or isolated cities and preparing for tour closure. In CVRP, the distilled rules explain why the surrogate bypasses the nearest customer to serve customers far from the depot while they are nearby or to reduce detours on the return trip (Figure~\ref{fig:pomo_cvrp_50_demo}). More examples are provided in~\S\ref{sec: more_inter_examaples}. 

\begin{figure*}[h]
  \centering
\includegraphics[width=\linewidth]{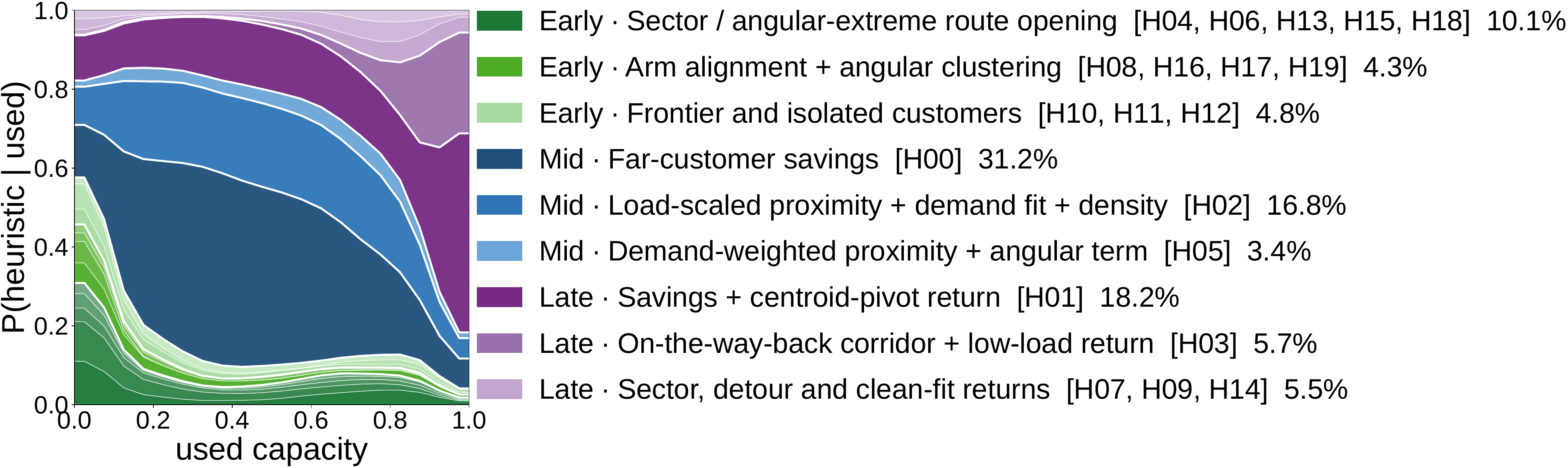}
  \caption{CVRP program activation across route progress (200 instances)}
  \label{fig:cvrp_heatmap}
\end{figure*}

\textbf{Interpreting aggregated decision patterns.} We aggregate decision patterns across 200 instances to reveal how strategies shift across route stages (Figure~\ref{fig:cvrp_heatmap}). It shows three phases: routes open around spatial structure, grow through savings and demand fit, and close toward the depot. Early decisions organize routes around angular sectors or prioritize peripheral and isolated customers. Midway, Clarke-Wright savings rules favor customers far from the depot but close to the vehicle, while demand-fit rules favor customers that leave little capacity unused. Near completion, depot-aware rules favor customers along the return corridor, discourage detours, and determine when to return. This suggests that the student balances vehicle utilization against the cost of returning to the depot.

\textbf{Faithfulness of the EPB surrogate.} We further test whether EPB’s explanations are causally faithful to its decisions using counterfactual and program-removal tests. Specifically, if a concept causally drives a decision, weakening it should weaken the student’s preference. For example, if the student prefers a city because it is isolated, making the city less isolated should reduce that preference. We perturb states to weaken a target city’s peripherality, isolation, or proximity to the homeward corridor, compare against matched controls, and track the effect from the programs that encode the concept, through the router, to the student’s preference for the target. We also measure the teacher’s response on the same perturbed states (Table~\ref{tab:counterfactual}) for reference. For all three concepts, the intervention reduces the relevant programs’ contribution and the student’s preference for the target, and the teacher’s preference shifts in the same direction on average. These results support the causal role of the tested concepts in student decisions and indicate that the explanations partially reflect the teacher’s behavior. Details are in \S\ref{app:counterfactual}. For a fourth concept, local demand density, the teacher also responds in the predicted direction on average, though more weakly than the student (\S\ref{app:cf-sweep}). 

Program-removal tests further show that removing the program with the highest routing weight usually changes the student’s action, whereas removing a randomly chosen program rarely does, confirming that dominant programs causally drive the student’s decisions (\S\ref{app: program_att}). 

\begin{table}[t]
\centering
\footnotesize
\caption{{Counterfactual faithfulness tests. We report the mean effect $g$ (control minus treatment, $g>0$ is the predicted direction) with 95\% instance-level bootstrap CIs, and the fraction of samples with $g>0$ in parentheses. $^\dagger$Manipulation check only.}}
\vspace{8pt}
\label{tab:counterfactual}
\setlength{\tabcolsep}{4pt}
\begin{tabular}{lccc}
\toprule
\textbf{Channel} & \textbf{POMO-TSP hull} & \textbf{LEHD-TSP isolation} & \textbf{POMO-CVRP corridor} \\
\midrule
Carrier score (Eq.~\ref{eq:E_S})$^\dagger$ & $+0.38$ [0.34, 0.43] (100\%) & $+1.20$ [1.18, 1.23] (100\%) & $+0.92$ [0.90, 0.94] (100\%) \\
Routing log-weight (Eq.~\ref{eq:R_S})     & $+1.54$ [1.08, 1.99] (73\%)  & $+0.29$ [0.20, 0.37] (72\%)  & $+0.39$ [0.32, 0.46] (61\%)  \\
Program contribution (Eq.~\ref{eq:C_S})    & $+0.38$ [0.33, 0.43] (92\%)  & $+0.08$ [0.06, 0.09] (74\%)  & $+0.25$ [0.24, 0.26] (91\%)  \\
Student log-odds (Eq.~\ref{eq:L_S})       & $+1.09$ [0.80, 1.38] (83\%)  & $+0.54$ [0.45, 0.63] (71\%)  & $+3.14$ [3.00, 3.28] (87\%)  \\
Teacher log-odds (Eq.~\ref{eq:L_T})       & $+9.57$ [7.69, 11.37] (84\%) & $+1.47$ [1.24, 1.74] (76\%)  & $+9.79$ [9.29, 10.29] (83\%) \\
\bottomrule
\end{tabular}
\end{table}

\textbf{Trade-off between code complexity and solution quality.} The complexity of the program affects interpretability.
Simpler programs are easier to interpret but may sacrifice fidelity. To study this trade-off, we run EPB on TSP-100 with per-program limits on the number of distinct feature primitives, executable statements, and branches (\S\ref{app:bank_readability}). In the simplest setting, each program uses at most two primitives. We then prune each bank one program at a time and keep the size with the lowest search gap, which gives 8 programs for POMO and 10 for LEHD. Including shared primitive implementations, the total statement count of the bank falls from 537 to 51 for POMO and from 419 to 49 for LEHD. Despite this reduction, solution cost gaps to the teachers remain small, rising from 0.03\% to 0.58\% for POMO and from 0.17\% to 0.33\% for LEHD. They are also well below those of the LLM coding-agent baselines, which use a similar amount of code. This shows that EPB can retain near-teacher solution quality with simple program code, facilitating a more interpretable surrogate.

\subsection{Generalization and Improvement}

\begin{table}[t]
\centering
\caption{Out-of-distribution generalization results on unseen problem sizes.}
\footnotesize
\vspace{5pt}
\label{tab:teacher_vs_student_generalization}
\setlength{\tabcolsep}{10pt}
\begin{tabular}{lcccccc}
\toprule
& \multicolumn{3}{c}{\textbf{Greedy}} & \multicolumn{3}{c}{\textbf{Search}} \\
\cmidrule(lr){2-4}\cmidrule(lr){5-7}
\textbf{Experiment} & \textbf{Teacher} & \textbf{EPB} & \textbf{Gap (\%)} & \textbf{Teacher} & \textbf{EPB} & \textbf{Gap (\%)} \\
\midrule
POMO-TSP-500  & 22.190 & 18.954 & $-14.58$ & 20.479 & 17.762 & $-13.27$ \\
POMO-CVRP-500 & 50.103 & 43.683 & $-12.81$ & 48.124 & 41.042 & $-14.72$ \\
LEHD-TSP-50   & 5.761  & 5.919  & $+2.74$  & 5.728  & 5.729  & $+0.03$ \\
LEHD-TSP-500  & 16.807 & 17.633 & $+4.91$  & 16.571 & 17.107 & $+3.24$ \\
\bottomrule
\end{tabular}
\end{table}

\textbf{Generalization to larger instances.} 
Table~\ref{tab:teacher_vs_student_generalization} reports zero-shot transfer to unseen sizes. EPB remains close to LEHD, which is designed to generalize across sizes. Notably, on both TSP and CVRP, the POMO surrogate significantly outperforms its teacher on larger instances, suggesting that distillation may yield more generalizable, less overfitted decision rules, possibly because EPB always combines relatively size-agnostic programs rather than relying on embeddings fitted to the training size.

\begin{figure*}[b]
\centering

\begin{minipage}[t]{0.48\textwidth}
\vspace{0pt}
\centering
\footnotesize
\captionof{table}{Intervention results on EPB surrogates. Gaps (\%) are relative to LKH for TSP and HGS for CVRP. Entries report greedy / search.}
\label{tab:intervention}
\setlength{\tabcolsep}{4pt}
\begin{tabular}{@{}lcc@{}}
\toprule
\textbf{Policy} & \textbf{TSP-100} & \textbf{CVRP-50} \\
               & Diagonal band   & Diagonal band   \\
\midrule
Teacher
& 38.18 / 20.90
& 13.34 / 7.66 \\
EPB
& \underline{24.16} / \underline{8.70}
& 10.88 / 4.17 \\
Edited Teacher
& 36.85 / 15.40
& \underline{7.40} / \underline{4.06} \\
Edited EPB
& \textbf{7.01 / 1.78}
& \textbf{5.94 / 2.43} \\
\bottomrule
\end{tabular}
\end{minipage}
\hfill
\begin{minipage}[t]{0.48\textwidth}
\vspace{0pt}
\centering
\includegraphics[width=0.7\linewidth]{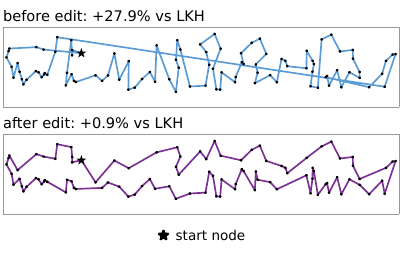}
\captionsetup{
    width=\textwidth,
    justification=justified,
    singlelinecheck=false
}
\vspace{-5pt}
\caption{EPB before and after editing on a TSP-100 diagonal-band instance.}
\label{fig:band_tsp_before_after}
\end{minipage}
\end{figure*}
\textbf{Intervention through program edits.}
The readable programs of EPB also let us diagnose and fix failures by directly editing code. We test this on out-of-distribution diagonal-band instances (nodes on a thin diagonal band). On TSP-100, all programs initially favor nearby nodes without reserving any for the return leg, leaving a long closing edge (Figure~\ref{fig:band_tsp_before_after}). We add distance penalties and three lane-keeping programs that go out along one side of the band and back along the other. As shown in Table~\ref{tab:intervention}, the edited EPB performs best. A black-box teacher, by contrast, cannot be edited directly. The most direct option is to add the same rule’s score as a bias to its log-probabilities, which helps but less than editing EPB’s programs. A similar edit also improves CVRP-50 (see Table~\ref{tab:intervention} and \S\ref{app:intervention}).

\subsection{Ablation Studies}
\label{sec:abl}

\noindent
\begin{minipage}[t]{0.53\linewidth}
\vspace{0pt}

\textbf{Ablation on hybrid gradients.}
Figure~\ref{fig:abl} (a) shows that numerical gradients alone (blue circles) yield limited gains. TextGrad (purple triangles) accelerates early training, but both saturate at fixed bank size.\\

\textbf{Ablation on bank size management.}
In Figure~\ref{fig:abl} (a), further activating bank size evolution (green squares) at epoch~25 produces a clear accuracy jump to break the plateau of fixed-size hybrid gradient descent. \\

\textbf{Ablation on evolving banks.}
Table~\ref{tab:bank_ablation} further shows that EPB achieves lower greedy and search gaps than two frozen baselines, which train only the router over a fixed bank initialized either by \emph{Expert} or by an \emph{LLM} (both released on GitHub). Expert’s larger gaps suggest that even a well-designed initial bank cannot match the teacher, so the programs should be evolved.\\

\textbf{Ablation on dynamic routing.} Figure~\ref{fig:abl}(b) shows that state-dependent dynamic keys (purple circles, \S\ref{sec:aacb}) outperform static keys (blue squares), with a clear gap already at epoch~1 despite identical initial program banks. A third variant, which builds keys from text embeddings of the program descriptions, performs similarly to static keys (\S\ref{app:key_design}), since both describe a program identically in every state, whereas dynamic keys reflect its behavior in each state.\\

\end{minipage}
\hfill
\begin{minipage}[t]{0.44\linewidth}
\vspace{0pt}
\centering
\includegraphics[width=0.95\linewidth]{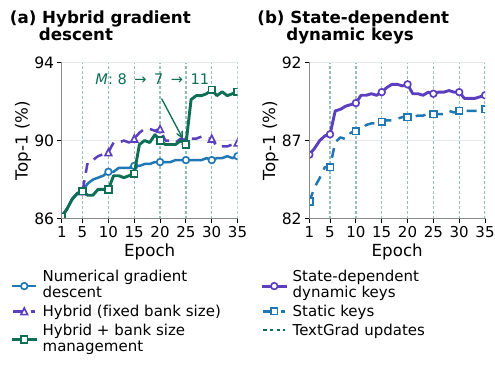}
\vspace{-3pt}
\captionof{figure}{Ablation studies.}
\label{fig:abl}

\vspace{-8pt}
\begin{table}[H]
\centering
\footnotesize
\caption{Cost gap (\%) to the teacher ($\downarrow$) with frozen banks (Expert- or LLM-initialized) versus evolving programmatic banks (EPB).}
\vspace{8pt}
\label{tab:bank_ablation}

\setlength{\tabcolsep}{3pt}
\renewcommand{\arraystretch}{1.0}
\resizebox{0.95\linewidth}{!}{%
\begin{tabular}{@{}l@{\hspace{4pt}}r*{3}{c}@{\hspace{6pt}}*{3}{c}@{}}
\toprule
& & \multicolumn{3}{c}{\textbf{TSP}} & \multicolumn{3}{c}{\textbf{CVRP}} \\
\cmidrule(lr){3-5}\cmidrule(l){6-8}
\textbf{Teacher} & $N$ & Expert & LLM & Ours & Expert & LLM & Ours \\
\midrule
\multicolumn{8}{@{}l}{\textit{Greedy gap (\%)} $\downarrow$}\\
POMO & 50  & \underline{13.76} & 16.26 & \textbf{3.02} & \underline{21.90} & 22.55 & \textbf{2.88} \\
POMO & 100 & 16.02 & \underline{12.45} & \textbf{3.79} & 23.27 & \underline{12.38} & \textbf{3.94} \\
LEHD & 100 & \underline{6.62} & 7.78 & \textbf{2.57} & 23.98 & \underline{7.40} & \textbf{3.53} \\
\midrule
\multicolumn{8}{@{}l}{\textit{Search gap (\%)} $\downarrow$}\\
POMO & 50  & 3.54 & \underline{3.53} & \textbf{0.03} & \underline{10.49} & 10.74 & \textbf{1.13} \\
POMO & 100 & 4.43 & \underline{2.65} & \textbf{0.07} & 12.34 & \underline{5.03} & \textbf{1.25} \\
LEHD & 100 & \underline{0.98} & 1.08 & \textbf{0.34} & 13.42 & \underline{5.23} & \textbf{2.16} \\
\bottomrule
\end{tabular}%
}
\end{table}
\end{minipage}
\par

\section{Conclusion}
\label{sec:conclusion}
We introduce Evolving Programmatic Bottlenecks (EPB), the first post-hoc interpretable surrogate framework that distills NCO policies into executable decision rules composed by a state-dependent router. Hybrid textual-numerical gradient descent refines the programs and router, while bank size management adapts the number of programs. Across POMO and LEHD on TSP and CVRP, EPB closely matches teacher behavior, and counterfactual and knockout tests support the causal faithfulness of its rules to the surrogate’s decisions. The surrogate also outperforms POMO on larger instances, and targeted program edits improve its out-of-distribution performance beyond the teacher.

While EPB takes a first step toward surrogate-based interpretability for NCO, it has several limitations. First, the router remains a neural network whose internal computation is not interpretable. A programmatic replacement may help but must retain sufficient expressiveness as suggested by our capacity ablation (\S\ref{sec: router_size}). Second, the failure-attribution score suits near-deterministic teachers, while stochastic teachers may require distribution-level measures. Third, EPB explains teacher behavior rather than internal mechanisms. Finally, discrete LLM rewrites can abruptly change a program’s behavior, destabilizing refinement. Future work could stabilize refinement and extend EPB, and more broadly the general idea of \emph{programmatic bottleneck models} to other sequential decision-making models, including more state-of-the-art NCO models, models with continuous action spaces, and other black-box solvers beyond neural networks, using interpretable surrogates to diagnose behavior patterns, guide policy improvement, and enable trustworthy deployment.

\begin{ack}
This research is supported by the National Research Foundation, Singapore under its AI Singapore AI Research Fundamental Research Collaborative (US-NSF Researcher Call) (AISG Award No: AISG3-RP-2025-036-USNSF). The authors would also like to thank Dr. Been Kim (Google DeepMind) and Prof. Bistra Dilkina (University of Southern California) for their helpful and insightful discussions during the early stages of this work.
\end{ack}

{
\small
\bibliographystyle{unsrtnat}
\bibliography{references}

@article{mankowitz_faster_2023,
	title = {Faster sorting algorithms discovered using deep reinforcement learning},
	volume = {618},
	pages = {257--263},
	number = {7964},
	journal = {Nature},
	author = {Mankowitz, Daniel J. and Michi, Andrea and Zhernov, Anton and Gelmi, Marco and Selvi, Marco and Paduraru, Cosmin and Leurent, Edouard and Iqbal, Shariq and Lespiau, Jean-Baptiste and Ahern, Alex and Köppe, Thomas and Millikin, Kevin and Gaffney, Stephen and Elster, Sophie and Broshear, Jackson and Gamble, Chris and Milan, Kieran and Tung, Robert and Hwang, Minjae and Cemgil, Taylan and Barekatain, Mohammadamin and Li, Yujia and Mandhane, Amol and Hubert, Thomas and Schrittwieser, Julian and Hassabis, Demis and Kohli, Pushmeet and Riedmiller, Martin and Vinyals, Oriol and Silver, David},
    year = {2023}
}

@inproceedings{koh2020concept,
  title={Concept bottleneck models},
  author={Koh, Pang Wei and Nguyen, Thao and Tang, Yew Siang and Mussmann, Stephen and Pierson, Emma and Kim, Been and Liang, Percy},
  booktitle={International conference on machine learning},
  pages={5338--5348},
  year={2020},
  organization={PMLR}
}

@article{oikarinen2023label,
  title={Label-free concept bottleneck models},
  author={Oikarinen, Tuomas and Das, Subhro and Nguyen, Lam M and Weng, Tsui-Wei},
  journal={arXiv preprint arXiv:2304.06129},
  year={2023}
}

@inproceedings{kool2019attention,
  title={Attention, Learn to Solve Routing Problems!},
  author={Kool, Wouter and Van Hoof, Herke and Welling, Max},
  booktitle={International Conference on Learning Representations},
  year={2019}
}

@article{zhang2025probing,
  title={Probing Neural Combinatorial Optimization Models},
  author={Zhang, Zhiqin and Ma, Yining and Cao, Zhiguang and Lau, Hoong Chuin},
  journal={arXiv preprint arXiv:2510.22131},
  year={2025}
}

@article{sun2023difusco,
  title={Difusco: Graph-based diffusion solvers for combinatorial optimization},
  author={Sun, Zhiqing and Yang, Yiming},
  journal={Advances in neural information processing systems},
  volume={36},
  pages={3706--3731},
  year={2023}
}

@article{luo2023neural,
  title={Neural combinatorial optimization with heavy decoder: Toward large scale generalization},
  author={Luo, Fu and Lin, Xi and Liu, Fei and Zhang, Qingfu and Wang, Zhenkun},
  journal={Advances in Neural Information Processing Systems},
  volume={36},
  pages={8845--8864},
  year={2023}
}

@inproceedings{kwon2020pomo,
  title={POMO: Policy Optimization with Multiple Optima for Reinforcement Learning},
  author={Kwon, Yeong-Dae and Choo, Jinho and Kim, Byoungjip and Yoon, Iljoo and Gwon, Youngjune and Min, Seungjai},
  booktitle={Advances in Neural Information Processing Systems},
  year={2020}
}

@article{yuksekgonul2022post,
  title={Post-hoc concept bottleneck models},
  author={Yuksekgonul, Mert and Wang, Maggie and Zou, James},
  journal={arXiv preprint arXiv:2205.15480},
  year={2022}
}

@article{yuksekgonul2024textgrad,
  title={Textgrad: Automatic" differentiation" via text},
  author={Yuksekgonul, Mert and Bianchi, Federico and Boen, Joseph and Liu, Sheng and Huang, Zhi and Guestrin, Carlos and Zou, James},
  journal={arXiv preprint arXiv:2406.07496},
  year={2024}
}

@article{li2022alphacode,
  title={Competition-level code generation with alphacode},
  author={Li, Yujia and Choi, David and Chung, Junyoung and Kushman, Nate and Schrittwieser, Julian and Leblond, R{\'e}mi and Eccles, Tom and Keeling, James and Gimeno, Felix and Dal Lago, Agustin and others},
  journal={Science},
  volume={378},
  number={6624},
  pages={1092--1097},
  year={2022},
  publisher={American Association for the Advancement of Science}
}

@article{romera2024funsearch,
  title={Mathematical discoveries from program search with large language models},
  author={Romera-Paredes, Bernardino and Barekatain, Mohammadamin and Novikov, Alexander and Balog, Matej and Kumar, M Pawan and Dupont, Emilien and Ruiz, Francisco JR and Ellenberg, Jordan S and Wang, Pengming and Fawzi, Omar and others},
  journal={Nature},
  volume={625},
  number={7995},
  pages={468--475},
  year={2024},
  publisher={Nature Publishing Group UK London}
}

@article{liu2024eoh,
  title={Evolution of heuristics: Towards efficient automatic algorithm design using large language model},
  author={Liu, Fei and Tong, Xialiang and Yuan, Mingxuan and Lin, Xi and Luo, Fu and Wang, Zhenkun and Lu, Zhichao and Zhang, Qingfu},
  journal={arXiv preprint arXiv:2401.02051},
  year={2024}
}

@article{shinn2023reflexion,
  title={Reflexion: Language agents with verbal reinforcement learning},
  author={Shinn, Noah and Cassano, Federico and Gopinath, Ashwin and Narasimhan, Karthik and Yao, Shunyu},
  journal={Advances in neural information processing systems},
  volume={36},
  pages={8634--8652},
  year={2023}
}

@article{yao2023tree,
  title={Tree of thoughts: Deliberate problem solving with large language models},
  author={Yao, Shunyu and Yu, Dian and Zhao, Jeffrey and Shafran, Izhak and Griffiths, Tom and Cao, Yuan and Narasimhan, Karthik},
  journal={Advances in neural information processing systems},
  volume={36},
  pages={11809--11822},
  year={2023}
}

@article{chen2021codex,
  title={Evaluating large language models trained on code},
  author={Chen, Mark and Tworek, Jerry and Jun, Heewoo and Yuan, Qiming and Pinto, Henrique Ponde De Oliveira and Kaplan, Jared and Edwards, Harri and Burda, Yuri and Joseph, Nicholas and Brockman, Greg and others},
  journal={arXiv preprint arXiv:2107.03374},
  year={2021}
}

@article{austin2021program,
  title={Program Synthesis with Large Language Models},
  author={Austin, Jacob and Odena, Augustus and Nye, Maxwell and Bosma, Maarten and Michalewski, Henryk and Dohan, David and Jiang, Ellen and Cai, Carrie and Terry, Michael and Le, Quoc and Sutton, Charles},
  journal={arXiv preprint arXiv:2108.07732},
  year={2021}
}

@article{bello2016neural,
  title={Neural combinatorial optimization with reinforcement learning},
  author={Bello, Irwan and Pham, Hieu and Le, Quoc V and Norouzi, Mohammad and Bengio, Samy},
  journal={arXiv preprint arXiv:1611.09940},
  year={2016}
}

@article{nazari2018reinforcement,
  title={Reinforcement learning for solving the vehicle routing problem},
  author={Nazari, Mohammadreza and Oroojlooy, Afshin and Snyder, Lawrence and Tak{\'a}c, Martin},
  journal={Advances in neural information processing systems},
  volume={31},
  year={2018}
}

@article{choo2022sgbs,
  title={Simulation-guided beam search for neural combinatorial optimization},
  author={Choo, Jinho and Kwon, Yeong-Dae and Kim, Jihoon and Jae, Jeongwoo and Hottung, Andr{\'e} and Tierney, Kevin and Gwon, Youngjune},
  journal={Advances in Neural Information Processing Systems},
  volume={35},
  pages={8760--8772},
  year={2022}
}

@inproceedings{hottung2022efficient,
  title={Efficient Active Search for Combinatorial Optimization Problems},
  author={Hottung, Andr{\'e} and Kwon, Yeong-Dae and Tierney, Kevin},
  booktitle={International Conference on Learning Representations},
  year={2022}
}

@inproceedings{ma2021learning,
  title={Learning to Iteratively Solve Routing Problems with Dual-Aspect Collaborative Transformer},
  author={Ma, Yining and Li, Jingwen and Cao, Zhiguang and Song, Wen and Zhang, Le and Chen, Zhenghua and Tang, Jing},
  booktitle={Advances in Neural Information Processing Systems},
  volume={34},
  pages={11096--11107},
  year={2021}
}

@inproceedings{sun2025cbllm,
  title={Concept Bottleneck Large Language Models},
  author={Sun, Chung-En and Oikarinen, Tuomas and Ustun, Berk and Weng, Tsui-Wei},
  booktitle={International Conference on Learning Representations},
  year={2025}
}

@inproceedings{kulkarni2025cbae,
  title={Interpretable Generative Models through Post-hoc Concept Bottlenecks},
  author={Kulkarni, Akshay and Yan, Ge and Sun, Chung-En and Oikarinen, Tuomas and Weng, Tsui-Wei},
  booktitle={Proceedings of the IEEE/CVF Conference on Computer Vision and Pattern Recognition},
  pages={8162--8171},
  year={2025}
}

@inproceedings{srivastava2024vlgcbm,
  title={VLG-CBM: Training Concept Bottleneck Models with Vision-Language Guidance},
  author={Srivastava, Divyansh and Yan, Ge and Weng, Tsui-Wei},
  booktitle={Advances in Neural Information Processing Systems},
  year={2024}
}

@inproceedings{kim2018tcav,
  title={Interpretability Beyond Feature Attribution: Quantitative Testing with Concept Activation Vectors (TCAV)},
  author={Kim, Been and Wattenberg, Martin and Gilmer, Justin and Cai, Carrie J. and Wexler, James and Viegas, Fernanda and Sayres, Rory},
  booktitle={International Conference on Machine Learning},
  year={2018}
}

@inproceedings{ghorbani2019ace,
  title={Towards Automatic Concept-based Explanations},
  author={Ghorbani, Amirata and Wexler, James and Zou, James Y. and Kim, Been},
  booktitle={Advances in Neural Information Processing Systems},
  year={2019}
}

@inproceedings{vinyals2015pointer,
  title={Pointer Networks},
  author={Vinyals, Oriol and Fortunato, Meire and Jaitly, Navdeep},
  booktitle={Advances in Neural Information Processing Systems},
  year={2015}
}

@article{wu2021learning,
  title={Learning Improvement Heuristics for Solving Routing Problems},
  author={Wu, Yaoxin and Song, Wen and Cao, Zhiguang and Zhang, Jie and Lim, Andrew},
  journal={IEEE Transactions on Neural Networks and Learning Systems},
  year={2021}
}

@article{michaud2024mips,
  title={Opening the ai black box: program synthesis via mechanistic interpretability},
  author={Michaud, Eric J and Liao, Isaac and Lad, Vedang and Liu, Ziming and Mudide, Anish and Loughridge, Chloe and Guo, Zifan Carl and Kheirkhah, Tara Rezaei and Vukeli{\'c}, Mateja and Tegmark, Max},
  journal={arXiv preprint arXiv:2402.05110},
  year={2024}
}

@article{geiger2023causal,
  title={Causal Abstraction: A Theoretical Foundation for Mechanistic Interpretability},
  author={Geiger, Atticus and Ibeling, Duligur and Zur, Amir and Chaudhary, Maheep and Chauhan, Sonakshi and Huang, Jing and Arora, Aryaman and Wu, Zhengxuan and Goodman, Noah and Potts, Christopher and Icard, Thomas},
  journal={arXiv preprint arXiv:2301.04709},
  year={2023}
}

@inproceedings{verma2018programmatically,
  title={Programmatically Interpretable Reinforcement Learning},
  author={Verma, Abhinav and Murali, Vijay and Singh, Rishabh and Kohli, Pushmeet and Chaudhuri, Swarat},
  booktitle={International Conference on Machine Learning},
  year={2018}
}

@inproceedings{trivedi2021synthesis,
  title={Learning to Synthesize Programs as Interpretable and Generalizable Policies},
  author={Trivedi, Dweep and Zhang, Jesse and Sun, Shao-Hua and Lim, Joseph J.},
  booktitle={Advances in Neural Information Processing Systems},
  year={2021}
}

@inproceedings{delfosse2023nudge,
  title={Interpretable and Explainable Logical Policies via Neurally Guided Symbolic Abstraction},
  author={Delfosse, Quentin and Shindo, Hikaru and Dhami, Devendra and Kersting, Kristian},
  booktitle={Advances in Neural Information Processing Systems},
  year={2023}
}

@inproceedings{zeiler2014visualizing,
  title={Visualizing and understanding convolutional networks},
  author={Zeiler, Matthew D and Fergus, Rob},
  booktitle={European conference on computer vision},
  pages={818--833},
  year={2014},
  organization={Springer}
}

@article{steinmann2023learning,
  title={Learning to intervene on concept bottlenecks},
  author={Steinmann, David and Stammer, Wolfgang and Friedrich, Felix and Kersting, Kristian},
  journal={arXiv preprint arXiv:2308.13453},
  year={2023}
}

@inproceedings{wu2023discover,
  title={Discover and cure: Concept-aware mitigation of spurious correlation},
  author={Wu, Shirley and Yuksekgonul, Mert and Zhang, Linjun and Zou, James},
  booktitle={International Conference on Machine Learning},
  pages={37765--37786},
  year={2023},
  organization={PMLR}
}

@article{craven1995extracting,
  title={Extracting tree-structured representations of trained networks},
  author={Craven, Mark and Shavlik, Jude},
  journal={Advances in neural information processing systems},
  volume={8},
  year={1995}
}

@article{frosst2017distilling,
  title={Distilling a neural network into a soft decision tree},
  author={Frosst, Nicholas and Hinton, Geoffrey},
  journal={arXiv preprint arXiv:1711.09784},
  year={2017}
}

@article{ghosh2023dividing,
  title={Dividing and conquering a blackbox to a mixture of interpretable models: route, interpret, repeat},
  author={Ghosh, Shantanu and Yu, Ke and Arabshahi, Forough and Batmanghelich, Kayhan},
  journal={Proceedings of machine learning research},
  volume={202},
  pages={11360},
  year={2023}
}

@article{moreira2024diconstruct,
  title={Diconstruct: Causal concept-based explanations through black-box distillation},
  author={Moreira, Ricardo and Bono, Jacopo and Cardoso, M{\'a}rio and Saleiro, Pedro and Figueiredo, M{\'a}rio AT and Bizarro, Pedro},
  journal={arXiv preprint arXiv:2401.08534},
  year={2024}
}

@article{feng2026bayesian,
  title={Bayesian concept bottleneck models with llm priors},
  author={Feng, Jean and Kothari, Avni and Zier, Lucas and Singh, Chandan and Tan, Yan Shuo},
  journal={Advances in Neural Information Processing Systems},
  volume={38},
  pages={93889--93920},
  year={2026}
}

@article{de2026mixture,
  title={Mixture of Concept Bottleneck Experts},
  author={De Santis, Francesco and Ciravegna, Gabriele and De Felice, Giovanni and Casanova, Arianna and Giannini, Francesco and Diligenti, Michelangelo and Schneider, Johannes and Giordano, Danilo and Zarlenga, Mateo Espinosa and Barbiero, Pietro},
  journal={arXiv preprint arXiv:2602.02886},
  year={2026}
}

@inproceedings{ye2025licorice,
  title={Licorice: Label-efficient concept-based interpretable reinforcement learning},
  author={Ye, Zhuorui and Milani, Stephanie and Gordon, Geoff and Fang, Fei},
  booktitle={International Conference on Learning Representations},
  volume={2025},
  pages={102754--102782},
  year={2025}
}

@article{ye2024reevo,
  title={Reevo: Large language models as hyper-heuristics with reflective evolution},
  author={Ye, Haoran and Wang, Jiarui and Cao, Zhiguang and Berto, Federico and Hua, Chuanbo and Kim, Haeyeon and Park, Jinkyoo and Song, Guojie},
  journal={Advances in neural information processing systems},
  volume={37},
  pages={43571--43608},
  year={2024}
}

@article{bastani2018verifiable,
  title={Verifiable reinforcement learning via policy extraction},
  author={Bastani, Osbert and Pu, Yewen and Solar-Lezama, Armando},
  journal={Advances in neural information processing systems},
  volume={31},
  year={2018}
}

@article{vasic2022moet,
  title={Mo{\"E}T: Mixture of expert trees and its application to verifiable reinforcement learning},
  author={Vasi{\'c}, Marko and Petrovi{\'c}, Andrija and Wang, Kaiyuan and Nikoli{\'c}, Mladen and Singh, Rishabh and Khurshid, Sarfraz},
  journal={Neural Networks},
  volume={151},
  pages={34--47},
  year={2022},
  publisher={Elsevier}
}

@article{agarwal2021neural,
  title={Neural additive models: Interpretable machine learning with neural nets},
  author={Agarwal, Rishabh and Melnick, Levi and Frosst, Nicholas and Zhang, Xuezhou and Lengerich, Ben and Caruana, Rich and Hinton, Geoffrey E},
  journal={Advances in neural information processing systems},
  volume={34},
  pages={4699--4711},
  year={2021}
}

@inproceedings{mall2025disciple,
  title={Disciple: Learning interpretable programs for scientific visual discovery},
  author={Mall, Utkarsh and Phoo, Cheng Perng and Chiquier, Mia and Hariharan, Bharath and Bala, Kavita and Vondrick, Carl},
  booktitle={2025 IEEE/CVF Conference on Computer Vision and Pattern Recognition (CVPR)},
  pages={29258--29267},
  year={2025},
  organization={IEEE}
}

@article{novikov2025alphaevolve,
  title={Alphaevolve: A coding agent for scientific and algorithmic discovery},
  author={Novikov, Alexander and V{\~u}, Ng{\^a}n and Eisenberger, Marvin and Dupont, Emilien and Huang, Po-Sen and Wagner, Adam Zsolt and Shirobokov, Sergey and Kozlovskii, Borislav and Ruiz, Francisco JR and Mehrabian, Abbas and others},
  journal={arXiv preprint arXiv:2506.13131},
  year={2025}
}

@inproceedings{lange2026shinkaevolve,
  title={Shinkaevolve: Towards open-ended and sample-efficient program evolution},
  author={Lange, Robert and Imajuku, Yuki and Cetin, Edoardo},
  booktitle={International Conference on Learning Representations},
  volume={2026},
  pages={74026--74078},
  year={2026}
}
}


\newpage
\appendix
\clearpage
\appendix

\vbox{
\hsize\textwidth
\hrule height 4pt
\vskip 0.25in%
\vskip -\parskip%
\centering
{\LARGE\bf Interpreting Neural Combinatorial Optimization via
Evolving Programmatic Bottlenecks (Appendix) \par}
\vskip 0.29in
\vskip -\parskip
\hrule height 1pt
\vskip 0.3in%
}

\startcontents[appendix]

\printcontents[appendix]{}{1}{\setcounter{tocdepth}{2}}

\clearpage
\section{Preliminary}
\subsection{TSP and CVRP}\label{sec: tsp_cvrp}
\paragraph{TSP}
The Traveling Salesman Problem (TSP) is defined on a set of $n$ nodes, where each node $i$ is associated with a coordinate $x_i \in \mathbb{R}^2$. A problem instance is denoted by $s = \{x_1, \dots, x_n\}$, typically sampled i.i.d. from a distribution over the plane (e.g., uniform in the unit square).
A feasible solution to TSP is a permutation $\pi = (\pi_1, \dots, \pi_n)$ of the nodes, representing a tour that visits each node exactly once and returns to the starting node. The total cost of a tour is defined as \[
L(\pi \mid s) = \sum_{t=1}^{n} d(x_{\pi_t}, x_{\pi_{t+1}}),
\] where $d(\cdot, \cdot)$ denotes the Euclidean distance between the two nodes and $\pi_{n+1} = \pi_1$.
The objective of TSP is to find a permutation $\pi^*$ that minimizes the total tour cost: $\pi^* = \arg\min_{\pi} L(\pi \mid s)$.
In neural combinatorial optimization settings, solutions are constructed by sequentially selecting the next node to visit conditioned on the current state:
$$p(\pi \mid s) = \prod_{t=1}^{n} p(\pi_t \mid s, \pi_{1:t-1}),$$ where previously visited nodes are excluded from the candidate set to ensure valid tours.
\paragraph{CVRP}The Capacitated Vehicle Routing Problem (CVRP) is defined on a set of $n$ customer nodes and a depot. Each customer node $i$ is associated with a location $x_i$ in $\mathbb{R}^2$ and a demand $d_i > 0$, and the depot has location $x_0$. Node locations and customer demands are drawn from a fixed distribution.  A feasible solution to CVRP consists of a set of routes, where each route starts and ends at the depot and visits a subset of customers. The total demand served in each route must not exceed the vehicle capacity $C$. The objective is to construct a set of routes that visits all customers while minimizing the total travel cost:
\[
L(\pi \mid s) = \sum_{\text{routes}}\sum_{t=1}^{|\pi|} d(x_{\pi_t}, x_{\pi_{t+1}}), 
\] where each route is represented as a sequence $\pi$ of visited nodes, and $\pi_{|\pi|+1}$ includes returning to the depot at the end of the route. In neural combinatorial optimization settings, solutions are generated by sequentially selecting the next node conditioned on the current partial route and the remaining vehicle capacity. Nodes who have already been served or whose demand exceeds the remaining capacity are masked to ensure feasibility.

\begin{figure}[ht]
  \centering
  \includegraphics[width=\textwidth]{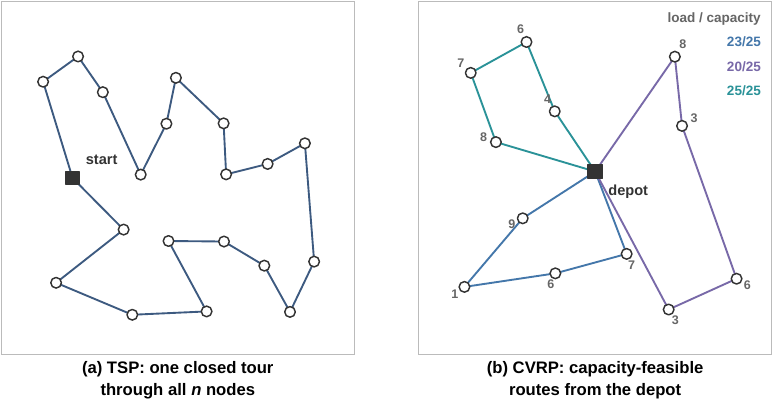}
  \caption{\textbf{(a)} TSP: a closed tour visiting each node once. \textbf{(b)} CVRP: routes start and end at the depot (square), each with total demand at most $C$. Numbers indicate customer demands.}
  \label{fig:tsp_cvrp}
\end{figure}

\subsection{Teacher/Student NCO model Structure}\label{sec: teacher_models}
We use POMO~\citep{kwon2020pomo} and LEHD~\citep{luo2023neural} as representative black-box NCO teachers. Both models define an autoregressive policy over feasible future actions, but they differ in architecture, training procedure, and inference strategy.
\paragraph{POMO} POMO is an attention-based reinforcement learning solver for combinatorial optimization. Its key idea is to harness the fact that many routing solutions have multiple equivalent solutions (evaluated by the relevant downstream metric). For example, a TSP solution cycle can be represented by starting from any city. POMO uses this symmetry by generating multiple rollout trajectories from different starting nodes for each problem instance during training and inference. This encourages the model to explore multiple equivalent optima rather than anchoring to a single starting point.

\paragraph{LEHD}
LEHD was proposed to improve generalization to large-scale problem instances. Earlier constructive NCO models often use a heavy-encoder/light-decoder structure, where node embeddings are computed once by the encoder and then reused throughout sequential decoding. This static embedding strategy can encourage the model to learn scale-dependent features, which may hurt performance when generalizing to larger instances. LEHD instead uses a lightweight encoder and a heavy decoder that dynamically captures relationships among the current partial solution and the available nodes at each construction step. LEHD is further trained with partial-solution augmentation. Given an optimal tour, it samples consecutive subsequences with random lengths and directions, and treats each subsequence as an augmented training example. The model then learns to reconstruct each partial solution step by step. At inference time, LEHD constructs full solutions greedily and can further improve them with Random Re-Construct, which repeatedly reconstructs sampled solution segments and accepts a replacement only if it improves the objective.

\paragraph{Student Model}
Similar to Concept Bottleneck Models, our student router is designed to follow the architecture of the corresponding teacher model while exposing an interpretable intermediate decision layer. For both TSP and CVRP, the goal is to obtain node embeddings that serve as the input representation for the router network. For POMO-based teachers, we train a POMO encoder to compute these embeddings. For LEHD-based teachers, we use the LEHD encoder together with few decoder layers to obtain the node embeddings.

\section{Methodology}
\subsection{Hierarchical Symbolic Optimization}\label{sec:hierarchical}
A heuristic can fail for two structurally different reasons. Its implementation may have a local bug, or its underlying strategy may be wrong for the failure mode it covers. This motivates a hierarchical strategy that updates the natural language description $d_m$ and the code implementation $c_m$ of each heuristic in a self-consistent way. A dispatcher selects which heuristics enter this update phase based on accumulated failure attribution scores from Section~\ref{sec:hybrid_gradient}. Phase 1 attempts local code refinement, and Phase 2 escalates to strategy-level revision only when Phase 1 cannot resolve the failure.
The TextGrad framework gives us two distinct channels per bank entry: backward calls on the code $c_m$, and backward calls on the natural-language description $d_m$.  We do not run the two channels simultaneously.  Instead, we organize them into a two-phase hierarchical procedure: \textbf{Phase 1} acts only on $c_m$, attempting local code refinement, and \textbf{Phase 2} acts only on $d_m$ (with $c_m$ subsequently re-implemented from the revised description), invoked only when Phase 1 cannot resolve the failure.

For each selected heuristic, Phase 1 attempts up to $R$ rounds of TextGrad on $c_m$. After each update, the candidate $c_m$ is evaluated on a held-out validation set. If the validation loss with the teacher distribution improves over the pre-substitution bank's by at least $\delta$, the candidate is retained and $d_m$ is regenerated from the new code to reflect the updated implementation.

If no update results from Phase 1, this indicates the failure is strategy-level rather than implementation-level. Phase 2 then updates $d_m$ via TextGrad on description-level gradients, and $c_m$ is re-implemented from the revised $d_m$. The new code is committed under the same validation criterion. If both phases fail to commit a new heuristic, the description is reverted and the heuristic is left unchanged. In this way, the bank maintains internally aligned $(c_m, d_m)$ pairs regardless of the sequence of changes.

\begin{tcolorbox}[
    colback=black!8!white,
    colframe=black,
    coltitle=white,
    title=\textbf{TextGrad backward output formats},
    fonttitle=\bfseries,
    fontupper=\small,
    boxrule=0.5pt,
    arc=2pt,
]
TextGrad's backward LLM call produces a structured natural-language
diagnostic in one of two formats, depending on the channel.

\medskip
\textbf{Code channel} (backward on $c_m$, four fields):
\begin{verbatim}
LOGIC:     [what the code currently computes for ranking 
            these top candidates]
TEACHER:   [what features of the teacher's pick the code 
            should recognize]
FLAW:      [why the code's score for the teacher's pick 
            fell behind the wrong argmax]
DIRECTION: [what change would push the teacher's pick 
            above the wrong argmax]
\end{verbatim}

\textbf{Description channel} (backward on $d_m$, three fields):
\begin{verbatim}
MISSING:    [what geometric reasoning the description 
             omits or under-emphasizes]
MISLEADING: [anything in the description that points 
             the implementation in the wrong direction]
DIRECTION:  [how the description should be revised]
\end{verbatim}
\end{tcolorbox}
\subsection{On the Necessity of the Stop-gradient}
\label{sec: stop_gradient_explaination}
The stop-gradient in Eq.~\ref{eq:K_dynamic} is a design choice rather than a requirement. The router parameters $\theta$ are continuous and updated by SGD, while the heuristic outputs $V_m(s)$ are produced by executable code and revised by TextGrad; $V_m(s)$ is never updated by gradients regardless of whether the path is stopped. Removing the stop-gradient would only add a second term to $\nabla_{V_m(s)}\ell_s$, flowing through the routing key $K_m$. Since this gradient is used not as an update but as an attribution signal for selecting which heuristic receives textual feedback (Eq.~\ref{eq:score}), stopping the routing path keeps the signal clean: it isolates $V_m(s)$'s direct effect on the mixture from its indirect effect on routing, so that the gradients on $\theta$ and on $V_m(s)$ no longer share a path and update their respective channels independently. The resulting gradient, $\nabla_{V_m(s)}\ell_s = w_m(s;\theta,\operatorname{sg}(V))r_s$, scales the shared mixture-level error $r_s$ by the program's routing weight. Since this gradient alone cannot determine whether a program is wrong, the attribution score in Eq.~\ref{eq:score} additionally incorporates each program's top-action disagreement and KL divergence from the teacher to allocate textual feedback.
\subsection{Routing Key Design}
\label{app:key_design}

\begin{table}[!htb]
\centering
\small
\caption{Distillation KL divergence for different routing keys with a fixed heuristic bank.}
\vspace{6pt}
\label{tab:key_design}
\setlength{\tabcolsep}{6pt}
\begin{tabular}{lrrrrrrr}
\toprule
& \multicolumn{7}{c}{\textbf{Training epoch}} \\
\cmidrule(lr){2-8}
\textbf{Key design} & 1 & 10 & 20 & 30 & 40 & 50 & 60 \\
\midrule
State-dependent Dynamic Keys
& \textbf{0.0252} & \textbf{0.0189} & \textbf{0.0176} & \textbf{0.0171} & \textbf{0.0167}
& \textbf{0.0166} & \textbf{0.0167} \\
Static Keys
& 0.0359 & 0.0242 & 0.0219 & 0.0203 & 0.0197
& 0.0192 & 0.0191 \\
Text Embedding Keys
& 0.0353 & 0.0231 & 0.0212 & 0.0203 & 0.0195
& 0.0193 & 0.0191 \\
\bottomrule
\end{tabular}
\end{table}

\paragraph{Setup.}
We compare EPB's dynamic keys with two static alternatives while keeping the heuristic bank fixed and training only the router. EPB constructs each key as the expectation of state-dependent node embeddings under the corresponding heuristic's action distribution. The \emph{static} variant assigns each heuristic a freely learnable embedding $e_m \in \mathbb{R}^{D}$ that is independent of the current state. The \emph{text} variant encodes each heuristic's natural-language description $d_m$ using the frozen OpenAI \texttt{text-embedding-3-large} model and projects the embedding to the shared key dimension:
\[
z_m = E_{\mathrm{text}}(d_m) \in \mathbb{R}^{3072},
\qquad
K_m^{\mathrm{text}} = W z_m,
\qquad
W \in \mathbb{R}^{D \times 3072}.
\]
The text embeddings remain fixed, while the projection is learned during router training. We report the distillation KL divergence to the teacher at selected training epochs.

\begin{wrapfigure}{r}{0.42\textwidth}
\centering
\includegraphics[width=\linewidth]{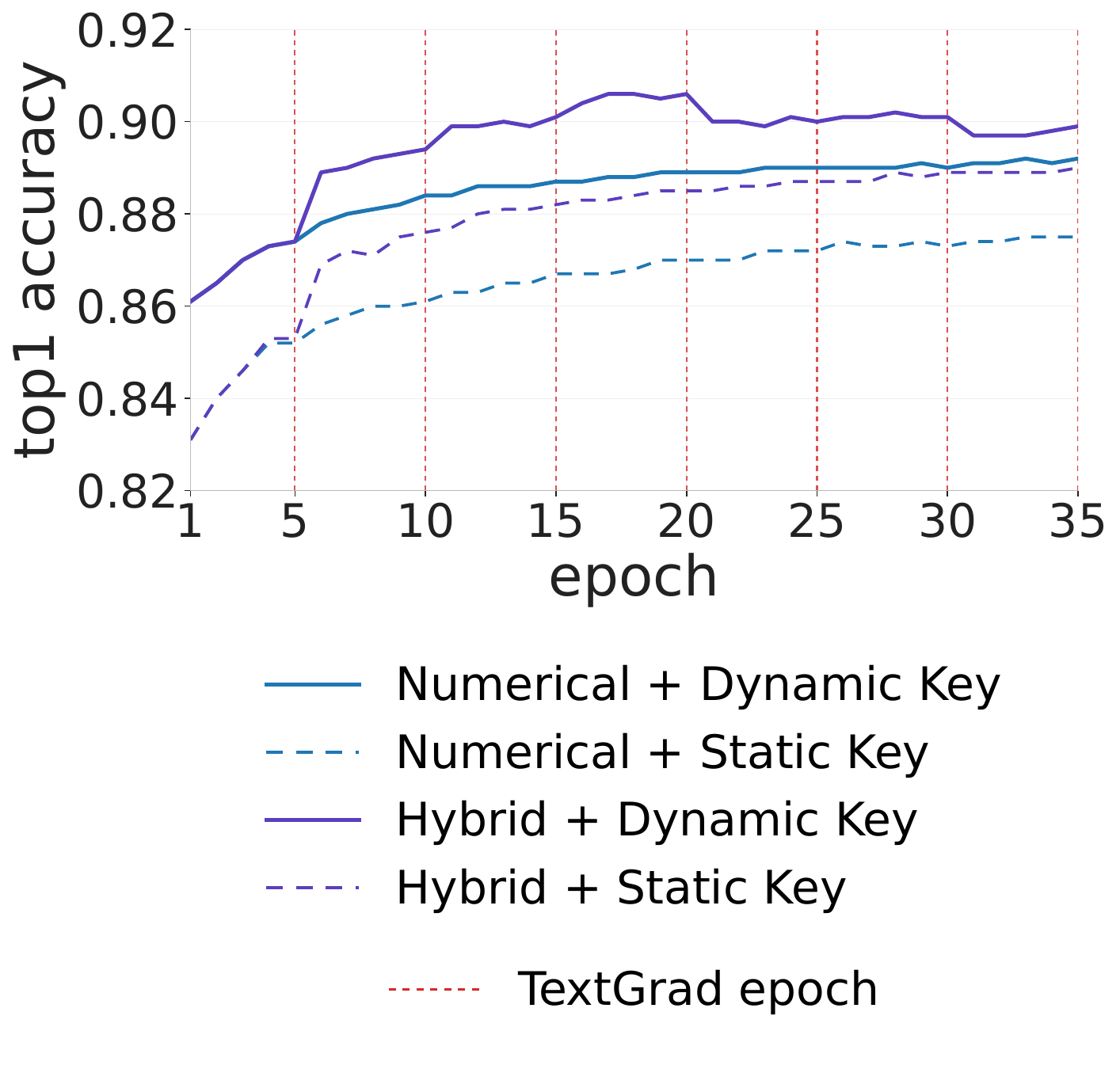}
\caption{Top-1 agreement with the teacher under numerical and hybrid training. Red vertical lines mark  epochs that perform TextGrad.}
\label{fig:key_top1}
\end{wrapfigure}

\paragraph{Results.}
EPB's dynamic keys achieve lower KL divergence at every reported epoch (Table~\ref{tab:key_design}). At epoch 60, both static alternatives reach 0.0191, compared with 0.0167 for EPB. The lowest reported value for EPB is 0.0166 at epoch 50. Text-derived keys closely track freely learned static keys and provide no improvement at the final reported epoch. Both static alternatives allow state-dependent routing through the learned query, but their keys do not reflect what each heuristic recommends in the current state.

Figure~\ref{fig:key_top1} compares dynamic and static keys under two training schemes. Numerical training updates only the router while keeping the heuristic bank fixed. Hybrid training also revises the programs through TextGrad. Dynamic keys yield higher top-1 agreement under both schemes, while hybrid training further improves agreement for either key design. The combination of hybrid training and dynamic keys achieves the highest agreement. These results support incorporating each heuristic's current action distribution into its routing key, both with a fixed bank and during program evolution.
\subsection{Derivation of the Hybrid Gradient Decomposition}
\label{app:hybrid_gradient_derivation}

We derive the gradient decomposition used in
\S\ref{sec:hybrid_gradient}. For a fixed decision state $s$, the student
policy is the routing-weighted mixture
\begin{equation}
  \student(\cdot\mid s)
  =
  \sum_{m=1}^{M}
  w_m(s;\theta,\operatorname{sg}(V))\,V_m(s).
  \label{eq:app_mixture}
\end{equation}
For readability, write
\[
  \bar V=\operatorname{sg}(V),
  \qquad
  w_m=w_m(s;\theta,\bar V),
  \qquad
  V_m=V_m(s).
\]
Then Eq.~\ref{eq:app_mixture} becomes
\begin{equation}
  \student(\cdot\mid s)
  =
  \sum_{m=1}^{M}
  w_m V_m.
  \label{eq:app_local_mixture}
\end{equation}

Consider small perturbations
\[
  V_m \mapsto V_m+\delta V_m,
  \qquad
  \theta \mapsto \theta+\delta\theta.
\]
The corresponding first-order perturbation of the student mixture is
\begin{equation}
\begin{aligned}
  \delta \student(\cdot\mid s)
  &=
  \sum_{m=1}^{M}
  \Big[
    (w_m+\delta w_m)(V_m+\delta V_m)-w_mV_m
  \Big]
  \\
  &=
  \sum_{m=1}^{M}
  \left(
    w_m\,\delta V_m
    +
    V_m\,\delta w_m
  \right)
  +
  O(\|\delta\|^2),
\end{aligned}
\label{eq:app_mixture_delta_raw}
\end{equation}
where the second-order term (i.e. 
$\sum_m \delta w_m\,\delta V_m$) is omitted.

The routing weights depend on $\theta$ and on the detached heuristic outputs
$\bar V=\operatorname{sg}(V)$, as in Eq.~\ref{eq:routing}. Hence their first-order perturbation is
\begin{equation}
  \delta w_m
  =
  \left\langle
    \nabla_\theta w_m(s;\theta,\bar V),
    \delta\theta
  \right\rangle
  +
  \left\langle
    \nabla_{\bar V} w_m(s;\theta,\bar V),
    \delta\bar V
  \right\rangle .
  \label{eq:app_delta_w_full}
\end{equation}
Since $\bar V=\operatorname{sg}(V)$, perturbations of $V$ do not propagate
through the router:
\begin{equation}
  \delta\bar V = 0,
  \qquad
  \frac{\partial w_j(s;\theta,\operatorname{sg}(V))}
       {\partial V_m(s)}
  =
  0  \qquad \forall j,m\in [M] .
  \label{eq:app_stop_gradient}
\end{equation}
Therefore,
\begin{equation}
  \delta w_m
  =
  \left\langle
    \nabla_\theta w_m(s;\theta,\operatorname{sg}(V)),
    \delta\theta
  \right\rangle .
  \label{eq:app_delta_w_theta_only}
\end{equation}

Substituting Eq.~\ref{eq:app_delta_w_theta_only} into
Eq.~\ref{eq:app_mixture_delta_raw} gives
\begin{equation}
  \delta \student(\cdot\mid s)
  =
  \sum_{m=1}^{M}
  w_m(s;\theta,\operatorname{sg}(V))\,\delta V_m(s)
  +
  \sum_{m=1}^{M}
  V_m(s)
  \left\langle
    \nabla_\theta w_m(s;\theta,\operatorname{sg}(V)),
    \delta\theta
  \right\rangle .
  \label{eq:app_mixture_delta_sg}
\end{equation}

Now define the mixture-level gradient
\begin{equation}
  r_s
  =
  \nabla_{\student(\cdot\mid s)}
  \ell_s(\theta,\bank).
  \label{eq:app_mixture_gradient}
\end{equation}
By the first-order Taylor expansion of the loss around the current student
output,
\begin{equation}
  \delta \ell_s
  \approx
  \left\langle
    r_s,
    \delta \student(\cdot\mid s)
  \right\rangle .
  \label{eq:app_loss_taylor}
\end{equation}
Combining Eq.~\ref{eq:app_mixture_delta_sg} and
Eq.~\ref{eq:app_loss_taylor}, we obtain
\begin{equation}
\begin{aligned}
  \delta \ell_s
  &\approx
  \sum_{m=1}^{M}
  \left\langle
    r_s,
    w_m(s;\theta,\operatorname{sg}(V))\,\delta V_m(s)
  \right\rangle
  \\
  &\quad+
  \sum_{m=1}^{M}
  \left\langle
    r_s,
    V_m(s)
  \right\rangle
  \left\langle
    \nabla_\theta w_m(s;\theta,\operatorname{sg}(V)),
    \delta\theta
  \right\rangle
  \\
  &=
  \sum_{m=1}^{M}
  \left\langle
    w_m(s;\theta,\operatorname{sg}(V))\,r_s,
    \delta V_m(s)
  \right\rangle
  \\
  &\quad+
  \left\langle
    \sum_{m=1}^{M}
    \left\langle r_s,V_m(s)\right\rangle
    \nabla_\theta w_m(s;\theta,\operatorname{sg}(V)),
    \delta\theta
  \right\rangle .
\end{aligned}
\label{eq:app_hybrid_gradient_decomposition}
\end{equation}

Since a first-order perturbation can also be written as
\[
  \delta \ell_s
  \approx
  \sum_{m=1}^{M}
  \left\langle
    \nabla_{V_m(s)}\ell_s,
    \delta V_m(s)
  \right\rangle
  +
  \left\langle
    \nabla_\theta \ell_s,
    \delta\theta
  \right\rangle,
\]
comparison with Eq.~\ref{eq:app_hybrid_gradient_decomposition} yields
\begin{equation}
  \nabla_{V_m(s)} \ell_s
  =
  w_m(s;\theta,\operatorname{sg}(V))\,r_s,
  \label{eq:app_textual_branch}
\end{equation}
and
\begin{equation}
  \nabla_\theta \ell_s
  =
  \sum_{m=1}^{M}
  \left\langle r_s,V_m(s)\right\rangle
  \nabla_\theta w_m(s;\theta,\operatorname{sg}(V)).
  \label{eq:app_numerical_branch}
\end{equation}
The first branch, $\nabla_{V_m(s)}\ell_s=w_m(s)r_s$, is the direct mixture-path sensitivity used for textual failure attribution. The second branch, $\nabla_\theta \ell_s$, is the standard numerical gradient used to update the routing network. Because the router observes $\operatorname{sg}(V)$, the indirect path from heuristic outputs to the loss through the router,
\[
V_m \rightarrow w \rightarrow \student \rightarrow \ell_s,
\]
is removed. Thus the heuristic-output sensitivity contains only the direct path
\[
V_m \rightarrow \student \rightarrow \ell_s.
\]
\subsection{Add and Drop Algorithms}
\label{sec:add_alg}
In this appendix, we present the algorithmic details of slot add and drop in Algorithms~\ref{alg:fault_pretrain} and~\ref{alg:drop}.
\begin{algorithm}[H]
\caption{Failure-Mode-Targeted Bank Expansion (Add)}
\label{alg:fault_pretrain}
\begin{algorithmic}[1]
\Require Existing bank $\mathcal{B}$, dataset $\mathcal{D}$, validation set $\mathcal{D}_{\text{val}}$, teacher, LLM, threshold $\rho_{\text{admit}}$
\Ensure Expanded bank $\mathcal{B}^+$
\Statex
\State $\mathcal{F} \gets$ decision steps on which every $h \in \mathcal{B}$ performs poorly
\State $\mathcal{C} \gets \textsc{LLM-Propose}(\mathcal{F})$ \Comment{LLM attributes failures into $K$ buckets}
\Statex
\Statex \textbf{// Pretrain candidates on $\mathcal{F}$}
\For{round $t = 1, \dots, T$}
  \For{each $s \in \mathcal{F}$}\Comment{assign each fault case to the best current candidate}
    \State assign $s$ to $h \in \mathcal{C}$ whose output is closest to the teacher's
  \EndFor
  \For{each $h \in \mathcal{C}$}\Comment{refine each candidate on its assigned fault cases}
    \State $\nabla_h \gets \textsc{TextGrad}(h, \text{samples assigned to } h)$
    \State $h' \gets \textsc{LLM-Edit}(h, \nabla_h)$
    \State replace $h \leftarrow h'$ \textbf{if} $h'$ improves agreement on a held-out split
  \EndFor
\EndFor
\Statex
\Statex \textbf{// Greedy admission by marginal coverage gain}
\State $\mathcal{B}^+ \gets \mathcal{B}$
\While{$\mathcal{C} \neq \emptyset$}
  \State $h^\star \gets \arg\max_{h \in \mathcal{C}} \big[ C(\mathcal{B}^+ \cup \{h\}; \mathcal{D}_{\text{val}}) - C(\mathcal{B}^+; \mathcal{D}_{\text{val}}) \big]$
  \State $\Delta^\star \gets C(\mathcal{B}^+ \cup \{h^\star\}; \mathcal{D}_{\text{val}}) - C(\mathcal{B}^+; \mathcal{D}_{\text{val}})$
  \If{$\Delta^\star < \rho_{\text{admit}}$}
    \State \textbf{break}
  \EndIf
  \State $\mathcal{B}^+ \gets \mathcal{B}^+ \cup \{h^\star\}$;\quad $\mathcal{C} \gets \mathcal{C} \setminus \{h^\star\}$
\EndWhile
\State \Return $\mathcal{B}^+$
\end{algorithmic}
\end{algorithm}

\begin{algorithm}[H]
\caption{Redundant-Target Removing (Drop)}
\label{alg:drop}
\begin{algorithmic}[1]
\Require Current bank $\mathcal{B}$, validation set $\mathcal{D}_{\text{val}}$, student model, tolerance $\epsilon$, max drops per call $D_{\max}$, minimum bank size $M_{\min}$
\Ensure Pruned bank $\mathcal{B}^-$
\Statex
\State $\mathcal{B}^- \gets \mathcal{B}$;\quad $d \gets 0$
\State $L_0 \gets \mathcal{L}_{\text{val}}(\mathcal{B}^-; \mathcal{D}_{\text{val}})$ \Comment{baseline validation loss}
\Statex
\While{$d < D_{\max}$ \textbf{and} $|\mathcal{B}^-| > M_{\min}$}
  \For{each $h \in \mathcal{B}^-$}
    \State $L_h \gets \mathcal{L}_{\text{val}}(\mathcal{B}^- \setminus \{h\}; \mathcal{D}_{\text{val}})$
    \State $\delta_h \gets L_h - L_0$ \Comment{degradation from removing $h$}
  \EndFor
  \State $h^\star \gets \arg\min_{h \in \mathcal{B}^-} \delta_h$
  \If{$\delta_{h^\star} > \epsilon$}
    \State \textbf{break} \Comment{no safe removal remains}
  \EndIf
  \State $\mathcal{B}^- \gets \mathcal{B}^- \setminus \{h^\star\}$;\quad $L_0 \gets L_{h^\star}$;\quad $d \gets d + 1$
\EndWhile
\State \Return $\mathcal{B}^-$
\end{algorithmic}
\end{algorithm}

\section{Experimental Setup}
\label{sec: setting}

All experiments were conducted on a workstation equipped with an
NVIDIA RTX A6000 GPU. Unless otherwise specified, we use Claude Sonnet 4.6 and GPT-5.4 as the LLMs for heuristic generation and TextGrad rewriting. For LEHD-TSP, we instead adopt GPT-5 as the
LLM optimizer. On a single NVIDIA RTX A6000, training EPB takes roughly 10 to 72 hours per setting.

\subsection{Hyperparameters}
\begin{table}[!htb]
\centering
\small
\setlength{\tabcolsep}{5pt}
\renewcommand{\arraystretch}{1.1}
\caption{Key hyperparameters of the selected student checkpoints.
The TextGrad interval is the number of batches between TextGrad
backward passes.}
\vspace{6pt}
\label{tab:hyperparameters}
\begin{tabular}{lcccc}
\toprule
\textbf{Parameter}
& \textbf{TSP-50}
& \textbf{TSP-100}
& \textbf{CVRP-50}
& \textbf{LEHD-TSP-100} \\
\midrule
Encoder layers
& 3 & 4 & 6 & 3 \\
Batch size
& 4,096 & 2,048 & 16,384 & 1,024 \\
Learning rate
& $6 \times 10^{-4}$
& $3 \times 10^{-4}$
& $4 \times 10^{-4}$
& $1 \times 10^{-4}$ \\
Epochs
& 144 & 80 & 50 & 50 \\
TextGrad interval
& 100 & 200 & 200 & 200 \\
\bottomrule
\end{tabular}
\end{table}
Table~\ref{tab:hyperparameters} summarizes the key hyperparameters of the selected student checkpoints. All student models use an embedding dimension of 128 and 8 attention heads. The LEHD-TSP student uses a three-layer light encoder and a three-layer heavy decoder.

\begingroup
\color{black}
\tcbset{
  appendixpromptbox/.style={
    enhanced,
    breakable,
    colback=black!8!white,
    colframe=black,
    colbacktitle=black,
    coltitle=white,
    coltext=black,
    fonttitle=\bfseries,
    fontupper=\small,
    boxrule=0.5pt,
    arc=2pt,
    before skip=10pt,
    after skip=10pt
  }
}
\lstdefinestyle{appendixprompttext}{
  language={},
  basicstyle=\ttfamily\small\color{black},
  identifierstyle=\color{black},
  keywordstyle=\color{black},
  commentstyle=\color{black},
  stringstyle=\color{black},
  numberstyle=\tiny\color{black},
  backgroundcolor={},
  frame=none,
  numbers=none,
  title={},
  caption={},
  breaklines=true,
  breakatwhitespace=false,
  columns=fullflexible,
  keepspaces=true,
  showspaces=false,
  showtabs=false,
  showstringspaces=false,
  mathescape=false,
  texcl=false,
  xleftmargin=0pt,
  xrightmargin=0pt,
  aboveskip=0pt,
  belowskip=0pt,
  literate={—}{{---}}3
}

\subsection{Prompt Setting}
\label{app:promptsetting}
\subsubsection{Pipeline overview}

The training loop has two LLM-driven phases. \textbf{(I) Proposal/init.}
For each new heuristic slot, the LLM is asked to (a) propose a strategy
in natural language, then (b) implement that strategy as a Python
module. (II) TextGrad refinement. In TextGrad epochs, every $n$ batches (Table 8), failure cases (where the heuristic's argmax disagrees with the LEHD teacher's) drive a backward pass that produces text gradients on the code and on the description; an optimizer prompt then synthesises those gradients into an updated module. A single shared context block
(\texttt{\_SPARSE\_TEACHER\_CONTEXT}) is embedded into every system prompt
so all calls share the same problem framing.

\subsubsection{Shared context}

\begin{tcolorbox}[appendixpromptbox,title={Shared context (\_SPARSE\_TEACHER\_CONTEXT)}]
\begin{lstlisting}[style=appendixprompttext]
**Subgraph decoder formulation (LEHD-style).**
At each decoding step the heuristic receives ONLY the unvisited subgraph
plus the start (first_loc) and current end (current_loc) of the partial tour.
Visited nodes are NOT shown — the input subgraph IS the remaining problem.
The output is a score over the K candidate next-nodes, where K shrinks across
the rollout from N-1 down to 1.  Your code MUST work for ANY K.

Progress within the tour is implicit: it is exposed ONLY through K
(= unselected_locs.size(1)).  No global N, step index, or step_norm is provided
— this matches what the LEHD teacher itself sees.  If your strategy needs a
"how-far-along" signal, derive it from the geometry of unselected_locs (e.g.
spread, mean distance to first_loc) or from K itself, and AVOID hard-coded
thresholds tied to a particular N.

**Critical: the teacher distribution is near one-hot.**
The teacher policy is a near-deterministic LEHD model:
  - top-1 candidate receives ~95% probability
  - top-2 candidate receives ~4.5% probability
  - the remaining K-2 candidates share less than 0.5% combined

Two implications for heuristic design:

1. **Output sharp, decisive scores.** The winning candidate's score should be
   substantially higher than the runner-up. After softmax with the training
   temperature, your output should look near-one-hot in clear-cut situations
   — not a smoothly graded distribution. A decisive winner gap matters more
   than carefully shaping the long tail.

2. **Top-1 is what matters; top-2 is a soft backup; the rest is irrelevant.**
   Don't waste complexity on shaping the bottom of the distribution. A simple
   "score = -distance + lookahead_bonus" can outperform an 8-signal weighted
   mixture if the simpler one consistently picks the right top-1.

Heuristics are evaluated on:
  (a) whether their argmax matches the teacher's argmax  (primary)
  (b) whether their top-2 contains the teacher's argmax  (secondary)
The full distribution shape does not matter. Avoid over-engineering for it.
\end{lstlisting}
\end{tcolorbox}

\subsubsection{Phase I --- initial proposal}

\textbf{Propose --- \texttt{\_make\_idea\_prompt} (system)}

\begin{tcolorbox}[appendixpromptbox,title={Propose strategy (system)}]
\begin{lstlisting}[style=appendixprompttext]
You are an expert in the 2D symmetric Euclidean Travelling Salesman Problem.
Training is on {problem_size}-city instances, but the heuristic must generalise to any
subgraph size K in [1, {problem_size - 1}] because it is called repeatedly during decoding
on a shrinking unvisited set.

{_SPARSE_TEACHER_CONTEXT}

Describe ONE strategy for choosing the next city at each decoding step (1-3 sentences).
Focus on a strategy that produces a CLEAR top-1 winner from the candidates, not one
that ranks many of them similarly. Top-1 hit rate is the primary metric.

The strategy must rely only on (unselected_locs, first_loc, current_loc).
There is NO global "visited mask" or full-instance node list available — visited nodes
are not in the input.  Progress within the tour is exposed ONLY via K =
unselected_locs.size(1).  No step index or step_norm is provided.

Diversity guideline:
Avoid proposing a strategy that is functionally identical to an existing one.
- You may reuse existing features as secondary signals, as long as the overall
  combination or decision logic is meaningfully new.
- Simpler strategies with one or two strong signals are preferred over complex
  multi-signal blends.

Use only TSP reasoning; do not mention VRP concepts.
Do NOT write code or mathematical formulas.
\end{lstlisting}
\end{tcolorbox}

\textbf{Implement --- \texttt{\_make\_system\_prompt} (system)}

\begin{tcolorbox}[appendixpromptbox,title={Implement strategy (system)}]
\begin{lstlisting}[style=appendixprompttext]
You are writing heuristic modules for the 2D symmetric Euclidean Travelling Salesman Problem (TSP).
Training is on {problem_size}-city instances, but at inference time K (the number of unvisited
candidate cities at each decoding step) varies from {problem_size - 1} down to 1.  Your code MUST
work for ANY K.

{_SPARSE_TEACHER_CONTEXT}

Your task is to write a Python module containing exactly one function:

  def heuristic(unselected_locs, first_loc, current_loc) -> torch.Tensor:
      """Score each unvisited candidate.

      Args:
        unselected_locs:  torch.Tensor [B, K, 2]  candidate (x,y) coords in [0,1]^2
        first_loc:        torch.Tensor [B, 2]    start node coords
        current_loc:      torch.Tensor [B, 2]    current end-of-tour coords

      Returns:
        torch.Tensor [B, K]  finite scores; higher = more preferred next-node.
      """

Rules:
- Imports allowed: torch, torch.nn.functional as F, math. Nothing else.
- All tensor operations must be batched (first dim = B).  K may differ across calls.
- No Python `for` loops over batch dim or candidate dim.  No `while` loops.
  No itertools.  All ops must be vectorised torch ops.
- No file I/O, network access, environment access, printing, or side effects.
- Do not modify input tensors in-place.
- Coordinates are in [0,1]^2.  Embed any constants directly in code.
- No PARAMS dict required.

**Critical: K=1 edge case.** Your code WILL be called with K=1 at the last decoding
step.  Common K=1 pitfalls that produce NaN/inf and cause your heuristic to be
discarded:
  - `tensor.std(dim=K_dim)` defaults to `unbiased=True` which needs >=2 samples;
    at K=1 it returns NaN.  Use `tensor.std(dim=K_dim, unbiased=False)` or just
    skip dispersion calcs when K==1.
  - `tensor.var(dim=K_dim)` has the same default-unbiased issue.  Use
    `var(dim=K_dim, unbiased=False)`.
  - Dividing by `K - 1`, `(K * (K - 1))`, or `count - 1` blows up at K=1.  Use
    `max(K - 1, 1)` or add a `+ 1e-8` denominator guard.
  - Pairwise distance matrices [B, K, K] degenerate to [B, 1, 1] with the only
    entry being self-distance 0; downstream divisions or normalisations on this
    can NaN.
  - When K=1, the answer is forced (one candidate); your code can early-return
    `torch.zeros(B, 1, ...)` if any computation would be degenerate.

Return a torch.Tensor of shape [B, K] with finite scores.
Return ONLY the complete Python module, no explanation, no markdown fences.
\end{lstlisting}
\end{tcolorbox}

\textbf{Implement --- user message + module template}

\begin{tcolorbox}[appendixpromptbox,title={Implement strategy (user) + HEURISTIC\_MODULE\_TEMPLATE}]
\begin{lstlisting}[style=appendixprompttext]
Implement the following heuristic strategy as a Python module.

Strategy: {description}

Use this structure:

import torch
import torch.nn.functional as F
import math


def heuristic(unselected_locs, first_loc, current_loc):
    """Score each unvisited candidate.

    Args:
        unselected_locs: torch.Tensor [B, K, 2]
        first_loc:       torch.Tensor [B, 2]
        current_loc:     torch.Tensor [B, 2]

    Returns:
        torch.Tensor [B, K] of finite scores; higher = preferred next-node.

    Note: progress within the tour is exposed only implicitly via K =
    unselected_locs.size(1).  No global N or step index is provided.
    """
    # TODO: Replace with strategy implementation.
    raise NotImplementedError
\end{lstlisting}
\end{tcolorbox}

\subsubsection{Phase II --- TextGrad refinement loop}

\textbf{Backward (code) --- \texttt{\_make\_backward\_prompt} (system)}

\begin{tcolorbox}[appendixpromptbox,title={Backward on code (system)}]
\begin{lstlisting}[style=appendixprompttext]
You are the BACKWARD PASS of a TextGrad optimizer for TSP next-node selection heuristics.
Training was on {problem_size}-city instances; K (candidate count) varies at inference.

{_SPARSE_TEACHER_CONTEXT}

The heuristic uses signature
  heuristic(unselected_locs [B,K,2], first_loc [B,2], current_loc [B,2]) -> [B,K]

You will see a heuristic function and one decision step where it picked a different
top-1 from the teacher. Produce a TEXT GRADIENT: a concise diagnosis of why the code's
score for the teacher's top-1 candidate fell behind its (wrong) argmax.

This is NOT a code fix. Output only a diagnosis in this exact format:
LOGIC: [what the code currently computes for ranking these top candidates]
TEACHER: [the teacher chose candidate X with ~95% mass — what about X would the code recognize?]
FLAW: [why the code's score for X fell below the score for the (wrong) argmax]
DIRECTION: [what change would push X's score above the wrong argmax — focus on
            making the winning gap larger or fixing the tiebreaker, NOT on shaping
            the full distribution]

Be concrete about geometry (distances, angles, tour-closing cost, cluster structure).
2-5 sentences total. No code. Note that K varies — your reasoning must hold for any K.
\end{lstlisting}
\end{tcolorbox}

\textbf{Backward (code) --- per-failure user message}

\begin{tcolorbox}[appendixpromptbox,title={Backward on code (per-failure user message)}]
\begin{lstlisting}[style=appendixprompttext]
Heuristic code:
{code}

Failure case [KL(teacher||heuristic)={kl_val:.4f}]:{rank_line}
  Step: {int(step)}/{int(step_max)} ({100*int(step)//int(step_max)}% complete)
  Unvisited candidates remaining (K): {n_unv}
  Current position: {cur_pos}
  Tour-start position: {first_pos}
  5 nearest unvisited (geometry only): {[f"pos={x['pos']} d={x['dist_to_cur']}" for x in nu]}
  Heuristic top-2: {h3}
  Teacher   top-2: {t3}

Generate the text gradient for this failure case.
Focus especially on why the heuristic's top-ranked candidate differs from the teacher's,
described purely in geometric terms (positions, distances, angles).  Candidates have NO
stable identifier — your reasoning must be over geometry, not node ids.
\end{lstlisting}
\end{tcolorbox}

\textbf{Backward (description) --- \texttt{\_make\_backward\_desc\_prompt} (system)}

\begin{tcolorbox}[appendixpromptbox,title={Backward on description (system)}]
\begin{lstlisting}[style=appendixprompttext]
You are the BACKWARD PASS of a TextGrad optimizer for TSP heuristic descriptions.
Training was on {problem_size}-city instances; the description must remain valid for any K.

{_SPARSE_TEACHER_CONTEXT}

Given a heuristic's natural language description and a decision step where the heuristic
picked the wrong top-1, produce a TEXT GRADIENT: what does the description fail to capture
that would help the implementation pick the right top-1 next time?

Format:
MISSING: [what geometric reasoning the description omits or under-emphasises]
MISLEADING: [anything in the description that points the implementation in the wrong direction]
DIRECTION: [how the description should be revised — focus on the top-1 ranking decision,
            not on shaping a full distribution]

2-4 sentences. No code. Focus on strategy-level geometric reasoning that affects top-1.
\end{lstlisting}
\end{tcolorbox}

\textbf{Backward (description) --- per-failure user message}

\begin{tcolorbox}[appendixpromptbox,title={Backward on description (per-failure user message)}]
\begin{lstlisting}[style=appendixprompttext]
Heuristic description: "{description}"

Heuristic code:
{code}

Failure case [KL={kl_val:.4f}]:
  Step: {int(step)}/{int(step_max)} ({100*int(step)//int(step_max)}% complete)
  Current pos: {sample_info.get('cur_pos', '?')}
  Heuristic top-2 (geometry only): {h3}
  Teacher   top-2 (geometry only): {t3}

Generate the text gradient for the DESCRIPTION (not the code).
Reason in geometric terms only — candidates have no stable identifier.
\end{lstlisting}
\end{tcolorbox}

\textbf{Optimizer step --- \texttt{\_make\_optimizer\_prompt} (system)}

\begin{tcolorbox}[appendixpromptbox,title={Optimizer step (system)}]
\begin{lstlisting}[style=appendixprompttext]
You are part of an optimization system that improves Python heuristic functions for TSP
next-node selection. Training was on {problem_size}-city instances; K (candidate count)
varies from {problem_size - 1} down to 1 at inference time.

{_SPARSE_TEACHER_CONTEXT}

Heuristic signature is fixed:
  heuristic(unselected_locs [B,K,2], first_loc [B,2], current_loc [B,2]) -> [B,K]

You will receive the current heuristic code and text gradients (diagnoses of failure cases).
Your task: synthesize the gradients and produce an improved version of the code that picks
the teacher's top-1 candidate more often (and at least keeps it in the top-2 when it can't win).

- Address the most common and severe failure patterns across all gradients
- Momentum gradients indicate persistent failure modes — treat them as high priority
- **Sharpen the winning gap.** If the existing code outputs scores that are too close
  together at the top, increase the multiplier on the dominant signal so the winner
  pulls ahead more decisively after softmax.
- Keep the function signature unchanged.
- Imports: torch, torch.nn.functional as F, math only.
- Hard constraint: NO Python `for` loops over batch dim or candidate dim. NO `while` loops.
  All ops must be vectorised torch ops.
- Code must work for ANY K (do not assume K = {problem_size - 1}).
- **K=1 numerical safety**: `tensor.std/var(dim, unbiased=True)` returns NaN at K=1.
  Always pass `unbiased=False` when reducing along the K dimension.  Avoid dividing
  by `K-1` or `K*(K-1)` without guarding (`max(K-1, 1)` or `+ 1e-8`).  Early-return
  `torch.zeros(B, 1, ...)` when K==1 if any computation would be degenerate.
- Embed constants directly in code. No markdown fences.
- Prefer simplicity. A 5-line scorer that picks top-1 right is better than a 50-line
  multi-signal blend that only gets top-3 right.

Put the improved Python module between {new_variable_start_tag} and {new_variable_end_tag}.
\end{lstlisting}
\end{tcolorbox}

\subsubsection{Teacher identity in prompts}
\label{app:teacher-identity}

We test whether EPB relies on explicit teacher identity by removing the POMO/LEHD names and all background information about these models from the LLM prompts. In these blind runs, the LLM receives only the optimization task, code interface, and action feedback from an unnamed teacher.

\begin{table}[ht]
\centering
\caption{Solution-cost gaps (\%) to the teacher with original and blind prompts. Lower is better.}
\vspace{6pt}
\label{tab:teacher-identity}
\setlength{\tabcolsep}{6pt}
\begin{tabular}{lcccc}
\toprule
& \multicolumn{2}{c}{\textbf{Greedy}}
& \multicolumn{2}{c}{\textbf{Search}} \\
\cmidrule(lr){2-3}\cmidrule(lr){4-5}
\textbf{Setting}
& Original & Blind & Original & Blind \\
\midrule
POMO CVRP-50
& $+2.88$ & $+3.32$ & $+1.13$ & $+1.12$ \\
LEHD TSP-100
& $+2.57$ & $+2.51$ & $+0.34$ & $+0.33$ \\
\bottomrule
\end{tabular}
\end{table}

The blind runs produce similar solution-cost gaps. These results suggest that explicit teacher identity is not required for the solution quality achieved in these settings. The LLM can still draw on its general knowledge of routing heuristics.

\endgroup
\section{Additional Results and Discussions}
\label{sec:appendixresults}
\subsection{Sensitivity to Random Seeds}
\label{app:seed-sensitivity}

\begin{figure}[ht]
  \centering
  \includegraphics[width=0.5\linewidth]{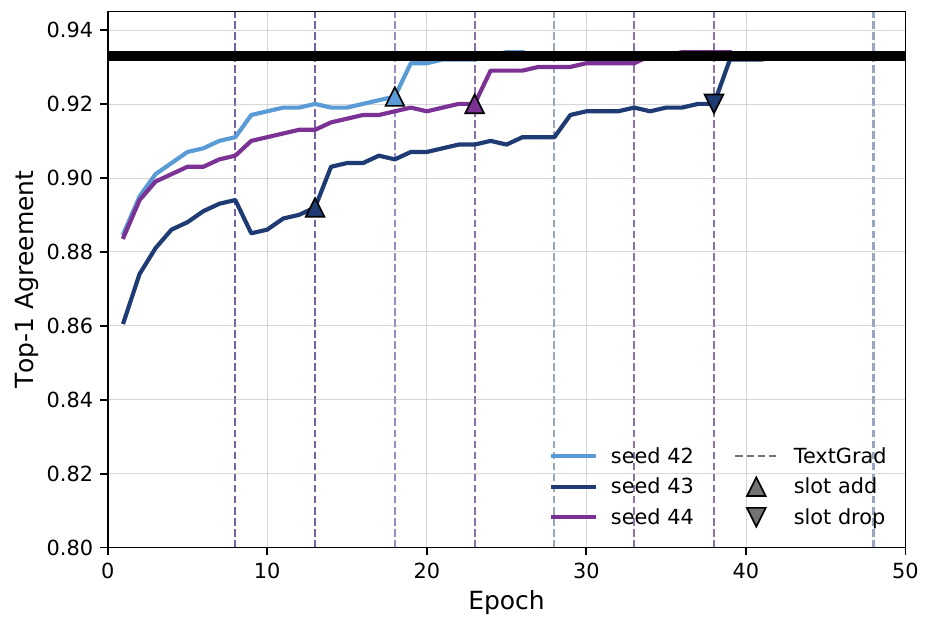}
  \caption{EPB convergence across seeds on POMO-TSP-100 with DeepSeek-V4-Flash.}
  \label{fig:seed-convergence}
\end{figure}

\noindent
\begin{minipage}[t]{0.55\linewidth}
  \vspace{0pt}
  We run EPB with DeepSeek-V4-Flash on POMO-TSP-100 using
  three random seeds (42, 43, and 44). Convergence speed varies
  with bank initialization, but all three runs reach similar
  final validation top-1 agreement
  (Figure~\ref{fig:seed-convergence}). The best greedy gaps
  range from 3.56\% to 5.08\%, while multi-start gaps remain
  below 0.21\% for all seeds
  (Table~\ref{tab:seed-performance}). These results support
  consistent final performance across seeds in this setting.
\end{minipage}\hfill
\begin{minipage}[t]{0.42\linewidth}
  \vspace{0pt}
  \centering
  \captionof{table}{Solution-cost gaps (\%) to the teacher across seeds.}
  \label{tab:seed-performance}
  \vspace{6pt}
  \setlength{\tabcolsep}{12pt}
  \begin{tabular}{ccc}
    \toprule
    \textbf{Seed} & \textbf{Greedy} & \textbf{Search} \\
    \midrule
    42 & $+3.56$ & $+0.036$ \\
    43 & $+4.66$ & $+0.138$ \\
    44 & $+5.08$ & $+0.204$ \\
    \bottomrule
  \end{tabular}
\end{minipage}
\par\medskip

\subsection{LLM Cost}
\label{app:cost}

\paragraph{One-time LLM usage.}
EPB uses the LLM only during offline distillation and requires no LLM calls at deployment. Table~\ref{tab:cost-usage} reports the usage of the three settings with the highest LLM usage.

\begin{table}[ht]
\centering
\small
\caption{LLM usage for one complete distillation run.}
\vspace{6pt}
\label{tab:cost-usage}
\setlength{\tabcolsep}{8pt}
\begin{tabular}{lrr}
\toprule
\textbf{Setting} & \textbf{LLM calls} & \textbf{Tokens} \\
\midrule
POMO CVRP-100 & 8{,}071 & 29.66M \\
LEHD TSP-100  & 5{,}471 & 30.33M \\
LEHD CVRP-100 & 9{,}704 & 44.24M \\
\bottomrule
\end{tabular}
\end{table}

\paragraph{Cost breakdown.}
TextGrad accounts for 90.6\% of all calls, of which 96\% are paired code and description backward passes. Input tokens account for 79\% of all tokens, and 93\% of these come from repeated prefixes. Thus, approximately 73\% of all tokens are potentially cacheable. In the runs reported in Table~\ref{tab:cost-open}, prompt caching served 70--79\% of the input tokens.

Larger gradient batches and less frequent TextGrad and description updates may further reduce calls, but we have not tested these changes. To our knowledge, this is the first work to distill sequential NCO policies into banks of executable programs evolved by an LLM. We prioritize establishing feasibility and fidelity and have not yet systematically optimized distillation cost.

\paragraph{Open models and parameter count.}\mbox{}\par
\noindent
\begin{minipage}[t]{0.5\linewidth}
  \vspace{0pt}
  To test lower-cost distillation and examine the effect of
  LLM size, we compare DeepSeek-V4-Flash and DeepSeek-V4-Pro
  on POMO-TSP-100. Flash has 284B total parameters and 13B
  activated parameters. Pro has 1.6T total parameters and
  49B activated parameters. Both runs follow the same
  training and evaluation protocol. Both models achieve
  solution quality close to the main results
  (Table~\ref{tab:cost-open}), showing that the smaller
  Flash model also supports effective distillation in
  this setting.
\end{minipage}\hfill
\begin{minipage}[t]{0.45\linewidth}
  \vspace{0pt}
  \centering
  \small
  \captionof{table}{Distilling POMO-TSP-100 with open models.
  Gaps are relative to the teacher.}
  \label{tab:cost-open}
  \vspace{6pt}
  \setlength{\tabcolsep}{6pt}
  \begin{tabular}{@{}lrr@{}}
    \toprule
    \textbf{LLM} & \textbf{Greedy (\%)} & \textbf{Search (\%)} \\
    \midrule
    Main results      & $+3.79$ & $+0.072$ \\
    DeepSeek-V4-Flash  & $+3.56$ & $+0.036$ \\
    DeepSeek-V4-Pro    & $+2.71$ & $-0.071$ \\
    \bottomrule
  \end{tabular}
\end{minipage}
\par\medskip
\paragraph{Comparison with program-search methods.}
EoH \cite{liu2024eoh} uses approximately 1{,}000 LLM queries for TSP and FSSP and 2{,}000 for online bin packing. FunSearch generates approximately $10^6$ program samples. EPB uses roughly 5{,}000--10{,}000 calls, broadly comparable in scale to EoH. These methods optimize task objectives, whereas EPB fits a teacher's sequential decisions. Since the objectives differ and neither method reports token usage, this comparison concerns only orders of magnitude.
\subsection{ Fidelity}
\label{sec:tour_fidlity}
\subsubsection{Absolute Solution Cost}
\label{sec:absolute_costs}
Table~\ref{tab:absolute_costs} reports the absolute mean solution costs of the teachers and all interpretable surrogate methods on TSP and CVRP, complementing the relative gaps in Table~\ref{tab:teacher_vs_student}.

\begin{table}[ht]
\centering\small
\caption{Absolute solution cost of the teacher and of every method in Table~\ref{tab:teacher_vs_student}. Ref.: teacher cost (reference, not a competitor). $N$: number of nodes. All methods in a row use the same test instances: 1{,}000 per setting, except 1{,}500 for POMO TSP-50 and 256 for LEHD CVRP-100.}
\vspace{6pt}
\label{tab:absolute_costs}
\setlength{\tabcolsep}{2pt}
\begin{tabular}{@{}l@{\hspace{4pt}}r*{7}{c}@{\hspace{3pt}}*{7}{c}@{}}
\toprule
& & \multicolumn{7}{c}{\textbf{TSP}} & \multicolumn{7}{c}{\textbf{CVRP}} \\
\cmidrule(lr){3-9}\cmidrule(l){10-16}
\textbf{Teacher} & $N$
& Ref. & GPT-6 & Fable & Tree & NAM & Mo\"ET & EPB
& Ref. & GPT-6 & Fable & Tree & NAM & Mo\"ET & EPB \\
\midrule
\multicolumn{16}{@{}l}{\textit{Greedy decoding} $\downarrow$}\\
POMO & 50  & 5.71 & 6.26 & \underline{6.17} & 6.65 & 6.35 & 6.76 & \textbf{5.88} & 10.75 & 12.35 & \underline{12.17} & 13.54 & 12.67 & 12.27 & \textbf{11.06} \\
POMO & 100 & 7.83 & 8.69 & \underline{8.54} & 9.40 & 8.91 & 9.46 & \textbf{8.13} & 16.19 & 18.84 & \underline{18.52} & 20.82 & 19.33 & 18.72 & \textbf{16.83} \\
LEHD & 100 & 7.81 & 8.71 & \underline{8.56} & 9.38 & 8.89 & 9.56 & \textbf{8.01} & 16.28 & 18.87 & \underline{18.58} & 19.85 & 19.25 & 18.79 & \textbf{16.86} \\
\midrule
\multicolumn{16}{@{}l}{\textit{Search decoding} $\downarrow$}\\
POMO & 50  & 5.70 & 5.84 & \underline{5.84} & 5.96 & 5.84 & 6.10 & \textbf{5.70} & 10.47 & 11.56 & 11.45 & 11.38 & 11.51 & \underline{11.26} & \textbf{10.59} \\
POMO & 100 & 7.80 & 8.15 & \underline{8.10} & 8.44 & 8.18 & 8.64 & \textbf{7.80} & 15.88 & 17.92 & 17.67 & 17.80 & 17.85 & \underline{17.33} & \textbf{16.08} \\
LEHD & 100 & 7.77 & 7.93 & \underline{7.90} & 8.04 & 7.97 & 8.33 & \textbf{7.80} & 15.97 & 18.26 & 17.99 & 18.28 & 17.91 & \underline{17.81} & \textbf{16.31} \\
\bottomrule
\end{tabular}
\end{table}

\subsubsection{Tour Comparison}
\label{sec: touragreement}
\begin{figure*}[ht]
  \centering
  \begin{minipage}[t]{0.48\linewidth}
    \centering
    \includegraphics[width=\linewidth]{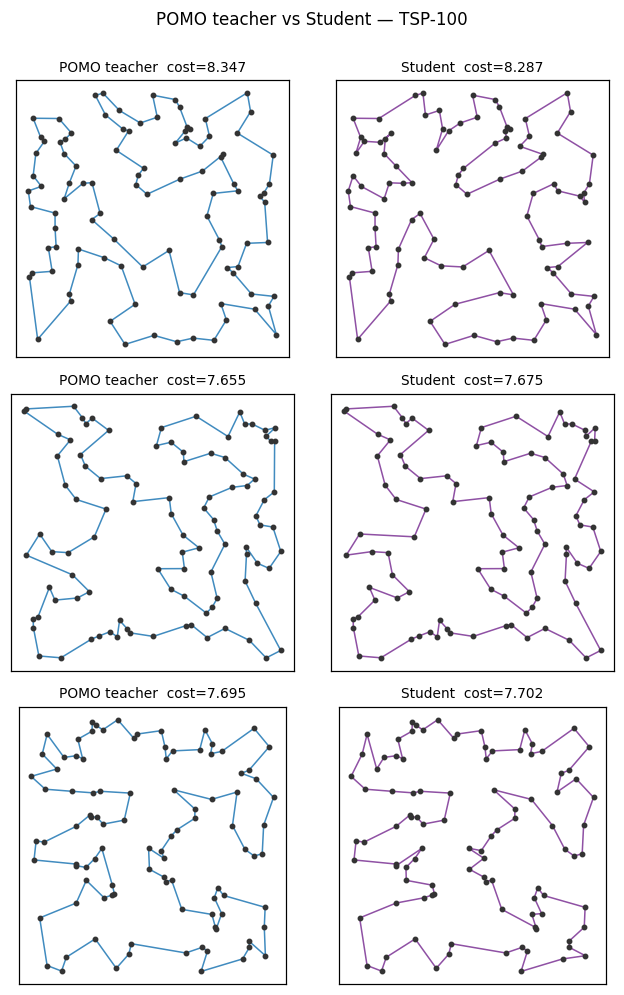}
    \caption{POMO-TSP-100 Tour Comparison}
    \label{fig: pomotsp100comp}
  \end{minipage}\hfill
  \begin{minipage}[t]{0.48\linewidth}
    \centering
    \includegraphics[width=\linewidth]{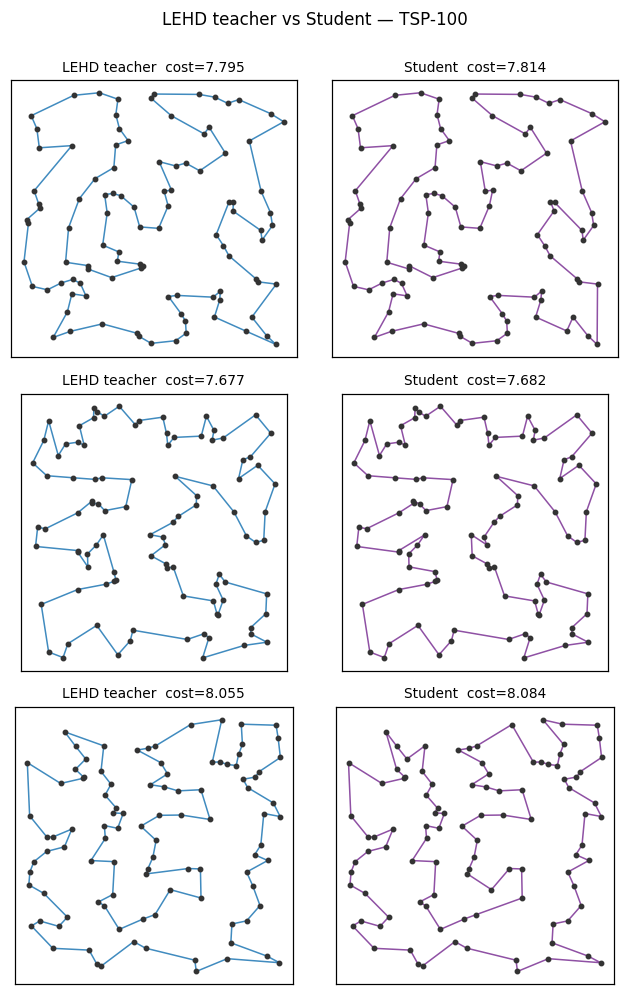}
    \caption{LEHD-TSP-100 Tour Comparison}
    \label{fig: lehdtsp100comp}
  \end{minipage}
\end{figure*}

We compare teacher and EPB tours under search decoding to reduce the impact of error accumulation in greedy decoding. Figures~\ref{fig: pomotsp100comp}--\ref{fig: lehdtsp100comp} show examples across teacher models.

\subsection{Heuristic Gallery\texorpdfstring{\protect\footnotemark}{}}
\label{sec: heu_gallery}
\footnotetext{To conserve space, this appendix reports condensed natural-language summaries in place of the heuristics' source description/code. Each summary is produced by prompting a large language model to refine the description.}

This appendix presents three example heuristic (program) banks distilled from POMO for TSP-100 (Table~\ref{tab:pomo_tsp100_heuristic_bank_descriptions}), LEHD for TSP-100 (Table~\ref{tab:lehd_tsp100_heuristic_bank}), and POMO-CVRP-50 (Table~\ref{tab:heuristic_bank_descriptions})
\newcolumntype{Y}{>{\raggedright\arraybackslash}X}
\setlength{\LTcapwidth}{\textwidth}  

\begin{table*}[!htb]
\centering
\footnotesize
\setlength{\tabcolsep}{5pt}
\renewcommand{\arraystretch}{1.12}
\caption{TSP-100 heuristic bank distilled from the POMO teacher.}
\label{tab:pomo_tsp100_heuristic_bank_descriptions}
\begin{tabularx}{\textwidth}{c >{\raggedright\arraybackslash}X}
\toprule
\textbf{Heuristic ID} & \textbf{Description} \\
\midrule
H00 &
Combines proximity with an early isolation bonus and a late pull
towards the start, then uses closing-cost estimates through two
or three remaining cities near completion. \\

H01 &
Favours nearby cities throughout, adds a far-from-start bonus late
in the tour, then penalises the distance through a candidate and
back to the start while further strengthening proximity. \\

H02 &
Rewards peripheral cities and outward steps early, strengthens
proximity as the tour progresses, and finishes with a nearest-onward
connection cost, a renewed periphery bonus and a pull towards the start. \\

H03 &
Adds mid-tour periphery and far-from-start bonuses to proximity,
briefly favours cities nearer the start later, then restores
a strong far-from-start preference near completion. \\

H04 &
Favours nearby cities with a large gap between their nearest and
second-nearest unvisited neighbours. With eight or fewer cities
remaining, it estimates up to six greedy onward steps and a return
to the start, with extra weight on proximity. \\

H05 &
Combines proximity, periphery and closing-cost estimates through
three cities. It emphasises peripheral and far-from-start cities
when few remain, with separate scores for the last one or two cities. \\

H06 &
Combines proximity, cheap onward connections and a far-from-start
bonus, then increasingly weights closing-cost estimates through
two or three cities, with a separate term for the last two cities. \\

H07 &
Adds a heading-alignment bonus among nearby candidates early and
mid-tour, then increasingly penalises the distance through a candidate
and back to the start. This closing term is disabled for the last
five cities. \\

H08 &
Combines proximity with a centroid-based geometric bonus, then
favours shorter distances through a candidate and back to the start.
With ten or fewer cities remaining, it uses greedy completion
estimates with extra weight on the immediate step. \\

H09 &
Rewards proximity, dense angular sectors and peripheral position
early, then gives more weight to isolation and short closing paths.
Near completion, it discourages visiting cities close to the start. \\

H10 &
Balances immediate distance against the distance to each candidate's
nearest unvisited neighbour, increasingly rewarding isolated cities
as the tour progresses. \\

H11 &
Favours nearby cities, with a mid-tour bonus for moving towards
the remaining cities' centroid. The directional bonus grows with
centroid distance, while a far-from-start bonus becomes stronger late. \\

H12 &
Combines strong proximity with early guidance towards the remaining
cities' centroid. Its start-distance term changes from a penalty
to a bonus, while the cost of continuing through the candidate's
nearest unvisited neighbour and back to the start gains weight. \\

H13 &
Combines normalised proximity with early bonuses for forward motion,
sparse neighbourhoods and closeness to the remaining cities'
centroid, then gradually shifts towards minimising the distance
through a candidate and back to the start. \\
\bottomrule
\end{tabularx}
\end{table*}

\begingroup
\footnotesize
\setlength{\tabcolsep}{5pt}
\renewcommand{\arraystretch}{1.12}
\begin{xltabular}{\textwidth}{c Y}
\caption{TSP-100 heuristic bank distilled from the LEHD teacher.}
\label{tab:lehd_tsp100_heuristic_bank}\\
\toprule
\textbf{Heuristic ID} & \textbf{Description} \\
\midrule
\endfirsthead
\caption[]{TSP-100 heuristic bank distilled from the LEHD teacher (continued).}\\
\toprule
\textbf{Heuristic ID} & \textbf{Description} \\
\midrule
\endhead
\midrule
\multicolumn{2}{r}{\emph{Continued on next page}}\\
\endfoot
\bottomrule
\endlastfoot
H00 &
Favours short steps, with additional bonuses for the two nearest cities.
It penalises isolation and increasingly favours cities near the start
as fewer cities remain. \\

H01 &
Combines the immediate step with the cheapest onward connection and
a weighted return from that same onward city to the start.
The return term gains weight as fewer cities remain. \\

H02 &
Scores the extra distance of visiting a candidate before returning
to the start, then adds the cheapest further insertion.
The insertion term gains weight as fewer cities remain. \\

H03 &
Combines proximity and an onward-step closure cost with bonuses for
isolated cities and cities far from the start, while postponing the
city closest to the start as closure becomes more important. \\

H04 &
Combines proximity with the distance through each candidate's nearest
remaining neighbour and back to the start.
A small alignment bonus favours steps directed towards the start. \\

H05 &
Combines short steps with a tangential sweep around the remaining
cities' centroid, penalises inward movement, and adds bonuses for
peripheral and isolated cities. Return-to-start terms guide closure. \\

H06 &
Balances proximity against isolation and the gap between the two
nearest onward connections, with additional bonuses for peripheral
cities and steps along the heading. \\

H07 &
Favours short steps while discouraging early use of the city closest
to the start. A preference for cities beyond the median distance
from the start strengthens as fewer cities remain. \\

H08 &
Favours a sideways step towards the denser side of the heading,
restricting choices to cities within a local distance allowance
of the nearest city. \\

H09 &
Favours directions near the centre of the largest empty forward
angular sector, penalises backward directions, and adds a small
bonus for longer steps within that sector. \\

H10 &
Favours short steps within a forward cone, but turns towards a
backward region when forward options are absent or sparse and
the backward region is dense or offers sufficient progress
towards the start. \\
\end{xltabular}
\endgroup

\begin{table*}[!htb]
\centering
\footnotesize
\setlength{\tabcolsep}{5pt}
\renewcommand{\arraystretch}{1.12}
\caption{CVRP-50 heuristic bank distilled from the POMO teacher.}
\label{tab:heuristic_bank_descriptions}
\begin{tabularx}{\textwidth}{c >{\raggedright\arraybackslash}X}
\toprule
\textbf{Heuristic ID} & \textbf{Description} \\
\midrule
H00 &
Far-customer savings: a dominant savings term favours customers far from the depot but close to the vehicle; it returns mainly when all feasible customers are closer to the depot than to the vehicle. \\

H01 &
Centroid pivot: savings plus inverse distance, with depot return driven by whether the depot is
closer than the current node to the remaining customers' centroid (ratio 0.75). \\

H02 &
Local demand density: rewards unvisited demand within radius 0.12 of a candidate, scaled by inverse
squared distance. The depot opens only when no customer fits and capacity is at most 0.15. \\

H03 &
Homeward corridor: penalises the perpendicular distance to the current-to-depot line and rewards
customers on that segment. Depot return gains weight below 0.25 capacity. \\

H04 &
Ordered sweep: visits customers in angular order around the depot in the direction set by the
first customer, favouring the next in order and the angular extreme of the unvisited set. \\

H05 & Capacity-pressure blend: demand-weighted proximity with an angular term favouring customers just ahead counter-clockwise around the depot while underway; plain proximity with at most three customers left. The depot penalty grows with feasible customers. \\

H06 &
Gap pivot: near the depot with more than 0.7 capacity it turns towards the largest angular gap
between feasible customers; at the depot it favours high demands far from the first arm. \\

H07 &
Angular continuity: keeps the route in the current direction from the depot, sharper once underway.
Returns when the remaining demand lies in opposite or distant $30^\circ$ sectors. \\

H08 &
Angular proximity to the current position as seen from the depot (Gaussian, $0.5$\,rad), with a
bonus for leaving a quarter of the capacity. Below 0.3 capacity it favours near-depot customers. \\

H09 &
Detour: penalises the extra distance of going home via the customer rather than directly. Depot
return gains weight below 0.5 capacity or with at most two feasible customers. \\

H10 & Frontier and isolation: rewards customers far from the remaining customers' centroid and with few unvisited neighbours within radius 0.15; below 0.2 capacity it favours customers near the depot. \\

H11 & Outward continuation: rewards customers far from the remaining customers' centroid, and those nearer to the vehicle than to the depot with high savings; below 0.3 capacity it favours exact fits and near-depot customers. \\

H12 & Frontier with perfect-fit trigger: favours nearby customers far from the remaining customers' centroid, with a strong bonus for demands matching the remaining capacity within 0.03; the depot is allowed only when no such fit exists and capacity is low. \\

H13 &
Sparse-sector pivot: when the current $60^\circ$ sector holds fewer than two feasible customers and
the opposite sector at least two, it favours the opposite sector or the depot. \\

H14 &
Clean-fit chaining: rewards leftovers that match another unvisited demand and customers in dense
$\pm30^\circ$ clusters; with at most three remaining it serves the farthest first. \\

H15 &
Bimodal sector test: returns to the depot when the vehicle is in the smaller of two dominant
$45^\circ$ demand sectors more than $90^\circ$ apart; starts from isolated, low-demand customers. \\

H16 &
Departure specialist: at the depot with a full vehicle it scores arm alignment, angular density
and average savings instead of proximity; elsewhere it is the nearest-customer rule. \\

H17 &
Three-regime switch: first-arm alignment and cluster density at departure, savings and current-arm
alignment underway, nearest low-demand customers at or below 0.15 capacity. \\

H18 &
Early arm lock: within the first three stops it penalises leaving the first-customer arm
exponentially in the angle; at the depot it favours the angular boundary of the unvisited set. \\

H19 &
Arm switching at low capacity: below 0.2 capacity it favours near-depot customers more than
$60^\circ$ from the current arm unless an exact fit exists; otherwise proximity, fit and angular clustering. \\
\bottomrule
\end{tabularx}
\end{table*}

\subsubsection{The surrogates emphasize tour progress and geometry differently} 
\label{sec:pomo_lehd}
\begin{figure}[t]
  \centering
  \includegraphics[width=.6\linewidth]{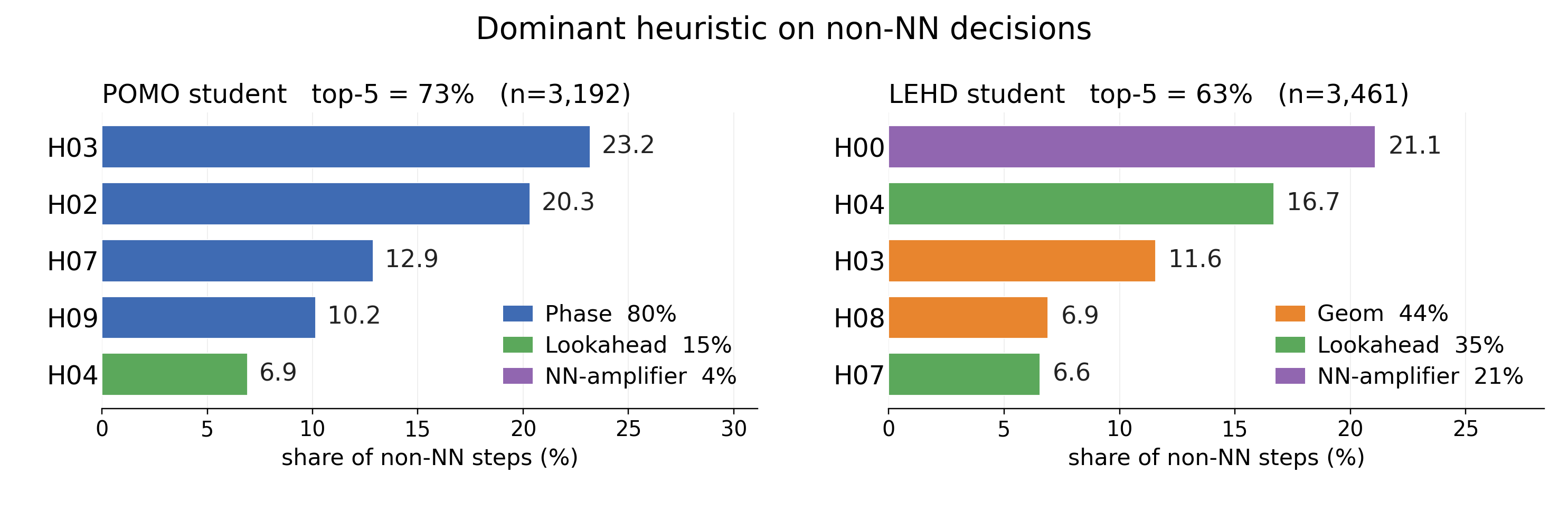}
  \caption{Dominant programs in the POMO and LEHD surrogates. Colors indicate categories.}
  \label{fig:nn_analysis}
\end{figure}

Figure~\ref{fig:nn_analysis} compares programs associated with non-NN decisions in the two surrogates. The POMO surrogate emphasizes phase-conditioned rules, shifting from early peripheral and isolated-city visits to late start-aware closure and multi-step completion. The LEHD surrogate emphasizes proximity, insertion and closure costs, directions relative to the start-to-current heading, and remaining-city geometry. Several LEHD programs strengthen closure preferences as fewer cities remain. Both combine geometry with progress-dependent adjustments. These differences may partly reflect the input constraints: POMO programs receive tour progress, whereas LEHD uses remaining-city counts.
\subsection{Code Example}
\label{sec: appendix_code}
In this appendix, we show two example codes learned by our student model.
\begin{figure}[!htb]
\centering
\begin{lstlisting}[style=pylight, caption={POMO TSP-100 H10 Code}, label={lst:h10}]
import torch

def heuristic(locs, current_node, mask, visited, step_norm, first_node):
    """
    POMO TSP-100 H10: Postponement-regret heuristic.

    Scores each feasible candidate by combining immediate proximity with
    a progress-scaled isolation bonus. As the tour progresses, the heuristic
    increasingly favors candidates whose nearest remaining neighbor is far away,
    i.e., nodes that may become expensive detours if postponed.
    """
    B, N, _ = locs.shape
    device = locs.device
    batch_idx = torch.arange(B, device=device)

    # Current city location: [B, 2]
    cur_loc = locs[batch_idx, current_node]

    # Immediate distance from current city to each candidate: [B, N]
    dist_to_cur = torch.norm(locs - cur_loc.unsqueeze(1), dim=-1)

    # Pairwise distances between all cities: [B, N, N]
    pairwise_dist = torch.norm(
        locs.unsqueeze(2) - locs.unsqueeze(1),
        dim=-1
    )

    # For each candidate i, find distance to its nearest other unvisited city.
    eye = torch.eye(N, device=device, dtype=torch.bool).unsqueeze(0)
    invalid_neighbor = (~mask).unsqueeze(1) | eye
    masked_pairwise = pairwise_dist.masked_fill(invalid_neighbor, 1e9)
    nearest_unvisited_dist = masked_pairwise.min(dim=-1).values

    # Progress-scaled postponement-regret weight.
    alpha = (1.5 + 2.0 * step_norm).unsqueeze(-1)

    # Higher is better: nearby nodes are good, but isolated nodes become
    # increasingly urgent as the tour progresses.
    logits = -dist_to_cur + alpha * nearest_unvisited_dist

    # Stand-alone safety: remove infeasible / already visited nodes.
    logits = logits.masked_fill(~mask, -1e9)

    return logits
\end{lstlisting}
\end{figure}

\newpage

\begin{figure}[ht]
\centering
\begin{lstlisting}[style=pylight, caption={LEHD TSP-100 H1 Code}, label={lst:h1}]
import math
import torch

def heuristic(unselected_locs, first_loc, current_loc):
    """
    LEHD TSP-100 H01: Coupled two-step closure lookahead.

    Scores each remaining city by combining the immediate current-to-candidate
    distance with the cheapest follow-up move from that candidate, where the
    follow-up node is also encouraged to be close to the tour start. This gives
    a local geometric correction rule that favors candidates forming a good
    two-step chain toward closure.
    """
    B, K, _ = unselected_locs.shape
    device = unselected_locs.device
    dtype = unselected_locs.dtype

    # K=1: forced choice.
    if K == 1:
        return torch.zeros(B, 1, device=device, dtype=dtype)

    # Distance from current city to each candidate: d(current, i)
    dist_current = torch.norm(
        unselected_locs - current_loc.unsqueeze(1),
        dim=-1
    )  # [B, K]

    # Distance from each candidate to the start city: d(i, start)
    dist_start = torch.norm(
        unselected_locs - first_loc.unsqueeze(1),
        dim=-1
    )  # [B, K]

    # Pairwise distances among remaining candidates: d(i, j)
    pairwise_dist = torch.cdist(unselected_locs, unselected_locs, p=2)
    eye = torch.eye(K, device=device, dtype=torch.bool).unsqueeze(0)
    pairwise_dist = pairwise_dist.masked_fill(eye, float("inf"))

    # K-aware closure weight: stronger when fewer nodes remain.
    sqrtK = math.sqrt(float(K))
    lam = min(1.0, 1.6 / sqrtK)
    w_step = 1.0 + 0.3 / sqrtK

    # Exact two-node tail behavior.
    if K == 2:
        lam = 1.0
        w_step = 1.0

    lam = torch.tensor(lam, device=device, dtype=dtype)
    w_step = torch.tensor(w_step, device=device, dtype=dtype)
    scale = torch.tensor(22.0 + 6.0 / sqrtK, device=device, dtype=dtype)

    # For each candidate i, choose the best follow-up j:
    # min_j [ d(i,j) + lambda * d(j,start) ], j != i.
    followup_cost = pairwise_dist + lam * dist_start.unsqueeze(1)
    best_followup = followup_cost.min(dim=2).values  # [B, K]

    # Lower cost is better, so return negative cost as score.
    total_cost = w_step * dist_current + best_followup
    scores = -scale * total_cost

    return scores
\end{lstlisting}
\end{figure}

\clearpage
\subsection{Interpreting each step - more examples}
\label{sec: more_inter_examaples}
\begin{wrapfigure}{r}{0.55\textwidth}
  \centering
  \includegraphics[width=\linewidth]{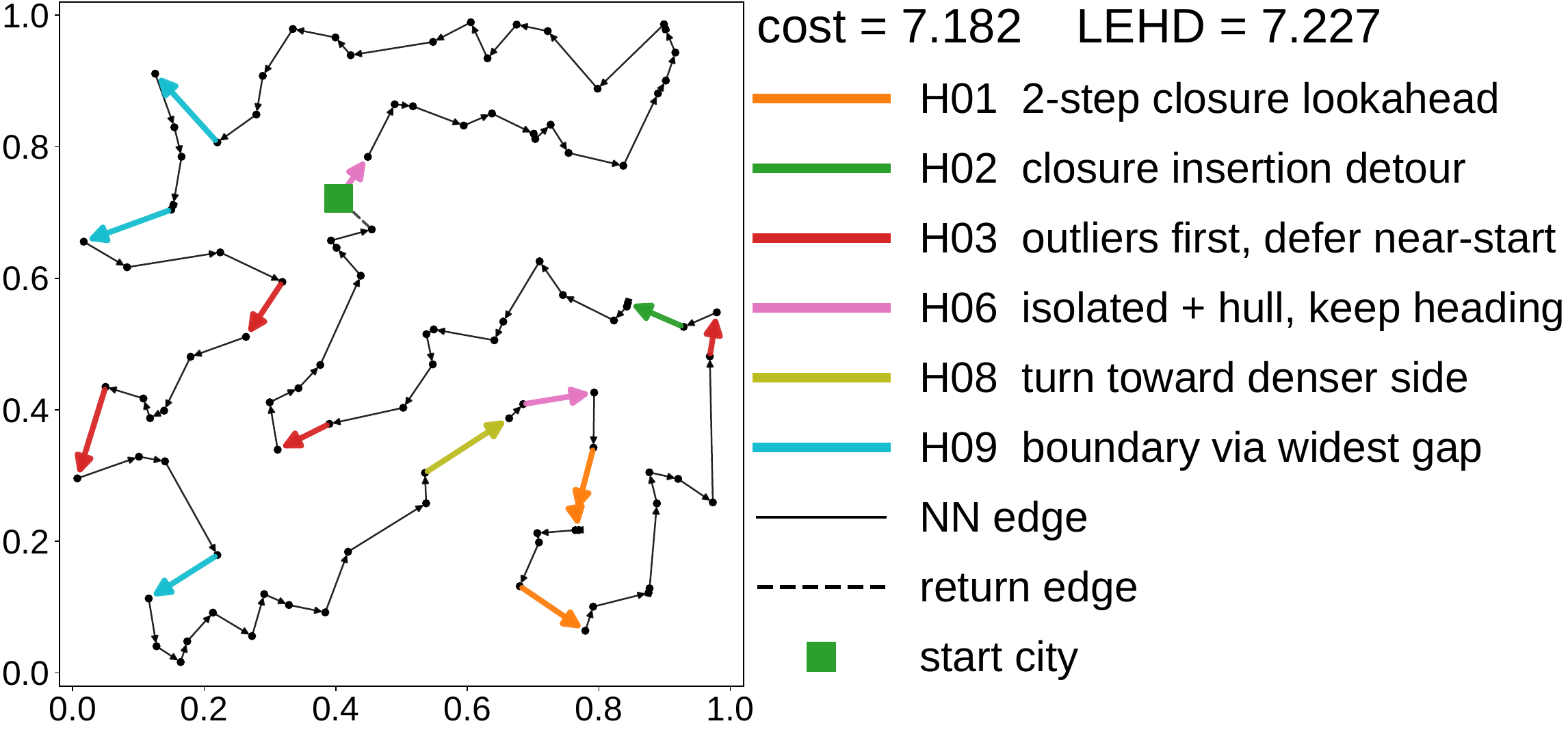}
  \caption{LEHD-TSP-100 Case study.}
  \label{fig:lehd_tsp_100_demo}
  \vspace{-5pt}
\end{wrapfigure}
\textbf{LEHD-TSP} Figure~\ref{fig:lehd_tsp_100_demo} links the selected LEHD student decisions to the highest-router-weight heuristic at each step. \textsc{H09} (\emph{boundary via widest gap}) favors candidates toward the widest empty sector facing away from the start. This matches the three cyan edges, which visit boundary cities before closer cities. \textsc{H06} (\emph{isolated + hull, keep heading}) rewards cities far from their nearest remaining neighbor. On the pink edge near $(0.7, 0.4)$, the student selects the most isolated of the 39 remaining cities over an NN candidate at almost the same distance from the current city, consistent with \textsc{H06}'s preferences for isolated cities. 

\begin{wrapfigure}{r}{0.55\textwidth}
  \centering
  \includegraphics[width=\linewidth]{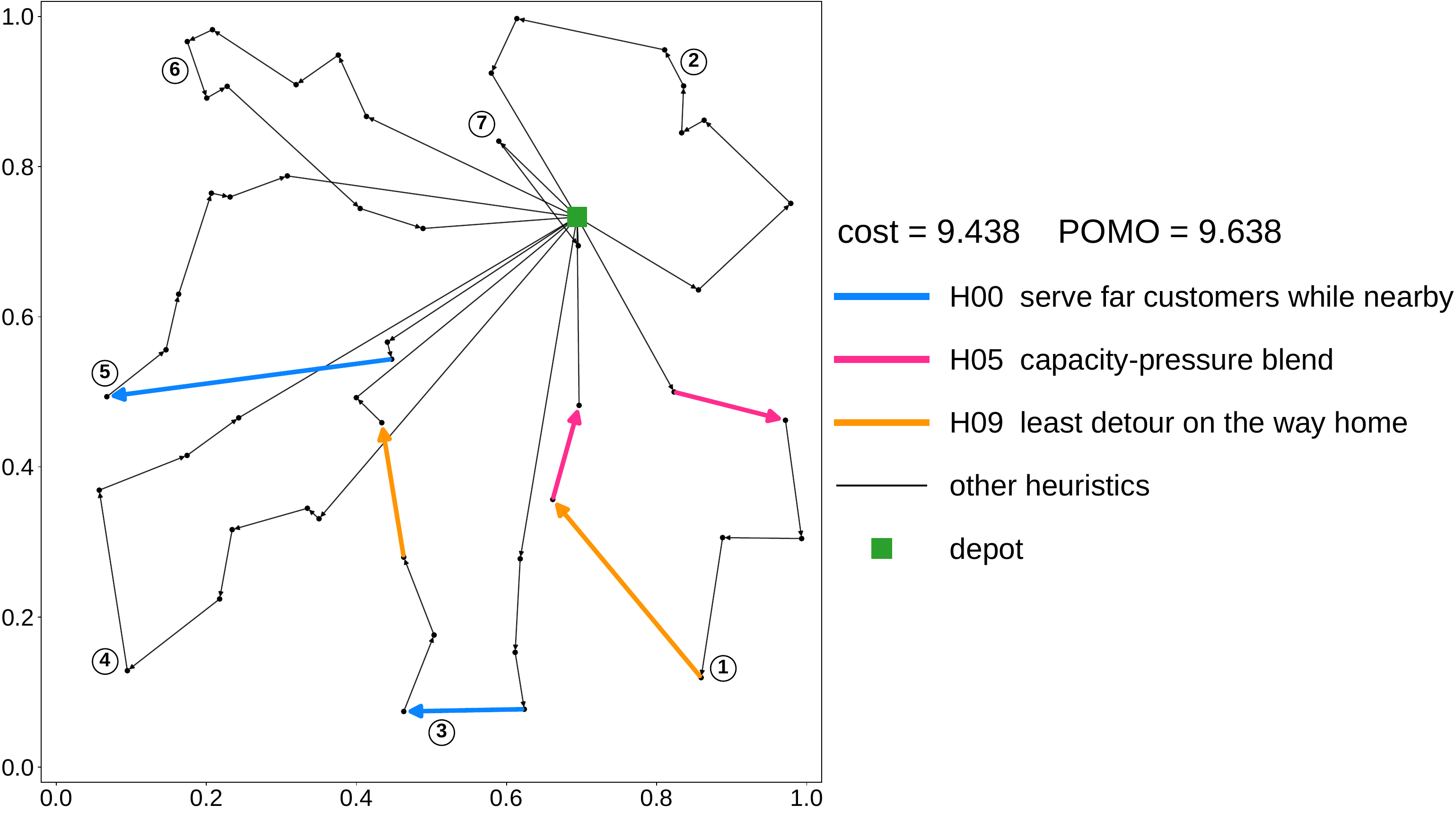}
  \caption{POMO-CVRP-50 Case study.}
  \label{fig:pomo_cvrp_50_demo}
  \vspace{-5pt}
\end{wrapfigure}
\textbf{POMO-CVRP} Figure~\ref{fig:pomo_cvrp_50_demo} links selected POMO student decisions to geometric preferences. Given the large number of heuristics, we focus on three for discussion. \textsc{H00} (\emph{serve far customers while nearby}) favors customers whose distance from the depot most exceeds their distance from the vehicle. On the blue edges of routes 3 and 5, the selected customers have the largest such difference among all feasible customers, so the vehicle reaches far customers while they are cheap to serve. \textsc{H05} (\emph{capacity-pressure blend}) favors, through its angular term, customers just ahead in the counter-clockwise direction around the depot.
 On the pink edges of route 1, the selected customers lie ahead of the vehicle in this direction, whereas the nearest feasible customers (NN) lie behind it, matching the program's preference for sweeping without stepping back. \textsc{H09} (\emph{least detour on the way home}) favors customers that add little detour on the way back to the depot. On the orange edges of routes 1 and 3, the selected customers add far less detour than the NN candidates, consistent with serving customers on the return trip.

\begin{wrapfigure}{r}{0.55\textwidth}
  \centering
  \includegraphics[width=\linewidth]{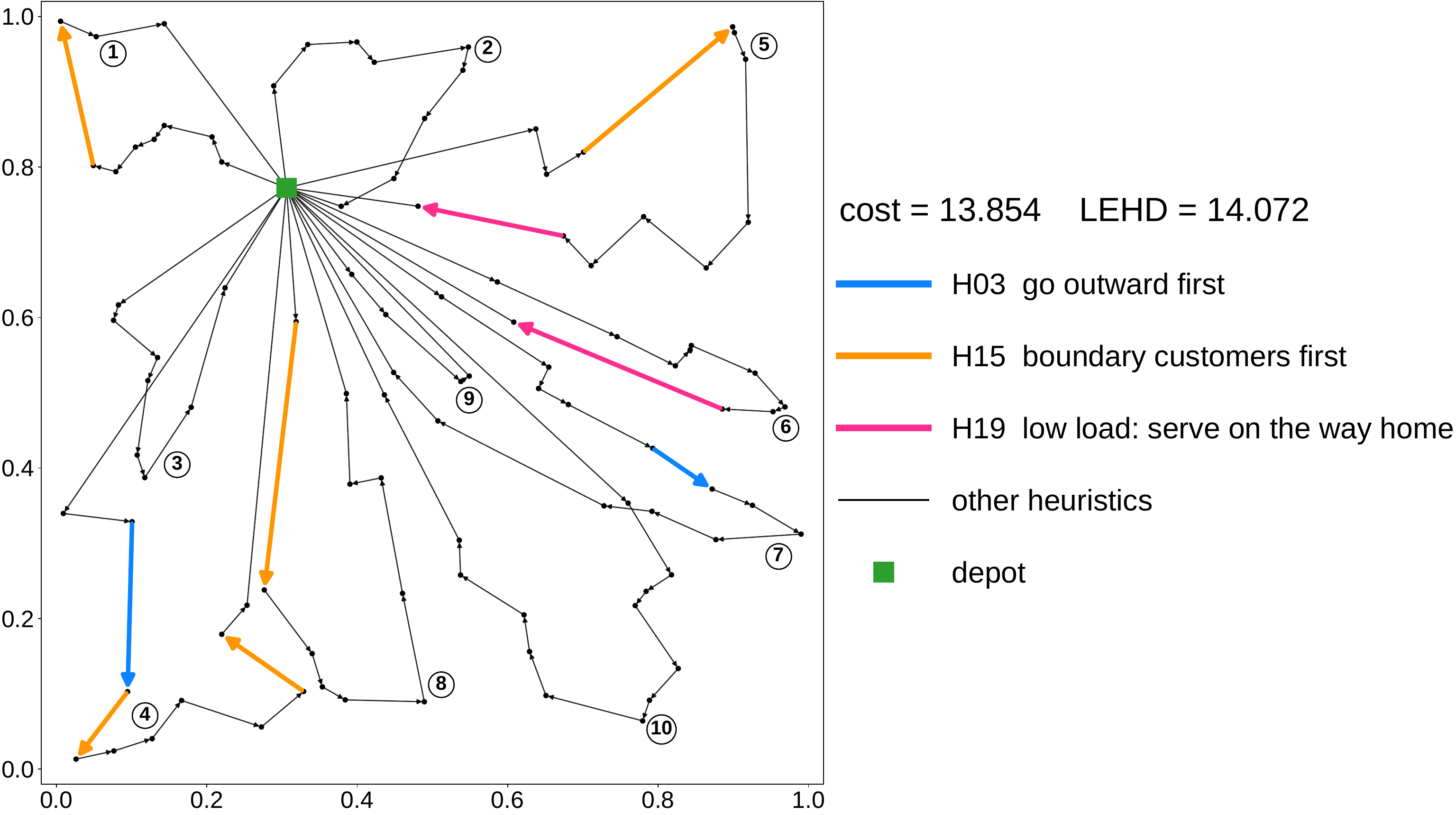}
  \caption{LEHD-CVRP-100 Case study.}
  \label{fig:lehd_cvrp_100_demo}
  \vspace{-5pt}
\end{wrapfigure}

\textbf{LEHD-CVRP} Figure~\ref{fig:lehd_cvrp_100_demo} links selected LEHD student decisions to geometric preferences. Given the large number of heuristics, we focus on three for discussion. \textsc{H15} (\emph{boundary customers first}) favors customers bordering the widest angular gap among unvisited customers around the depot. The orange edges, for example, in routes 4 and 8 are consistent with prioritizing the boundary of the remaining customer set. \textsc{H19} (\emph{serve on the way home}) favors depot toward customers when at most a quarter of the vehicle capacity remains. On the pink edges of routes 5 and 6, the selected customers lie almost directly along the current-to-depot direction, whereas the NN candidates are less aligned, consistent with serving customers along the return direction. \textsc{H03} (\emph{go outward first}) balances proximity to the current customer with a preference for customers farther from the depot. On the blue edges, it selects customers farther from the depot than the NN alternatives, matching the program's preference for outward continuation.

\subsection{Trustworthy Concept Attributions (Counterfactual Tests)}
\label{app:counterfactual}
We examine whether the concepts represented by the programs influence the student's predictions. We test hull peripherality in POMO-TSP, isolation in LEHD-TSP, and the homeward corridor in POMO-CVRP through matched interventions, tracking changes in program scores, routing weights, and action probabilities. We apply the same interventions to the teacher to compare the policies' responses.

\textbf{Intervention setup.} For each base state $B$, we construct a treatment state $T$ and a matched control state $P$. The student's choice in $B$ is the target $a$, and a reference action $r$ is selected using the rule for each test. We analyze base states where the teacher and student choose the same action and the student's preference for $a$ over $r$ is attributed mainly to programs implementing the tested concept. We compare the same $a$ and $r$ across the three states, keeping the reference node in place.

The treatment perturbs the problem instance to weaken the tested concept's support for $a$. The control matches the perturbation magnitude while preserving or strengthening that support. We compare the student's probability ratio for $a$ versus $r$, which we expect to be lower under treatment than control.

\textbf{Test Metrics.}
Let $\mathcal M$ denote the set of programs implementing the tested concept. For state $s$, program $m$ produces raw action scores $z_m(\cdot\mid s)$ and an action distribution $V_m(\cdot\mid s)$ through masked softmax. The router assigns each program a weight $w_m(s)$. The student distribution is
\begin{equation}
\pi_S(a\mid s)=\sum_m w_m(s)V_m(a\mid s).
\label{eq:cf_student_distribution}
\end{equation}

The carrier score sums the selected programs' score differences between $a$ and $r$:
\begin{equation}
E_{\mathcal M}(s)
=
\sum_{m\in\mathcal M}
\left[z_m(a\mid s)-z_m(r\mid s)\right].
\label{eq:E_S}
\end{equation}
This is a manipulation check and is not counted as independent evidence.

The routing log-weight compares the total routing weights of the selected programs and the rest of the bank:
\begin{equation}
R_{\mathcal M}(s)
=
\log\sum_{m\in\mathcal M}w_m(s)
-
\log\sum_{m\notin\mathcal M}w_m(s).
\label{eq:R_S}
\end{equation}

The program contribution to the student's probability margin between $a$ and $r$ is
\begin{equation}
C_{\mathcal M}(s)
=
\sum_{m\in\mathcal M}
w_m(s)\left[V_m(a\mid s)-V_m(r\mid s)\right].
\label{eq:C_S}
\end{equation}
Summing over the entire bank gives $\pi_S(a\mid s)-\pi_S(r\mid s)$ exactly. $C_{\mathcal M}$ measures the contribution of the selected programs as a whole.

Student log-odds and teacher log-odds measure each policy's relative preference for $a$ over $r$:
\begin{equation}
L_S(s)
=
\log\left[\pi_S(a\mid s)+\varepsilon\right]
-
\log\left[\pi_S(r\mid s)+\varepsilon\right],
\label{eq:L_S}
\end{equation}
\begin{equation}
L_T(s)
=
\log\left[\pi_T(a\mid s)+\varepsilon\right]
-
\log\left[\pi_T(r\mid s)+\varepsilon\right].
\label{eq:L_T}
\end{equation}
Here, $\pi_T$ is the teacher's action distribution and $\varepsilon$ is a small constant for numerical stability. For each metric $f$, the matched effect is
\begin{equation}
g_f=f(P)-f(T).
\label{eq:matched_effect}
\end{equation}
Each test is oriented so that $g_f>0$ is the predicted direction.

\subsubsection{POMO-TSP: hull peripherality}
\label{app:cf-hull}
We examine hull peripherality in the POMO-TSP-100 student. During early tour construction, H02 scores candidates as
\begin{equation}
\begin{aligned}
z_2(i)
&=1.5\,\mathrm{hb}(i)+0.5\,\mathrm{cont}(i)-4\,d_{\mathrm{cur}}(i),\\
\mathrm{hb}(i)
&=\frac{\lVert x_i-\mu\rVert}
{\max_{j\in U}\lVert x_j-\mu\rVert},\\
\mathrm{cont}(i)
&=\left\langle
\frac{x_i-x_c}{\lVert x_i-x_c\rVert},
\hat u_{\mathrm{out}}
\right\rangle.
\end{aligned}
\label{eq:hull-carrier}
\end{equation}
Here, $U$ is the set of unvisited cities and $\mu$ is their centroid. The positions of candidate $i$ and the current city are $x_i$ and $x_c$, respectively, with $d_{\mathrm{cur}}(i)=\lVert x_i-x_c\rVert$. The unit vector $\hat u_{\mathrm{out}}$ points from $\mu$ to $x_c$. The $\mathrm{hb}$ term rewards normalized distance from the centroid, while $\mathrm{cont}$ rewards alignment of the move from $x_c$ to $x_i$ with this outward direction. Together, they express the peripherality concept. The distance penalty favors nearby cities. We choose the candidate with the second-highest student probability in the base state as the reference $r$.

\begin{figure}[ht]
\centering
\includegraphics[width=0.8\linewidth]{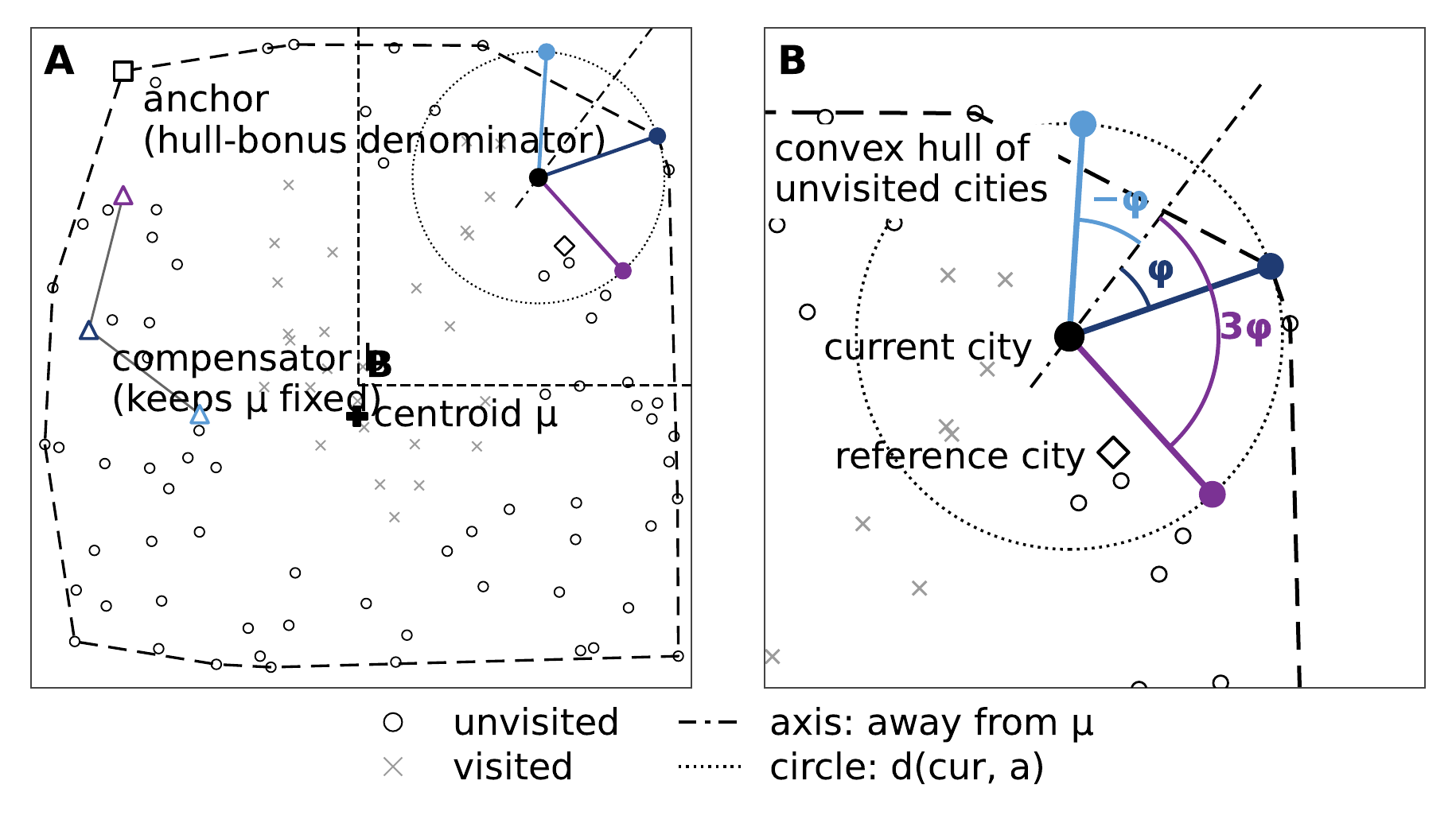}
\caption{Hull peripherality test on POMO-TSP. (A) Full instance. (B) Treatment reduces peripherality, while control preserves it.}
\label{fig:appx-hull}
\end{figure}
As shown in Figure~\ref{fig:appx-hull}, both the treatment and control move the target along a circle centered at the current city, preserving $d_{\mathrm{cur}}(a)$. The treatment rotates the target toward the centroid, reducing both $\mathrm{hb}(a)$ and $\mathrm{cont}(a)$. The control reflects the target across the line through the current city and the centroid, preserving both terms. The target moves by the same distance in both states.

To keep the centroid fixed, we move another unvisited city $b$ by an equal and opposite displacement in each state. We also preserve the maximum distance to the centroid used to normalize $\mathrm{hb}$ and match the change in the target's distance to the start city across treatment and control.

Table~\ref{tab:counterfactual} reports positive mean matched effects for the routing balance, program contribution, and student log-odds. On average, the treatment reduces the program's support for $a$ and the student's preference for $a$ over $r$ relative to the control. The teacher responds in the same direction.

\subsubsection{LEHD-TSP: isolation}
\label{app:cf-isolation}

We examine isolation in the LEHD-TSP-100 student. For candidate $i$, let $d_1(i)$ denote its distance to the nearest other unvisited city. The programs in $\mathcal{M}$ contain the following isolation terms:
\begin{equation}
\begin{aligned}
\mathrm{H03}:\quad
&2.0\,\kappa\,
\frac{d_1(i)-\min_k d_1(k)}
{\max_k d_1(k)-\min_k d_1(k)},\\
\mathrm{H05}:\quad
&2.0\,g\,
\frac{d_1(i)-\mathrm{mean}_k\,d_1(k)}
{\mathrm{sd}_k\,d_1(k)},\\
\mathrm{H06}:\quad
&2.1\,
\frac{d_1(i)}
{\mathrm{mean}_k\,d_1(k)}.
\end{aligned}
\label{eq:isolation-carriers}
\end{equation}
The normalization statistics are recomputed over the candidates in each state. The positive gates $\kappa$ and $g$ depend on the tour stage and remain fixed across $B$, $T$, and $P$. Each term favors candidates farther from their nearest neighbour. H03 also uses distance to the start city, and H06 includes a partner-gap term. We choose the candidate nearest to the current city as the reference $r$ and consider states where $a\neq r$.

\begin{figure}[ht]
\centering
\includegraphics[width=0.8\linewidth]{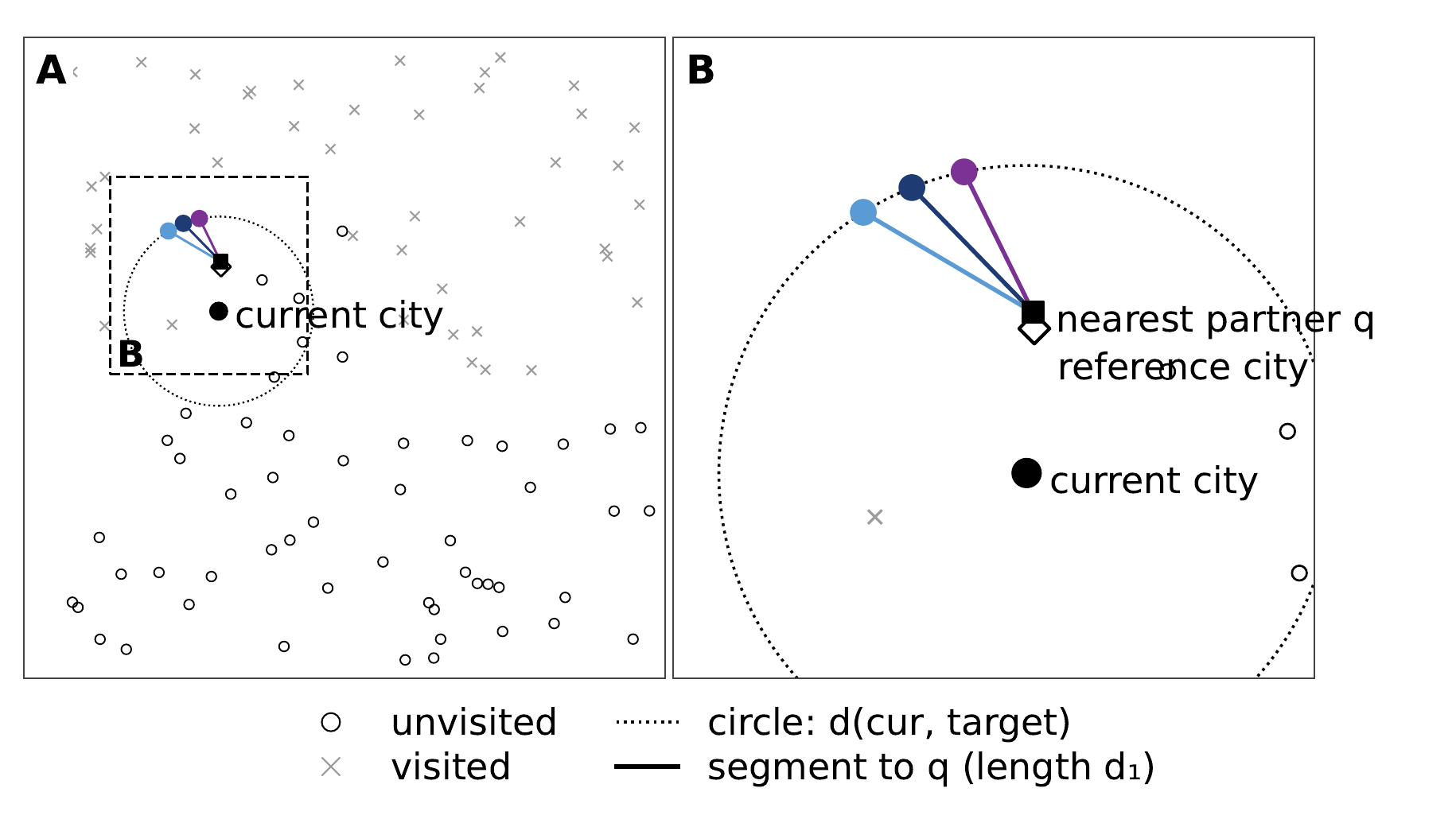}
\caption{Isolation test on LEHD-TSP. Treatment moves $a$ toward $q$, and control moves it away. Both match displacement and preserve distance to the current city.}
\label{fig:appx-isolation}
\end{figure}

As shown in Figure~\ref{fig:appx-isolation}, the treatment rotates the target $a$ about the current city toward its nearest unvisited neighbour $q$, reducing $d_1(a)$. The control rotates it by the same angle in the opposite direction, increasing $d_1(a)$. Both preserve $d_{\mathrm{cur}}(a)$ and move the target by the same distance.

Only the target moves. We require $q\neq r$ and preserve $q$ as the target's nearest neighbour in both states. The reference and its two nearest neighbours remain fixed, and the target stays outside that neighbourhood. This preserves the reference's first and second nearest-neighbour distances.

Since the control increases isolation, we report both changes relative to the base state:
\begin{equation}
\Delta_{\mathrm{down}}=f(B)-f(T),
\qquad
\Delta_{\mathrm{up}}=f(P)-f(B).
\label{eq:isolation-components}
\end{equation}
Their sum is the matched effect $g_f=f(P)-f(T)$.

Table~\ref{tab:appx-isolation} reports positive mean effects for both components across all five metrics. On average, the student favors $a$ less relative to $r$ when isolation decreases and more when it increases. The teacher shows the same pattern. 

\begin{table}[!htb]
\centering\footnotesize
\caption{Isolation intervention on LEHD-TSP. Entries give mean effects with 95\% instance-bootstrap confidence intervals and the percentage of states with positive effects in parentheses.}
\vspace{6pt}
\label{tab:appx-isolation}
\setlength{\tabcolsep}{3pt}
\begin{tabular}{@{}lccc@{}}
\toprule
\textbf{Metric} & \textbf{Down} & \textbf{Up} & \textbf{Total}\\
\midrule
Carrier score$^\dagger$
& $0.59$ [0.58, 0.60] (100\%)
& $0.61$ [0.60, 0.63] (100\%)
& $1.20$ [1.18, 1.23] (100\%)\\
Routing log-weight
& $0.16$ [0.10, 0.22] (74\%)
& $0.13$ [0.08, 0.19] (70\%)
& $0.29$ [0.20, 0.37] (72\%)\\
Program contribution
& $0.05$ [0.04, 0.06] (75\%)
& $0.03$ [0.02, 0.03] (70\%)
& $0.08$ [0.06, 0.09] (74\%)\\
Student log-odds
& $0.36$ [0.29, 0.43] (71\%)
& $0.18$ [0.15, 0.22] (66\%)
& $0.54$ [0.45, 0.63] (71\%)\\
Teacher log-odds
& $0.83$ [0.69, 1.02] (77\%)
& $0.64$ [0.55, 0.74] (74\%)
& $1.47$ [1.24, 1.74] (76\%)\\
\bottomrule
\end{tabular}

\smallskip
$^\dagger$Manipulation check only.
\end{table}

\subsubsection{POMO-CVRP: homeward corridor}
\label{app:cf-corridor}

We examine the homeward corridor concept in the POMO-CVRP-50 student. Let $c$ denote the current customer, $o$ the depot, $L$ the remaining capacity, and $\alpha_i$ the angle between the directions from $c$ to $o$ and from $c$ to candidate $i$. We measure the detour through $i$ and its perpendicular distance to the line through $c$ and $o$:
\begin{equation}
\begin{aligned}
D(i)&=d(c,i)+d(i,o)-d(c,o),\\
\rho_\perp(i)&=d(c,i)\,|\sin\alpha_i|,
\end{aligned}
\label{eq:corridor-features}
\end{equation}
where $d$ denotes Euclidean distance.

The programs in $\mathcal M=\{\mathrm{H03},\mathrm{H09}\}$ contain the following scoring terms:
\begin{equation}
\begin{aligned}
\mathrm{H03}:\quad
&-3\rho_\perp(i)-1.5d(c,i)\\
&+\mathbb{1}\big[0\leq d(c,i)\cos\alpha_i\leq d(c,o)\big]
+\text{demand terms},\\
\mathrm{H09}:\quad
&-6D(i)-2d(c,i)+\text{demand terms}.
\end{aligned}
\label{eq:corridor-carriers}
\end{equation}
Here, the demand terms depend only on the demand $\mathrm{dem}_i$ of customer $i$ and on $L$. H03 penalizes distance from the return line and rewards customers whose projection lies between $c$ and $o$. H09 penalizes detours on the return to the depot. We examine states with low remaining capacity and choose the nearest feasible customer as the reference $r$.

\begin{figure}[ht]
\centering
\includegraphics[width=0.8\linewidth]{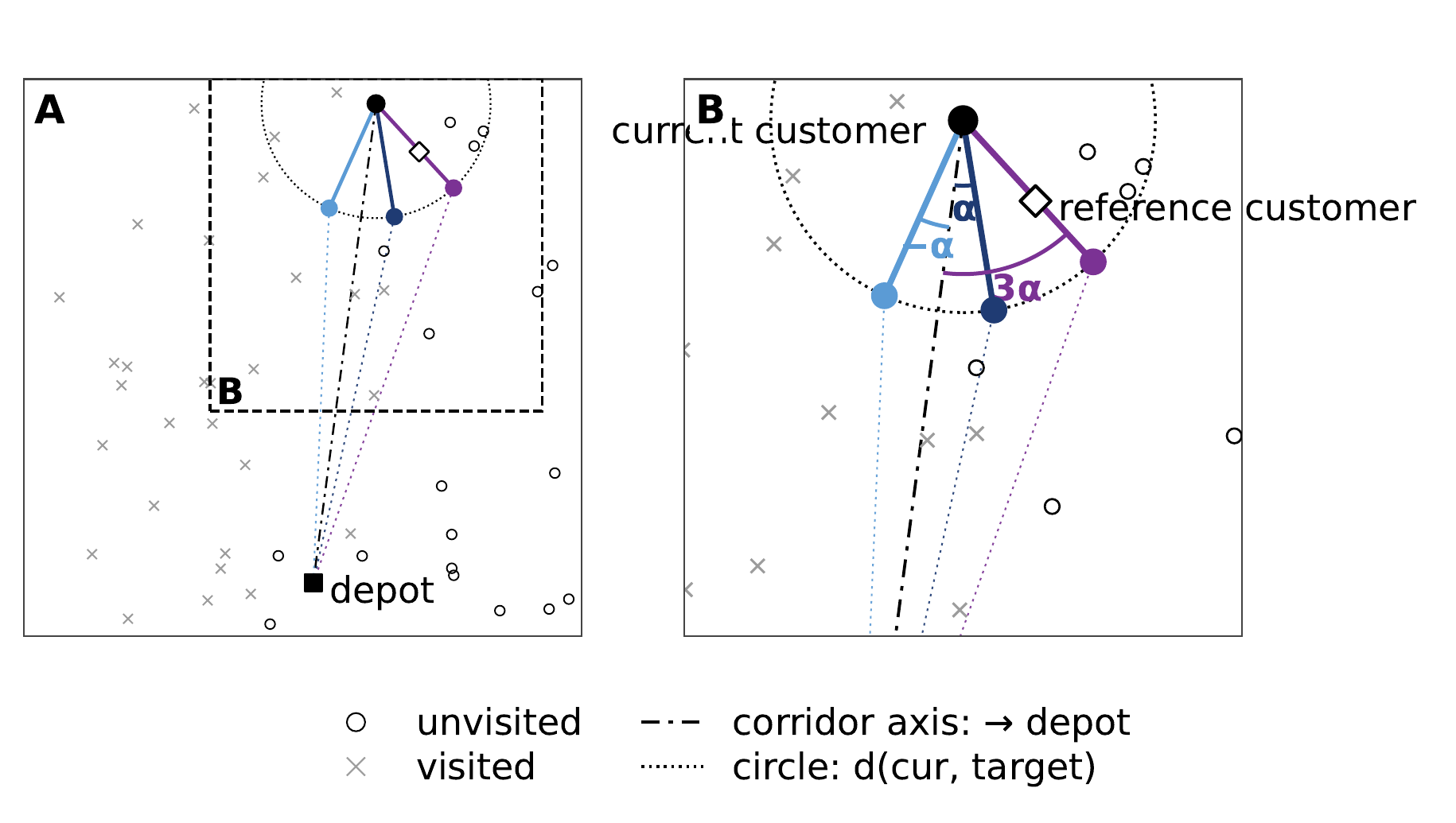}
\caption{Homeward corridor intervention on POMO-CVRP. Treatment and control match target displacement and preserve distance to the current customer. Demands, remaining capacity, and feasible actions stay fixed.}
\label{fig:appx-corridor}
\end{figure}

As shown in Figure~\ref{fig:appx-corridor}, the treatment rotates the target away from the direction to the depot, increasing both $D(a)$ and $\rho_\perp(a)$. The control reflects the target across the line through $c$ and $o$, preserving both quantities and its distance to the depot. Treatment and control move the target by the same distance and preserve $d(c,a)$.

Only the target moves. Demands, remaining capacity, the visited set, and feasible actions remain unchanged. The target's projection stays on the segment from $c$ to $o$, keeping the indicator term in H03 fixed. The reference remains the nearest feasible customer because all distances to the current customer are preserved.

We also fix the action distributions of all programs outside $\mathcal M$ to their base-state values and recompute routing weights to check whether the student's response persists.

\begin{table}[ht]
\centering
\caption{Homeward corridor intervention on POMO-CVRP. Mean student log-odds effects with 95\% instance-bootstrap CIs. Parentheses give the fraction of states with positive effects.}
\vspace{6pt}
\label{tab:appx-corridor}
\begin{tabular}{@{}lc@{}}
\toprule
\textbf{Setting} & \textbf{Student log-odds effect} \\
\midrule
Standard intervention
& $3.14$ [3.00, 3.28] (87\%) \\
Other program distributions fixed
& $2.65$ [2.53, 2.77] (88\%) \\
\bottomrule
\end{tabular}
\end{table}

Routing log-weight, program contribution, and student and teacher log-odds all show positive mean effects (Table~\ref{tab:counterfactual}). The mean student log-odds effect remains positive when all program distributions outside $\mathcal M$ are fixed to their base-state values and routing weights are recomputed (Table~\ref{tab:appx-corridor}).

\subsubsection{POMO-CVRP: local demand density}
\label{app:cf-sweep}

H02 rewards candidates surrounded by unvisited demand, scaled by their distance to the current node:
\begin{equation}
z^{\mathrm{dens}}(i)
=
\frac{D_{0.12}(i)}{d(c,i)^{2}}.
\label{eq:sweep-term}
\end{equation}
Here, $D_{0.12}(i)$ is the unvisited demand within radius $0.12$ of candidate $i$, including $i$ itself, and $c$ is the current customer. We use $\mathcal M=\{\mathrm{H02}\}$ and use the nearest feasible customer as the reference~$r$.

The treatment transfers just enough demand from customers near the target to customers outside the neighborhoods of both the target and the reference to make H02 prefer the reference. The control gives the same receivers the same amount from customers outside both neighborhoods, preserving the local demand of the target and the reference as well as the total unvisited demand. Customer locations, remaining capacity, the visited set, and feasible actions remain unchanged.

Holding H02's density term at its base-state values while recomputing the other programs and the router eliminates the student effect. This supports the density term's role in the student's response.

\begin{wraptable}{r}{0.50\textwidth}
\vspace{-\intextsep}
\centering
\footnotesize
\caption{Local demand density intervention on POMO-CVRP. Mean matched effects with 95\% instance-bootstrap CIs. Parentheses give the fraction of states with $g>0$.}
\label{tab:appx-sweep}
\vspace{4pt}
\setlength{\tabcolsep}{3pt}
\begin{tabular}{@{}lc@{}}
\toprule
\textbf{Metric} & \textbf{Effect} \\
\midrule
Carrier score$^\dagger$
& $5.65$ [5.05, 6.36] (100\%) \\
Routing log-weight
& $2.68$ [2.46, 2.91] (93\%) \\
Program contribution
& $0.43$ [0.41, 0.44] (100\%) \\
Student log-odds
& $1.38$ [1.27, 1.50] (97\%) \\
Teacher log-odds
& $0.97$ [0.57, 1.45] (56\%) \\
\bottomrule
\end{tabular}

\smallskip
$^\dagger$Manipulation check only.
\end{wraptable}

Table~\ref{tab:appx-sweep} shows positive mean effects for routing log-weight, program contribution, and student log-odds. The teacher responds in the same direction but more weakly than the student. Its mean log-odds effect is positive in each of three independent batches, yet only 56\% of states show a positive effect, compared with 97\% for the student, and its probability margin changes by $0.09$, compared with $0.30$ for the student. Its top choice changes under treatment in 10\% of states, compared with 43\% for the student.

\subsection{Trustworthy Program attribution (Knockout Tests).}
\label{app: program_att}
We remove each program in states where it dominates and renormalize the router weights. Mean next-node flip rates across programs are 65.2--98.1\%, versus 0.2--0.6\% when removing a uniformly sampled other program in the same states (Table~\ref{tab:dominance}). Removing the dominant program yields a higher flip rate than the control for all but H01 on POMO-CVRP, whose output is nearly uniform over customers and has little effect on the next-node choice.

\begin{table}[!htb]
\centering\footnotesize
\setlength{\tabcolsep}{4pt}
\caption{Dominance knockout. Program $m$ has the highest router weight in the evaluated states. Flip rates are means across programs with 95\% instance-bootstrap intervals.}
\vspace{6pt}
\label{tab:dominance}
\begin{tabular}{lccc}
\toprule
\textbf{Setting} & \textbf{Remove $m$ ($F_m$)} & \textbf{Remove random ($F_{\mathrm{rand}}$)} & \textbf{\# programs ($F_m>F_{\mathrm{rand}}$) / total} \\
\midrule
POMO TSP  & 94.3\% [93.8, 94.8] & 0.6\% [0.4, 0.8] & 14/14 \\
LEHD TSP  & 98.1\% [97.6, 98.5] & 0.2\% [0.1, 0.4] & 11/11 \\
POMO CVRP & 65.2\% [64.6, 65.7] & 0.5\% [0.5, 0.6] & 19/20 \\
\bottomrule
\end{tabular}
\end{table}
\subsection{Trade-off Between Code Complexity and Solution Quality}
\label{app:bank_readability}

\paragraph{Setup.}
We repeat distillation on TSP-100 with limits on program complexity. Constrained programs access the state through a fixed set of primitives: distance to the current node, distance to the start node, nearest-neighbour distance, local density, heading cosine, distance to the centroid of unvisited nodes, and the number of unvisited nodes. POMO also includes tour progress. These primitives are implemented in a shared handwritten library with 33 statements for POMO and 21 for LEHD. Unconstrained programs compute their features directly from the raw state.

We check each program's syntax tree during initialization, revision, and addition. The \emph{simple} setting allows at most 2 distinct primitives, 15 statements, and 1 branch. The \emph{medium} setting allows 3 primitives, 30 statements, and 2 branches. Branches include \texttt{if}, \texttt{where}, and \texttt{masked\_fill}. Both settings forbid loops, sorting, and access to raw inputs outside primitive calls. The \emph{unconstrained} setting uses the default pipeline without these restrictions.

All distillation settings use the same LLM (\texttt{deepseek-v4-flash}). We evaluate solution quality along the pruning path of each learned bank, removing programs one at a time using validation top-1 agreement. Table~\ref{tab:complexity} compares coding-agent policies with EPB banks at the sizes with the lowest evaluated search gaps.

\paragraph{Metrics.}
We remove comments, blank lines, and docstrings and regenerate the source from its syntax tree before measuring complexity. \emph{Statements} counts statement nodes, including nested statements, but excludes imports, the function signature, and the final return of the score vector. \emph{Halstead volume} is
\[
V=(N_1+N_2)\log_2(n_1+n_2),
\]
where $n_1,n_2$ count distinct operators and operands, and $N_1,N_2$ count their occurrences. We compute it with \texttt{radon}~6.0.1. \emph{Coefficients} counts numeric weights, thresholds, and exponents, excluding indices, dimension arguments, numerical-stability constants, and mask fill values.

All complexity measurements in Table~\ref{tab:complexity} include primitive implementations. Per-program measurements expand each called primitive and its helper functions before averaging across programs. Whole-bank measurements count each used primitive and helper once. For coding-agent policies, per-program and whole-bank measurements coincide and are reported only in the whole-bank columns.

\begin{figure}[ht]
\centering
\begin{minipage}[t]{0.49\linewidth}
\centering
\includegraphics[width=\linewidth]{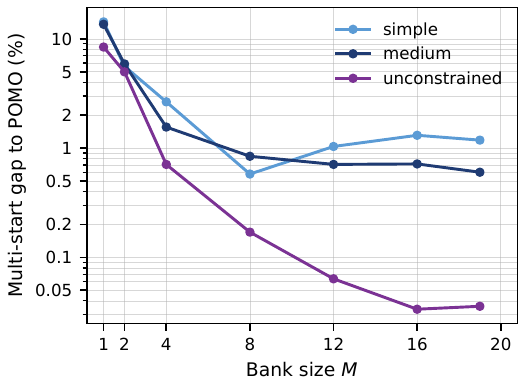}\\[-2pt]
{\small (a) POMO-TSP-100}
\end{minipage}\hfill
\begin{minipage}[t]{0.49\linewidth}
\centering
\includegraphics[width=\linewidth]{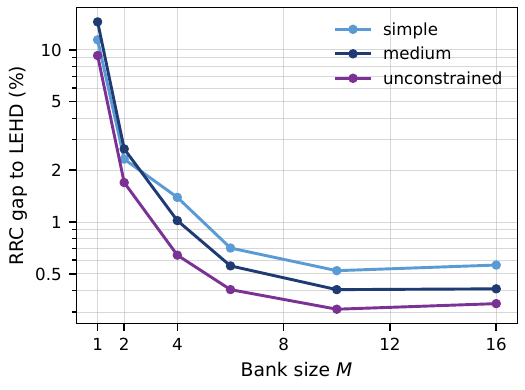}\\[-2pt]
{\small (b) LEHD-TSP-100}
\end{minipage}
\caption{Search gap to the teacher along the pruning path of each bank, for bank sizes $M\le20$. Programs are removed one at a time using validation top-1 agreement.}
\label{fig:prune-cost}
\end{figure}

\begin{table}[ht]
\centering
\footnotesize
\caption{Program complexity and search gaps to the teacher on TSP-100. Complexity includes primitives and is reported as per-program means and whole-bank totals.}
\vspace{6pt}
\label{tab:complexity}
\label{tab:agent-vs-bank}
\setlength{\tabcolsep}{3.5pt}
\begin{tabular}{lrrrrrrrr}
\toprule
& & & \multicolumn{3}{c}{\textbf{Per program}}
    & \multicolumn{3}{c}{\textbf{Whole bank}} \\
\cmidrule(lr){4-6}\cmidrule(lr){7-9}
\textbf{Policy} & $M$ & \textbf{Gap (\%)}
& Stmt. & Vol. & Coef.
& Stmt. & Vol. & Coef. \\
\midrule

\multicolumn{9}{l}{\textbf{POMO-TSP-100}} \\
\addlinespace[2pt]
Agent: GPT-6
& 1 & 4.58
& \multicolumn{3}{c}{}
& 52 & 1{,}262 & 29 \\
Agent: Fable 5.1
& 1 & 3.90
& \multicolumn{3}{c}{}
& 40 & 919 & 17 \\

\addlinespace
EPB bank: simple
& 8 & 0.579
& 11.4 & 116 & 2.9
& 51 & 573 & 20 \\
EPB bank: medium
& 29 & 0.455
& 14.5 & 138 & 4.5
& 154 & 2{,}147 & 85 \\
EPB bank: unconstrained
& 16 & 0.034
& 33.6 & 655 & 13.0
& 537 & 10{,}485 & 208 \\

\midrule
\multicolumn{9}{l}{\textbf{LEHD-TSP-100}} \\
\addlinespace[2pt]

Agent: GPT-6
& 1 & 2.09
& \multicolumn{3}{c}{}
& 45 & 1{,}072 & 28 \\
Agent: Fable 5.1
& 1 & 1.64
& \multicolumn{3}{c}{}
& 38 & 831 & 16 \\

\addlinespace
EPB bank: simple
& 10 & 0.333
& 8.8 & 129 & 2.5
& 49 & 609 & 23 \\
EPB bank: medium
& 28 & 0.255
& 14.5 & 242 & 6.5
& 206 & 4{,}089 & 152 \\
EPB bank: unconstrained
& 10 & 0.172
& 41.9 & 943 & 19.1
& 419 & 9{,}426 & 191 \\
\bottomrule

\end{tabular}
\end{table}

\paragraph{Complexity and solution quality.}
Table~\ref{tab:complexity} reports complexity at both the program and bank levels. Including shared primitive implementations, the simple banks contain 51 statements for POMO and 49 for LEHD, compared with 537 and 419 for the unconstrained banks. Their Halstead volumes are also substantially lower, decreasing from 10{,}485 to 573 for POMO and from 9{,}426 to 609 for LEHD. Despite these reductions in program code, the simple banks retain search gaps of 0.579\% and 0.333\% to their respective teachers. These results show that combining compact programs can preserve near-teacher solution quality with substantially less program code.

\paragraph{Bank size and complexity trade-off.}
Figure~\ref{fig:prune-cost} shows that, within a given program complexity setting, improvements in solution quality tend to saturate as the bank grows, with limited benefits from further expansion. For the simple and medium settings, this may partly reflect the fixed primitive vocabulary: additional programs combine the same features and may overlap with existing decision rules, limiting the benefits of increasing their number. Meanwhile, relaxing the constraints from simple to medium and unconstrained further reduces the search gaps, but also substantially increases whole-bank code complexity (Table~\ref{tab:complexity}). From simple to unconstrained, the absolute gap reductions are approximately 0.55 percentage points for POMO and 0.16 for LEHD, accompanied by increases in statement counts, Halstead volume, and coefficient counts. Bank configuration therefore requires jointly considering program count and per-program complexity, balancing additional gains in solution quality against the resulting code complexity.

\paragraph{Comparison with coding-agent policies.}
Table~\ref{tab:complexity} compares standalone coding-agent policies with EPB program banks. Including primitive implementations, the simple EPB banks contain 51 and 49 statements, comparable in scale to the agents' 40 to 52 statements for POMO and 38 to 45 for LEHD. Their search gaps are 0.579\% and 0.333\%, compared with 3.90\% and 1.64\% for the better coding-agent policy in each setting. Thus, the simple banks achieve lower solution costs with a similar amount of program code. This comparison concerns program code, as EPB additionally uses a learned router.

\begin{figure}[!htb]
\centering

\begin{lstlisting}[
  style=pylight,
  caption={POMO TSP Simple Tier Code: 2 primitives, 3 statements, Halstead volume 37, 2 coefficients.},
  label={lst:pomo-simple}
]
import torch
from tsp_feats import feat

def heuristic(locs, current_node, mask, visited, step_norm, first_node):
    # Uses 2 concept(s): dist_cur, heading_cos
    d = feat.dist_cur(locs, current_node)
    h = feat.heading_cos(locs, current_node, visited)

    # Combines proximity with a heading-alignment bonus.
    logits = -6.0 * d + 1.5 * h
    return logits
\end{lstlisting}

\vspace{4pt}

\begin{lstlisting}[
  style=pylight,
  caption={POMO TSP Medium Tier Code: 3 primitives, 4 statements, Halstead volume 66, 3 coefficients.},
  label={lst:pomo-medium}
]
import torch
from tsp_feats import feat

def heuristic(locs, current_node, mask, visited, step_norm, first_node):
    # 3 concepts: dist_cur (greedy), dist_start (closing leg), nn_dist (isolation)
    d_cur = feat.dist_cur(locs, current_node)
    d_start = feat.dist_start(locs, first_node)
    nn = feat.nn_dist(locs, mask)

    # Favors nearby candidates and proximity to the start,
    # with a small isolation bonus.
    logits = -3.0 * d_cur - 1.5 * d_start + 0.2 * nn
    return logits
\end{lstlisting}

\caption{Example programs from the POMO-TSP-100 banks.
\texttt{dist\_cur} and \texttt{dist\_start} give distances to the current
and start nodes. \texttt{heading\_cos} measures heading alignment,
and \texttt{nn\_dist} gives the distance to the nearest other unvisited node.
Complexity values in the listing captions refer to the displayed program bodies and exclude primitive implementations.}
\label{fig:complexity-examples}
\end{figure}

\subsection{Improvements through Program Edits}
\label{app:intervention}

We edit the student on two instance families outside the training distribution: TSP-100 and CVRP-50 instances on a thin diagonal band. Both the teacher and the student perform poorly on these families. The interventions change only program code, and the router and all network weights remain frozen.

Each intervention introduces a closed-form rule that scores candidate nodes without search or route planning. An instance-level shape check restricts the rule to its target family, so its contribution is zero on the uniform instances used for training. The rule modifies program scores, while the router continues to combine the programs and choose the next node.

The rules were designed on a separate set of instances. All results below are on 256 held-out instances per family that played no part in their design. Gaps are the mean excess over LKH for TSP and HGS for CVRP. To test whether the same rule also helps the teacher, we add its score as a bias to teacher's (the POMO model) log-probabilities, $\arg\max_j[\log p_{T}(j)+\gamma z_j]$, where $z$ is the standardised rule score, and report the best $\gamma$.

\subsubsection{TSP-100, diagonal band}
\label{app:int_band}

On a thin band, an efficient tour travels along one side and returns along the other. All 14 programs of the student favour candidates close to the current node, and none reserves nodes for the return leg. The student consequently sweeps both sides on the way out, crosses the band frequently, and leaves scattered nodes to collect before closing the tour with a long edge back to the start.

The edit encourages the student to follow one side outward and the other back, without skipping nodes on the return leg. We add three lane-keeping programs and modify the original 14 programs where they compute the distance from the current node to each candidate. Immediately after that computation, we add penalties to the distances of candidates that break the two-lane pattern, and the rest of each program is unchanged. The penalties discourage choosing nodes on the opposite side during the outward leg, moving against the direction of travel, and skipping nodes on the return leg. They are assigned to the 7 nearest neighbor, 3 phase-based, and 4 closing programs according to their roles. We also add the lane-keeping rule's score to the final scores of all 17 programs. Before addition, the rule score is standardised and scaled to one standard deviation of the receiving program's scores.

For TSP-100, the edit reduces EPB's greedy gap from 24.16\% to 7.01\% and its search gap from 8.70\% to 1.78\% (Table~\ref{tab:intervention}). Edited EPB outperforms both the teacher and the teacher with the same rule in both settings. After the edit, the number of nodes left for the return lane rises from 15.4 to 40.6, close to LKH's 42.1. The median closing-edge length falls from 21.2 to 3.2 nearest-neighbour spacings, compared with 1.6 for LKH. Figure~\ref{fig:band_tsp_before_after} illustrates this change on one held-out instance.  

\subsubsection{CVRP-50, diagonal band}
\label{app:int_band_cvrp}

On diagonal-band instances, the customers lie on a thin band along the diagonal of the square. Good solutions cut the band into consecutive stretches and serve each stretch with one route. EPB's routes instead drive out to the far parts of the band and take only part of the customers there, so later routes cover the same stretches again: 4.79 routes per instance reach the outer half of the band, against 4.25 for HGS. The 20 programs score customers by distance, angle and demand fit, and none of them divides the band into stretches, giving the router little support for this structure.

We introduce three rules. At the depot, the start rule favours the customers farthest out along the band on one side of the depot, as many as fit in one vehicle, unless going out along one side of the band and back along the other is shorter. While customers of the current stretch lie farther out, the route rule favours keeping to the vehicle's side of the band and moving outward; otherwise it favours the outermost remaining customer. The return rule penalises returning to the depot when this would force one more route and a cheap continuation exists. Each score is standardised, scaled to 8, 1 and 2 times the standard deviation of each program's own scores, and added to the final scores of all 20 programs. For instances that are not bands, the rules are zero, and the programs' outputs are unchanged.

The edit changes EPB's routes as intended. The routes reaching the outer half of the band fall from 4.79 to 4.40 per instance (HGS 4.25), and the excess of twice the farthest depot distance summed over routes falls from 7.2\% to 2.0\% of the HGS cost. EPB's greedy gap falls from 10.88\% to 5.94\%, and its search gap from 4.17\% to 2.43\% (Table~\ref{tab:intervention}). Both gaps are below those of the teacher and the teacher with the same rules.

\subsection{Overly shallow routing}
\label{sec: router_size}

\noindent
\begin{minipage}[t]{0.46\textwidth}
\vspace{0pt}
With the program bank fixed, the one-layer router yields higher held-out KL and larger greedy and multi-start gaps than every deeper configuration. At $D=128$, increasing depth from one to two layers reduces the greedy gap from 16.33\% to 10.03\% and the multi-start gap from 3.70\% to 2.10\%. The narrower four-layer router also outperforms the one-layer model despite having fewer parameters. These results support retaining a sufficiently expressive neural router for state-dependent program composition, while further increases in model size do not consistently improve performance.
\end{minipage}\hfill
\begin{minipage}[t]{0.52\textwidth}
\vspace{0pt}
\centering
\small
\captionof{table}{Router capacity with a frozen program bank ($M=14$). Cost gaps (\%) are relative to the POMO teacher ($\downarrow$).}
\label{tab:router_capacity}
\vspace{6pt}
\setlength{\tabcolsep}{3pt}
\begin{tabular}{lrrr}
\toprule
\textbf{Configuration} & \textbf{Parameters} & \textbf{Greedy gap} & \textbf{MS gap} \\
\midrule
$L=4,\ D=64$  & 213,504   & \textbf{8.73} & \textbf{1.63} \\
$L=1,\ D=128$ & 249,984   & 16.33 & 3.70 \\
$L=2,\ D=128$ & 448,256   & 10.03 & 2.10 \\
$L=3,\ D=128$ & 646,528   & 11.57 & 2.34 \\
$L=4,\ D=128$ & 844,800   & 11.15 & 2.15 \\
$L=6,\ D=128$ & 1,241,344 & 9.56  & 1.72 \\
$L=4,\ D=256$ & 3,360,768 & 10.83 & 2.05 \\
\bottomrule
\end{tabular}
\end{minipage}
\par\medskip
\section{LLM Statement}
\label{sec:llm}
LLMs are a core component of EPB, generating and revising programs as detailed in \S\ref{sec:method} and \S\ref{app:promptsetting}. They were also used to assist with writing, drawing figures, and routine coding. The authors reviewed and verified all AI-assisted content and take full responsibility for the paper.
\section{Social Impact Statement}
\label{app:social_impact}

EPB may improve the transparency and auditability of neural routing policies by exposing executable decision rules and enabling targeted modifications. These capabilities may help researchers inspect model behavior, diagnose failures, and evaluate policy changes.

This study focuses on synthetic routing benchmarks and does not involve human subjects, sensitive personal data, or real-world deployment. We do not foresee direct negative societal impacts within this scope. Nevertheless, the surrogate approximates the teacher's behavior, and its explanations do not establish the teacher's internal reasoning. Practical deployment should therefore independently validate the surrogate's fidelity, solution quality, and robustness under the intended conditions.

\section{License}
\label{sec:license}
Our experiments use publicly available checkpoints and code assets from prior neural combinatorial optimization models, including POMO~\citep{kwon2020pomo} and LEHD~\citep{luo2023neural}, as teacher policies and/or baselines. 
For POMO, we use the publicly released repository associated with the NeurIPS 2020 paper. We did not find an explicit license file or named software license in the public repository, and therefore use the checkpoint only for academic research comparison and policy distillation. We do not redistribute the original or modified POMO checkpoints.
For LEHD, we use the publicly released repository associated with the NeurIPS 2023 paper. The repository states that the code is copyrighted by CIAM Group and can only be used for non-commercial purposes; commercial use requires contacting the authors. We use the LEHD checkpoint only for non-commercial academic research, follow the stated usage restriction, and do not redistribute the original or modified LEHD checkpoints.
\end{document}